\documentclass{article}

\usepackage{amssymb}
\usepackage{latexsym}

\usepackage{url}
\usepackage[table]{xcolor}
\definecolor{newcolor}{rgb}{.8,.349,.1}
\usepackage{multirow}
\usepackage{color}

\usepackage{float}
\usepackage[utf8]{inputenc}  
\usepackage{subfiles} 
\usepackage[square,numbers,sort&compress]{natbib}
\usepackage{amssymb}
\usepackage{mathtools}
\usepackage{gensymb}
\usepackage{authblk}
\usepackage[margin=0.8in]{geometry}
\usepackage[font=footnotesize]{subfig}
\usepackage{subfiles}
\usepackage[frenchb]{babel}

\usepackage{amsmath}
\usepackage[pdftex]{graphicx}

\usepackage{array}
\newcolumntype{M}[1]{>{\centering\arraybackslash}m{#1}}

\usepackage{hyperref}

\definecolor{lighterGray}{gray}{0.95}
\definecolor{Gray}{gray}{0.9}
\definecolor{darkerGray}{gray}{0.8}

\begin{document}

\title{Surrogate-based variational data assimilation for tidal modelling}

\author[1,2]{Rem-Sophia Mouradi}
\author[1]{Cédric Goeury}
\author[2,3]{Olivier Thual}
\author[1]{Fabrice Zaoui}
\author[1,4]{Pablo Tassi}

\affil[1]{EDF R\&D, National Laboratory for Hydraulics and Environment (LNHE), 6 Quai Watier, 78400 Chatou, France}
\affil[2]{Climate, Environment, Coupling and Uncertainties research unit (CECI) at the European Center for Research and Advanced Training in Scientific Computation (CERFACS), French National Research Center (CNRS), 42 Avenue Gaspard Coriolis, 31820 Toulouse, France}
\affil[3]{Institut de M\'ecanique des Fluides de Toulouse (IMFT), Universit\'e de Toulouse, CNRS, Toulouse, France}
\affil[4]{Saint-Venant Laboratory for Hydraulics (LHSV), Chatou, France}

\date{\today}
\maketitle

\begin{abstract}
Data assimilation (DA) is widely used to combine physical knowledge and observations. It is nowadays commonly used in geosciences to perform parametric calibration. In a context of climate change, old calibrations can not necessarily be used for new scenarios. This raises the question of DA computational cost, as costly physics-based numerical models need to be reanalyzed. Reduction and metamodelling represent therefore interesting perspectives, for example proposed in recent contributions as hybridization between ensemble and variational methods, to combine their advantages (efficiency, non-linear framework). They are however often based on Monte Carlo (MC) type sampling, which often requires considerable increase of the ensemble size for better efficiency, therefore representing a computational burden in ensemble-based methods as well. To address these issues, two methods to replace the complex model by a surrogate are proposed and confronted: (i) PODEn3DVAR directly inspired from PODEn4DVAR, relies on an ensemble-based joint parameter-state Proper Orthogonal Decomposition (POD), which provides a linear metamodel; (ii) POD-PCE-3DVAR, where the model states are POD reduced then learned using Polynomial Chaos Expansion (PCE), resulting in a non-linear metamodel.  Both metamodels allow to write an approximate cost function whose minimum can be analytically computed, or deduced by a gradient descent at negligible cost. Furthermore, adapted metamodelling error covariance matrix is given for POD-PCE-3DVAR, allowing to substantially improve the metamodel-based DA analysis. Proposed methods are confronted on a twin experiment, and compared to classical 3DVAR on a measurement-based problem. Results are promising, in particular superior with POD-PCE-3DVAR, showing good convergence to classical 3DVAR and robustness to noise.

\end{abstract}


\shorthandoff{:}
\section{Introduction}
 Data assimilation (DA) \citep{Asch2016} is a powerful mathematical approach to produce trustworthy simulations, accounting for both measurements and results of physics-based numerical models, while taking into consideration their respective uncertainties. It is applied in geosciences \citep{Carrassi2018}, from atmospheric and oceanographic forecasting models \citep{Navon2009,Martin2015}, to fire front tracking \citep{Rochoux2014a}, hydrodynamics \citep{Sorensen2004,Altaf2009,Larnier2020}, morphodynamics \citep{Scott2007,Smith2013,Evangelista2017}, etc. Interest in such methods is enhanced in a context of climate change, where new data constantly need to be accounted for \citep{Rolnick2019}, and standard calibrations, specifically optimal values obtained from fitting on old measurements, are not necessarily suitable to be applied for new scenarios \citep{Rochoux2014a}. Thus, the same model, which is often complex and computationally costly, has to be reanalyzed for different scenarios. Additionally, a constant increase in data sources is registered \citep{Karpatne2019}, as the new SWOT satellite mission \cite{Swot2021,Morrow2019}.  \\

DA can for example be employed to fit modelling parameters as new data arrive, which is called \textit{parametric calibration} and is the focus of the proposed study. In this context, two DA techniques are commonly used: (i) ensemble methods \citep{Evensen2009} where a set of parameterizations is studied according to their probabilities, and (ii) variational methods \cite{Asch2016}, where a mismatch between different data sources, formalized by a \textit{cost function}, is minimized. Each technique has its own advantages and drawbacks. On the one hand, ensemble-based methods are known to be computationally efficient for moderate control/state dimensions, as they allow faster exploration of the inputs and outputs spaces. They are able to account for non-Gaussian information through sampling, although relying on a Gaussian framework \citep{Carrassi2018}. However, sampling errors accumulate through assimilation cycles and increase with large assimilation windows, which impacts the accuracy of ensemble methods. Small ensembles can overestimate the inter-ensemble correlations, resulting in a single trajectory through assimilation cycles for all members, which is the so-called \textit{filter divergence} \citep{Bannister2017}. Furthermore, small ensembles are known to systematically underestimate the variances, resulting in underestimated calibration errors \citep{Bannister2017}. Solutions as \textit{localization} (moderation of covariances) or \textit{inflation} (forced increase of variances) are classically used, but are generally case specific and consequently require case-by-case modelling effort \citep{Bannister2017}. This problem can only be resolved by increasing the ensemble size \citep{Li2008} which in turn increases the computational cost, and spoils the efficiency advantage. Additionally, ensemble methods rely most of the time on a linearity assumption, or alternatively linearization of the operators, which can result in a loss of accuracy for highly non-linear cases. Conversely, variational methods provide a general non-linear framework which is convenient for complex physical models, and do not rely on sampling. However, they are known to be time consuming for moderate control/state dimensions, oppositely to ensemble methods. This is caused by the classical use of iterative descent for minimization \citep{Asch2016}. As a result, both ensemble-based and variational techniques can suffer from computational burden, either caused by sampling size and inefficiency of Monte Carlo (MC) type estimation, or due to iterative descent which only gradually explores the parameters space. Consequently, methods to decrease the cost of DA are desired. \\

\begin{figure}[H]
  \centering
    \includegraphics[trim={0cm 0cm 0cm 0cm},clip,width=0.7\textwidth]{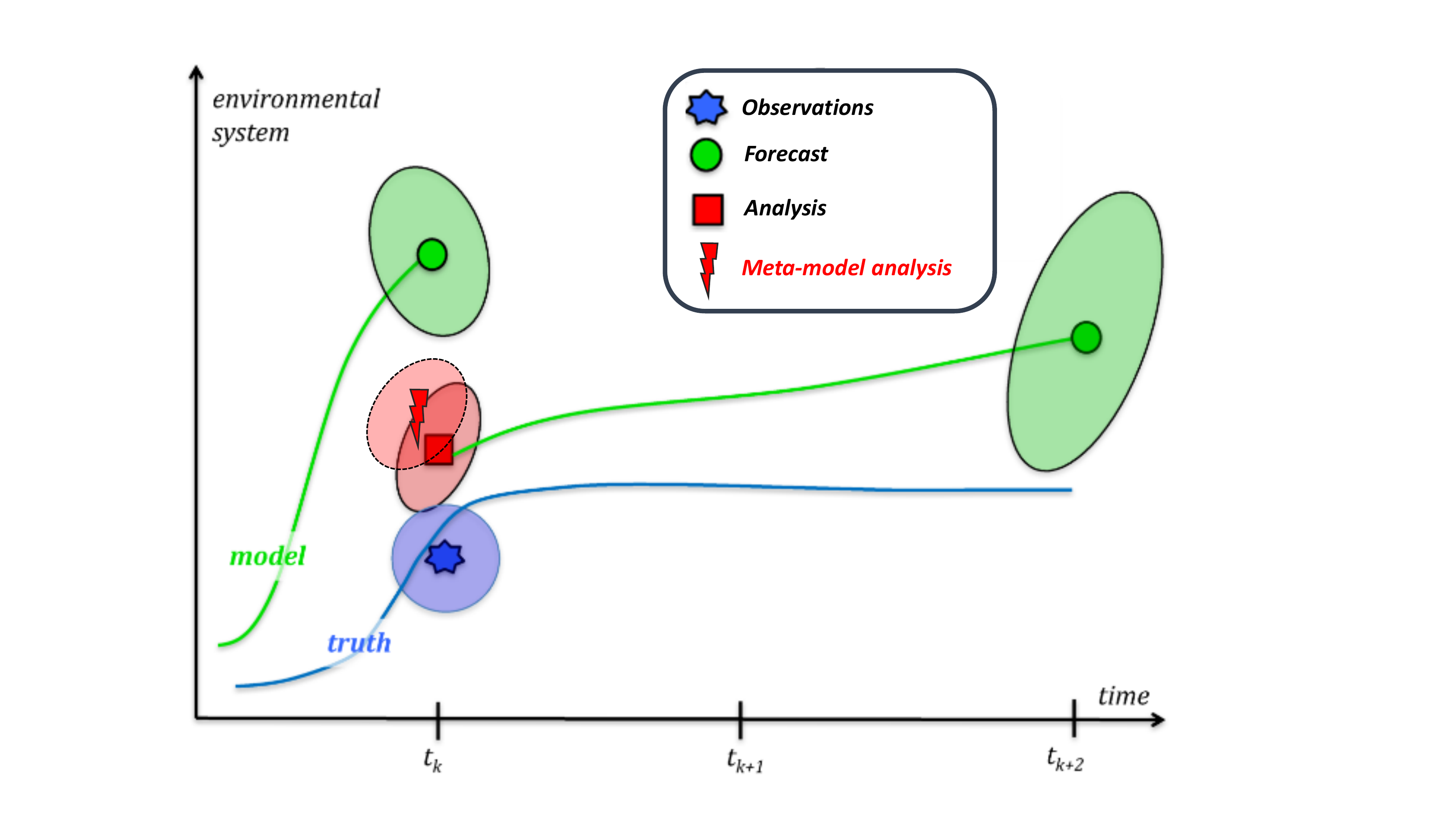}
    \caption{Representation of the analysis provided by a metamodel as a compromise between observations (in blue), a physics-based numerical model (in green) and their resepective uncertainties (in surrounding circles), adapted from \citep{Carrassi2018}}
    \label{fig:metamodelAnalysis}
\end{figure}

To overcome these limitations, the use of Dimensionality Reduction (DR) and input-to-output probabilistic modelling is investigated. Two approaches are proposed and compared: (i) an ensemble of parameters realizations and associated states are jointly reduced using Proper Orthogonal Decomposition (POD). Their covariances are represented through a common orthogonal basis. They are both replaced by their POD projections in a variational cost function. This is referred to as PODEn3DVAR (POD Ensemble-based three-Dimensional VARiational), which is a direct adaptation of PODEn4DVAR originally introduced by \citet{Tian2008},  to parametric cases; (ii) model states are first independently reduced using POD. Their variations are therefore represented by a low dimensional vector, whose components are learned as a function of control parameters using Polynomial Chaos Expansion (PCE) \citep{Lemaitre2001_a,Lemaitre2002_b}. This provides a coupled POD-PCE \textit{metamodel}, replacing the original model in the 3DVAR cost function, and hence called POD-PCE-3DVAR. The model replacement by a metamodel allows performing a DA analysis at lower cost, that in principle is an approximation of the analysis provided by the full model, as represented in Figure \ref{fig:metamodelAnalysis}. \\

POD application to increase efficiency of variational DA was given particular attention in recent literature contributions. For example, an intrusive Galerkin scheme is used to derive a reduced model and corresponding adjoint, based on orthogonal basis of model states in \citep{Park2001}. This was originally proposed for Boussinesq equations in \citep{Park2001}, and later applied to ocean modelling in \citep{Cao2007} and tidal equations in \citep{Qian2016}. Previous contributions are however case dependent due to the use of intrusive approaches and adjoint calculation. Conversely, \citet{Vermeulen2006} proposed an alternative based on linearization of model responses around a prior, combined to POD reduction and finite differences estimations of partial derivatives. This results with an approximate adjoint derivation, applicable to arbitrary models \citep{Vermeulen2006}. Their method was applied to Shallow Water Equations (SWE) for example in \citep{Altaf2009}, and coastal hydro-morphodynamics in \citep{Garcia2013}. To completely avoid adjoint calculation, \citet{Tian2008} propose the so-called PODEn4DVAR, to assimilate temporal states by calibrating a dynamic model's initial condition. POD is used to jointly reduce all spatio-temporal states including initial condition, which produces a common orthogonal basis to generate both the control and state spaces \citep{Tian2008}. Using twin experiments, authors show that PODEn4DVAR solution is better than both iterative 4DVAR and Ensemble-based Kalman Filter (EnKF) \citep{Evensen2009}). \citet{Mons2017} adapted PODEn4DVAR for parametric calibration by linearly spanning the parameters space using the model states POD basis. However, the relationship between parameters and states is not explicitly modelled. \\

The use of PCE as metamodel in DA is not novel. It has been applied by \citet{Marzouk2007}, within a Bayesian DA framework, also using a Galerkin scheme, to replace the model. This allows efficient approximation of posterior distribution compared to MC sampling. Their method is generalized for functional outputs (e.g. spatiotemporal fields) in \citep{Marzouk2009} using DR prior to PCE learning, in order to reduce computational cost of Markov Chain Monte Carlo (MCMC). In other examples, PCE, known to converge significantly faster than MC sampling, is used to improve accuracy of ensemble methods by generating larger ensembles, as in \citep{Li2009}. It is also used in the latter to obtain error covariance matrices involved in the filter, instead of calculating ensemble-anomaly matrices, by straightforward estimation of statistical moments due to the orthonormality of the basis \citep{Li2009}. Same strategy is adopted in \citep{Rochoux2014a} for forest fire front tracking application. However, none of the previous strategies used PCE in a purely variational framework for functional outputs.

Hereby, we therefore attempt to: (i) adapt PODEn4DVAR to a purely parametric framework, in such way to capture parameter-state covariances; (ii) apply PCE as metamodel in a variational framework, and more precisely for functional outputs by coupling to POD and (iii) account for the errors resulting from the use of a POD-PCE surrogate in the optimization process, by considering a new error covariance matrix, which considerably improves the variational estimates. The proposed methods present interesting characteristics for efficient DA. Firstly, joint parameter-state reduction in PODEn3DVAR enforces the model dynamics in the cost function, and avoids learning the non-linear relationships \citep{Tian2008}. The resulting cost function linearly depends on a unique reduced vector that represents parameters and states. Its minimum can be computed analytically, without iterative descent. In particular, POD has shown to be accurate for non-linear problems \citep{Taira2017}, although being a linear decomposition. Its use in DA shows "\textit{superiority [...] with regard to the other families of vectors}" \citep{Robert2005}, not only in terms of representation efficiency, but also due to straightforward calculation of error covariance matrices for the reduced vector \citep{Durbiano2001}. Indeed, POD results with an eigenvalue matrix, that is a reduced size estimate of covariances, diagonal even for highly correlated vectors. \citet{Robert2005} for example uses the eigenvalues as a background error covariance matrix, making the variational problem four to five times faster, and the error of DA solution smaller \citep{Robert2005}. In addition, \citet{Vermeulen2006} emphasis the interest of POD to avoid local minima in variational frameworks, with the smoothing inherent to the methods. Secondly, coupled use of POD and PCE benefits from the previously cited advantages of POD, in addition to the \textit{spectral decay} assured by PCE \citep{Sudret2008}. Indeed, the latter is based on orthogonal polynomials, implying learning efficiency and fast convergence of statistical moments, compared to MC type methods based on sampling. Furthermore, in a previous contribution \citep{Mouradi2021}, POD-PCE efficiency to learn complex and non-linear phenomena was demonstrated for point-wise prediction of multi-dimensional fields. \\

\begin{figure}[H]
  \centering
    \includegraphics[trim={0.5cm 0.9cm 0.5cm 3.2cm},clip,scale=0.52]{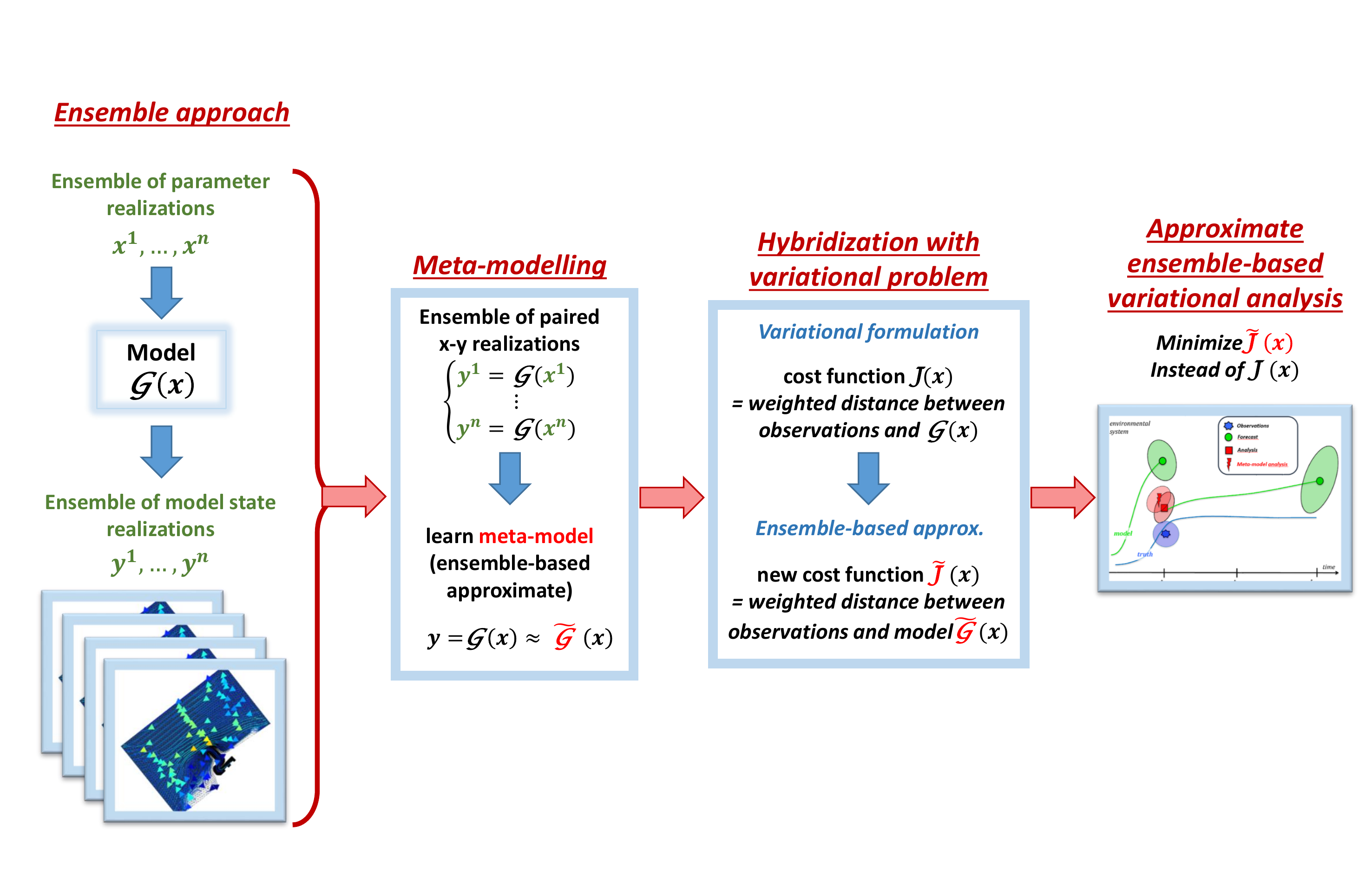}
    \caption{Representation of the hybrid ensemble-variational metamodelling approaches}
    \label{fig:ensembleVarDA}
\end{figure}

Both approaches, PODEn3DVAR (linear surrogate with joint parameter-state POD) and POD-PCE-3DVAR (non-linear surrogate by POD-PCE coupling), are based on hybridization between ensemble and variational methods. The variational cost function formulation is used, but the model is replaced with an ensemble-based approximate. This is performed by generating an ensemble of parameters and related model states that allow to build appropriate emulators, as represented in Figure \ref{fig:ensembleVarDA}.

Hybridization is used in an attempt to benefit from the advantages of both variational and ensemble approaches. While still relying on a cost function with a general non-linear framework, the computational burden of the descent is greatly reduced by the use of a surrogate instead of the full model. Additionally, while still using an ensemble of states, the sample size for convergence is reduced by means of POD that accurately summarizes the variances, especially when using POD-PCE-3DVAR with PCE that is much more efficient than MC sampling. In the presented study, performances of both approaches are investigated on a twin experiment as well as on measurements, applied to tidal modelling in a coastal area, by solving the Shallow Water Equations (SWE). Assessment of the robustness to different levels of noise is proposed on the twin experiment. In addition, sensitivity to the considered number of POD basis components is investigated. Hybrid ensemble-variational POD-PCE-3DVAR and PODEn3DVAR are confronted, and compared to classical 3DVAR on measurements. \\

The paper is outlined as follows: theoretical elements are introduced in Section \ref{section:materials}. Parametric 3DVAR framework is briefly presented in Subsection \ref{subsection:DA}, while proposed PODEn3DVar and POD-PCE-3DVAR are detailed in Subsections \ref{subsection:PODEn3DVAR} and \ref{subsection:PODPCE3DVAR} respectively, with a focus on POD-PCE metamodel error covariance calculation in Section  \ref{subsection:error}. An application case and benchmarks are presented in Section \ref{section:application}, and a discussion and conclusion are proposed in Section \ref{subsection:application:discussion}.



\shorthandoff{:}

\section{Materials}
\label{section:materials}

In this Section, two metamodelling approaches, based on an ensemble generation, are proposed to accelerate variational parametric calibration using 3DVAR. Firstly, classical 3DVAR framework is briefly reminded in Section \ref{subsection:DA}. Secondly, PODEn3DVAR, which consists in joint parameter-state reduction, is presented in Subsection \ref{subsection:PODEn3DVAR} based on Proper Orthogonal Decomposition (POD). Thirdly, POD-PCE-3DVAR is proposed in Subsection \ref{subsection:PODPCE3DVAR}, based on a coupling between POD and Polynomial Chaos Expansion (PCE). Lastly, an appropriate calculation of metamodelling error, in order to update the model error covariance matrix in the optimization process, is proposed for POD-PCE-3DVAR in Section \ref{subsection:error}.

\subsection{Data Assimilation: parametetric calibration using 3DVAR}
\label{subsection:DA}

Data Assimilation (DA) is a sub-class of inverse problems, where the objective is to estimate an unknown (state, parameters, etc.) using a compromise between observations and a model outputs, with a background idea (or prior knowledge) about the unknown \citep{Asch2016,Blayo2011}. The most widely used methods are ensemble methods \citep{Evensen2009} and 3D- or 4DVAR (three- or four-Dimensional VARiational) \citep{Asch2016}. Particular attention is here given to hybrid approaches, where variational formulation is used, but the model is replaced with a ensemble-based surrogate. Brief explanation of the framework is given below, and reader interested in details can refer to \citep{Asch2016,Evensen2009}. \\

This study focuses on \textit{parametric estimation} problems, where the objective is to inversely deduce a set of parameters denoted $\mathbf{x}$ from optimal calibration of an interest state denoted $\mathbf{y} \coloneqq \mathcal{H} \circ \mathcal{M} (\mathbf{x}) \coloneqq \mathcal{G} (\mathbf{x}) $, where $\mathcal{M}$ designates a physics based numerical model, and $\mathcal{H}$ an observation operator (projection, interpolation, selection, transformation, etc.), combined through a general model denoted $\mathcal{G} \coloneqq \mathcal{H} \circ \mathcal{M}$ \citep{Carrassi2018}. Attributes are given to variables as superscripts: true state $\mathbf{y}^{(t)}$ and associated true parameters $\mathbf{x}^{(t)}$ designate a perfect model calibration. They are in general unknown, and the modeller, in principle, has an uncertain idea about parameters values, denoted $\mathbf{x}^{(b)}$. The objective is to find the best estimate called analysis and denoted $\mathbf{x}^{(a)}$, that provides an equilibrium between observed state values denoted $\mathbf{y}^{(o)}$ and the background $\mathbf{y}^{(b)} = \mathcal{G}(\mathbf{x}^{(b)})$, taking into consideration their respective uncertainties. In variational formulation, finding the analysis consists in minimizing a cost function denoted $J(\mathbf{x})$. \\

In a discrete (finite) multidimensional setting, where the parameters, state, and observations are arranged in vectors $\mathbf{x} = [x_1, \hdots, x_{m_x}] \in \mathbb{R}^{m_x}$, $\mathbf{y}=[y_1,\dots, y_{m_y}]^T \in \mathbb{R}^{m_y}$ and $\mathbf{y}^{(o)}=[y^{(o)}_1,\dots,y^{(o)}_{m_y}]^T \in \mathbb{R}^{m_y}$ respectively (typically, uncertain parameters, multivariate response of a numerical model and spatio-temporal measures), the cost function to minimize \citep{Asch2016} can be written as in Equation \ref{eq:materials:DA:3DVARcostFunction},
\begin{equation}
\label{eq:materials:DA:3DVARcostFunction}
J(\mathbf{x})  =  \frac{1}{2} \left\| \mathbf{x} -  \mathbf{x}^{(b)}  \right\|^2_{\mathbf{B}^{-1}} + \frac{1}{2}\left\|\mathcal{G}(\mathbf{x}) - \mathbf{y}^{(o)}\right\|^2_{\mathbf{R}^{-1}} \ ,
\end{equation}
where $||\mathbf{v}-\mathbf{w}||_{\mathbf{A}} \coloneqq (\mathbf{v}-\mathbf{w})^T\mathbf{A}(\mathbf{v}-\mathbf{w})$ is the weighted Mahalanobis distance, and matrices $\mathbf{B} \in \mathbb{R}^{m_x\times m_x}$ and $\mathbf{R} \in \mathbb{R}^{m_y\times m_y}$ are the symmetrical covariance matrices of the background and state/observation error respectively. Indeed, the background on the one hand, and the model, observation and its operator on the other hand, are both characterized with errors, denoted $\boldsymbol{\epsilon}^{(b)}$ and $\boldsymbol{\epsilon}^{(m,o)}$ respectively, as expressed in Equation \ref{eq:materials:DA:backgroundObservationError}. 
\begin{equation}
\label{eq:materials:DA:backgroundObservationError}
\left\{  \begin{matrix}
\mathbf{x}^{(t)}& = & \mathbf{x}^{(b)} + \boldsymbol{\epsilon}^{(b)} \\
& & \\
\mathbf{y}^{(o)} & = & \mathcal{G}\left(\mathbf{x}^{(t)}\right) + \boldsymbol{\epsilon}^{(m,o)}
\end{matrix} \right. \ .
\end{equation}
In a variational DA framework, errors are considered Gaussian and unbiased (zero expectations), strictly defined by their covariance matrices $\mathbf{B} \coloneqq [cov(\epsilon^{(b)}_i,\epsilon^{(b)}_j)]_{i,j \in \{1,\hdots,m_x\}}$ and $\mathbf{R} \coloneqq [cov(\epsilon^{(m,o)}_i,\epsilon^{(m,o)}_j)]_{i,j \in  \{1,\hdots,m_y\}}$. All these errors should be minimized, which sometimes results in a high dimensional problem, principally due to the dimension of model response $\mathbf{y}$. Additionally, model errors denoted $\boldsymbol{\epsilon}^{(m)}$ are in general difficult to define. A \textit{perfect model} hypothesis is therefore added (i.e. $\boldsymbol{\epsilon}^{(m)}=0$ and $\boldsymbol{\epsilon}^{(m,o)}=\boldsymbol{\epsilon}^{(o)}$), and resulting problem is said \textit{strong-constraint} \cite{Carrassi2018}. \\

A minimum corresponds to a null gradient of $J$. In the general case of non-linear models, classical approach consists in identifying an \textit{adjoint model} \citep{Asch2016}, whose solution is equivalent to cancelling the gradient \citep{Asch2016}. This method is however intrusive and presents drawbacks: it is conditioned by analytical derivation of the adjoint model, not possible for all systems of equations, and any update of the direct model and its discretization comes with the supplementary cost of updating the adjoint, which is not always trivial and requires resources. Non-intrusive solutions are therefore often preferred. For example, iterative descent methods, where the gradient is approximated using finite-differences, can be used \citep{Asch2016}. Due to the numerous approximations (cost function definition, error covariance matrices approximation, descent algorithms), optimal solution is only an approximation (analysis) of the true state (perfect knowledge). It is therefore also characterized with errors, which can be expressed as $\mathbf{x}^{(a)}=\mathbf{x}^{(t)} + \boldsymbol{\epsilon}^{(a)}$. An associated error covariance matrix is denoted $\mathbf{P}^{(a)} \coloneqq [cov(\epsilon^{(a)}_i,\epsilon^{(a)}_j)]_{i,j \in \{1,\hdots,m_x\}}$. \\

Minimizing the cost function $J$ can be costly, principally due to the cost of the model $\mathcal{G}$. Hence, in the following Sections \ref{subsection:PODEn3DVAR} and \ref{subsection:PODPCE3DVAR}, methods to construct a linear and a non-linear approximate $\widetilde{\mathcal{G}}$ for the model $\mathcal{G}$ are respectively proposed. This results with metamodelling errors that should be accounted for in the error covariance matrix $\mathbf{R}$, which is proposed in Section \ref{subsection:error} for POD-PCE-3DVAR.

\subsection{Linear Surrogate: PODEn3DVAR}
\label{subsection:PODEn3DVAR}

This section is dedicated to the use of Proper Orthogonal Decomposition (POD) as an ensemble-based linear surrogate within 3DVAR parametric calibration. POD is briefly presented in Subsection \ref{subsubsection:PODEn3DVAR:POD}, while its use to construct a linear metamodel is detailed in \ref{subsubsection:PODEn3DVAR:PODEn3DVAR}. It is also used as a reduction method prior to the construction of a non-linear metamodel in Section \ref{subsection:PODPCE3DVAR}. Readers interested in supplementary theoretical details about POD and demonstrations can refer to \citep{Sirovich1987,Muller2008_phd}.  

\subsubsection{Proper Orthogonal Decomposition}
\label{subsubsection:PODEn3DVAR:POD}
POD is a Dimensionality Reduction (DR) technique \citep{Lumley1967}, consisting in a linear variable separation for continuous bi-variate functions denoted $\mathbf{u}(\psi,\omega): \Psi \times \Omega \rightarrow \mathbb{D}$, where  $\mathbb{D}$ is a Hilbert space characterized by its scalar product $(.~,.)_{\mathbb{D}}$ and induced norm $||.||_{\mathbb{D}}$. This is written as in Equation \ref{eq:materials:POD:expansion},
\begin{equation}
\label{eq:materials:POD:expansion}
u(\psi,\omega) = \sum_{k=1}^{\infty} \nu_k(\omega) \sqrt{\lambda_k} \boldsymbol{\phi}_k(\psi) \ ,
\end{equation}
where $\{\lambda_k\}_{k=1}^{\infty} \subset \mathbb{R}$, $\{\nu_k(.)\}_{k=1}^{\infty} \subset \mathcal{C}(\Omega ,\mathbb{R})$ and $\{\boldsymbol{\phi}_k(.)\}_{k=1}^{\infty}  \subset \mathcal{C}(\Psi,\mathbb{D})$, with $\mathcal{C}(\mathbb{A},\mathbb{B})$ denoting the space of continuous functions defined over $\mathbb{A}$ and arriving at $\mathbb{B}$. In particular, $\{\boldsymbol{\phi}_k(.)\}_{k=1}^{\infty}$ is called POD basis and is orthonormal with respect to $(.~,.)_{\mathbb{D}}$. Its members are ordered according to their importance in the interest field representation in terms of variance, which can be calculated as in Equation \ref{eq:POD:EVR}, where $e_d$ is called Explained Variance Rate (EVR). It quantifies the proportion of variance captured by POD at a given rank $d \in \mathbb{N}$, with an approximation $\sum_{k=1}^{d} \nu_k(\omega) \sigma_k \boldsymbol{\phi}_k(\psi)$. When an order $d<<min(dim(\Psi),dim(\Omega))$  corresponds to a high EVR, we speak of DR, because $u$ is spanned to a sub-space of much smaller dimension than $\Psi \times \Omega$.
\begin{equation}
\label{eq:POD:EVR}
e_d=\dfrac{ \sum_{k \leq d} \lambda_k}{ \sum_{k=1}^{+\infty} \lambda_k} \ .
\end{equation}

At given rank $d$, POD approximation has the lowest error (in terms of the $||.||_{\mathbb{D}}$ norm and averaged over $\Omega$) compared to any other linear expansion \citep{Lumley1967}. \\

In discrete form, POD is usually written as in Equation \ref{eq:materials:POD:SVD}, for a real valued physical variable measured at different coordinates $\{\psi_1,\dots,\psi_m\}$ (e.g. spatio-temporal locations)  and for an \textit{ensemble} of events $\{\omega_1,\dots,\omega_m\}$ (e.g. realizations in different configurations), and stored in matrix $\mathbf{U} \coloneqq [u(\psi_i,\omega_j)]_{i,j} \in \mathbb{R}^{m \times n}$,
\begin{equation}
\label{eq:materials:POD:SVD}
\mathbf{U} =  \overline{\mathbf{U}} + \boldsymbol{\Phi} \boldsymbol{\Lambda}^{1/2}\mathbf{N}^T  = \overline{\mathbf{U}} + \boldsymbol{\Phi} \boldsymbol{\Sigma}\mathbf{N}^T ~ ,
\end{equation}
where $\boldsymbol{\Phi} \coloneqq [\phi_k(\psi_i)]_{i,k} \in \mathbb{R}^{m \times e}$,  $\mathbf{N} \coloneqq [\nu_k(\omega_j)]_{j,k} \in \mathbb{R}^{n \times e}$, and $\boldsymbol{\Lambda} \coloneqq [\lambda_k]_{k,k} \in \mathbb{R}^{e \times e}$, with $e=min(m,n)$, and $\overline{\mathbf{U}}$ the ensemble mean. Matrices $\mathbf{\Phi}$ and $\mathbf{N}$ are orthonormal, so that $\boldsymbol{\Phi}\boldsymbol{\Phi}^T=\mathbf{I}_m$ and $\mathbf{N} \mathbf{N}^T= \mathbf{I}_n $, where $\mathbf{I}_m$ is the $m \times m$ identity matrix. The POD basis $\boldsymbol{\Phi}$ corresponds to the eigenvectors of matrix $(\mathbf{U}-\overline{\mathbf{U}})(\mathbf{U}-\overline{\mathbf{U}})^T$ decomposed as $\boldsymbol{\Phi} \boldsymbol{\Lambda} \boldsymbol{\Phi}^T$, with $\boldsymbol{\Lambda}$ the eigenvalues arranged in decreasing order, or using the equivalent Singular Value Decomposition (SVD) as $(\mathbf{U} -  \overline{\mathbf{U}}) = \boldsymbol{\Phi} \boldsymbol{\Sigma}\mathbf{N}^T$, where $\boldsymbol{\Sigma} = \boldsymbol{\Lambda}^{1/2}$ is the singular values matrix. Matrix $\mathbf{N}^T$ is deduced using a projection $ \mathbf{N}^T = \boldsymbol{\Sigma}^{-1} \boldsymbol{\Phi}^T \left(\mathbf{U} - \overline{\mathbf{U}}\right)$. Each column $j$ of $\mathbf{N}^T$ is a vector of expansion coefficients that represent a realization $\omega_j$ of the decomposed field. When POD approximation is truncated at rank $d$, it is written as $\mathbf{U} \approx  \overline{\mathbf{U}} + \boldsymbol{\Phi}^{(d)} \boldsymbol{\Sigma}^{(d)}\mathbf{N}^{(d)}$, where column $j$ of $\mathbf{N}^{(d)}$ contains reduced vector variable $\boldsymbol{\nu}^{(d)} = [\nu_1(\omega_j), \hdots, \nu_d(\omega_j)]^T \in \mathbb{R}^{d}$. \\

\subsubsection{PODEn4DVAR adapted to parametric 3DVAR}
\label{subsubsection:PODEn3DVAR:PODEn3DVAR}
When diverse enough records $\{\omega_1,\dots,\omega_m\}$ are available, resulting POD basis can be considered as a generator of all possible states. Then, any state $u(\psi,\omega_j)$ can be approximated on the same basis $\boldsymbol{\Phi}^{(d)}$ by fitting $\boldsymbol{\nu}^{(d)}=[\nu_1(\omega_j), \hdots \nu_d(\omega_j)]^T$. Hence, \citet{Tian2008} proposes to replace the model response at any time in a 4DVAR cost function by its POD approximation, which they call PODEn4DVAR. Consideration of the model dynamics is then enforced using POD rather than an adjoint. \\

We propose the same procedure using a joint parameter-state POD and cost function in Equation \ref{eq:materials:DA:3DVARcostFunction}.  A number $n$ of paired realizations of parameters $[x_1, \hdots, x_{m_x}]$ and associated states $[y_1, \hdots, y_{m_y}]$ (through model $\mathcal{G}$) are arranged in a single vector $[x_1, \hdots, x_{m_x}, y_1, \hdots, y_{m_y}]$ of size $m_x + m_y$ and stored in an ensemble matrix of size $ (m_x + m_y)\times n$. The latter is POD reduced which gives an approximation in the form of Equation \ref{eq:materials:PODEn3DVAR:jointParamStatePOD} for each realization. The POD basis matrix can therefore be written as a block vector in the form $\boldsymbol{\Phi}^{(d)} = [\boldsymbol{\Phi}^{(d)}_{\mathbf{x}}, \boldsymbol{\Phi}^{(d)}_{\mathbf{y}}]^T$, where parameter block denoted $\boldsymbol{\Phi}^{(d)}_{\mathbf{x}}$ is of size $m_x \times d$, and state block denoted $\boldsymbol{\Phi}^{(d)}_{\mathbf{y}}$ is of size $m_y \times d$, formed simply by selecting adequate lines of the full matrix $\boldsymbol{\Phi}^{(d)}$. 
\begin{equation}
\label{eq:materials:PODEn3DVAR:jointParamStatePOD}
\left[x_1, \hdots,x_{m_x}, y_1, \hdots, y_{m_y}\right]^T \approx \left[~\overline{x}_1, \hdots, \overline{x}_{m_x}, \overline{y}_1,\hdots, \overline{y}_{m_y}\right]^T +\boldsymbol{\Phi}^{(d)} \boldsymbol{\Sigma}^{(d)}\boldsymbol{\nu}^{(d)} ~ .
\end{equation}

This provides a common linear generator for parameters and state. Hence, they both can be replaced in the parametric 3DVAR cost function in Equation \ref{eq:materials:DA:3DVARcostFunction}. A new approximate cost function is written as in Equation \ref{eq:materials:PODEn3DVAR:costFunction} (where superscript $d$ for $\boldsymbol{\nu}^{(d)}$ is dropped for the simplicity of following formulas). 
\begin{equation}
\label{eq:materials:PODEn3DVAR:costFunction}
J(\mathbf{x})  \approx \widetilde{J} (\boldsymbol{\nu}) \coloneqq \frac{1}{2} \left\| \boldsymbol{\Phi}_{\mathbf{x}}^{(d)} \boldsymbol{\Sigma}^{(d)} \left(\boldsymbol{\nu} -  \boldsymbol{\nu}^{(b)} \right) \right\|^2_{\mathbf{B}^{-1}} + \frac{1}{2}\left\|\overline{\mathbf{y}} + \boldsymbol{\Phi}_{\mathbf{y}}^{(d)} \boldsymbol{\Sigma}^{(d)}\boldsymbol{\nu} - \mathbf{y}^{(o)}\right\|^2_{\mathbf{R}^{-1}} \ ,
\end{equation}

It can be in particular noted that the linear metamodel reads:
\begin{equation}
\label{eq:materials:PODEn3DVAR:metamodely}
 \mathcal{G} \approx \widetilde{\mathcal{G}} \coloneqq \overline{\mathbf{y}} + \boldsymbol{\Phi}_{\mathbf{y}}^{(d)} \boldsymbol{\Sigma}^{(d)}\boldsymbol{\nu}  \ .
\end{equation}

The problem becomes linear and explicit, and the gradient can be analytically calculated as in Equation \ref{eq:materials:PODEn3DVAR:CostFunctionGradient}.
\begin{equation}
\label{eq:materials:PODEn3DVAR:CostFunctionGradient}
\begin{matrix}
\nabla_{\boldsymbol{\nu}} \widetilde{J}  & = & \left(\boldsymbol{\Phi}_{\mathbf{x}}^{(d)} \boldsymbol{\Sigma}^{(d)}\right)^T\mathbf{B}^{-1}\left(\boldsymbol{\Phi}_{\mathbf{x}}^{(d)} \boldsymbol{\Sigma}^{(d)}\right)\left( \boldsymbol{\nu} - \boldsymbol{\nu}^{(b)} \right) \\
& + & \left(\boldsymbol{\Phi}_{\mathbf{y}}^{(d)} \boldsymbol{\Sigma}^{(d)}\right)^T \mathbf{R}^{-1} \left(\overline{\mathbf{y}} + \boldsymbol{\Phi}_{\mathbf{y}}^{(d)} \boldsymbol{\Sigma}^{(d)}\boldsymbol{\nu}  -  \mathbf{y}^{(o)} \right) \end{matrix} \ .
\end{equation}

The analysis is obtained by cancelling the gradient, as in Equation \ref{eq:materials:PODEn3DVAR:analysis}.
\begin{equation}
\label{eq:materials:PODEn3DVAR:analysis}
\begin{matrix}
\boldsymbol{\nu}^{(a)}  & = &  \left[\left(\boldsymbol{\Phi}_{\mathbf{x}}^{(d)} \boldsymbol{\Sigma}^{(d)}\right)^T\mathbf{B}^{-1}\left(\boldsymbol{\Phi}_{\mathbf{x}}^{(d)} \boldsymbol{\Sigma}^{(d)}\right) + \left(\boldsymbol{\Phi}_{\mathbf{y}}^{(d)} \boldsymbol{\Sigma}^{(d)}\right)^T \mathbf{R}^{-1} \left(\boldsymbol{\Phi}_{\mathbf{y}}^{(d)} \boldsymbol{\Sigma}^{(d)}\right)  \right]^{-1} \\
& & \\
& \times & \left[ \left(\boldsymbol{\Phi}_{\mathbf{x}}^{(d)} \boldsymbol{\Sigma}^{(d)}\right)^T\mathbf{B}^{-1}\left(\boldsymbol{\Phi}_{\mathbf{x}}^{(d)} \boldsymbol{\Sigma}^{(d)}\right)\boldsymbol{\nu}^{(b)} -
\left(\boldsymbol{\Phi}_{\mathbf{y}}^{(d)} \boldsymbol{\Sigma}^{(d)}\right)^T \mathbf{R}^{-1} \left(\overline{\mathbf{y}} - \mathbf{y}^{(o)}\right) \right]
\end{matrix} \ .
\end{equation}

Solution to the minimization is hence straightforward and does not require an iterative algorithm. 

\subsection{Non-linear probabilistic surrogate: POD-PCE-3DVAR}
\label{subsection:PODPCE3DVAR}

Previously presented metamodel in Section \ref{subsection:PODEn3DVAR} relies on a linearization of the relationships between interest model states and control parameters, which can be limiting for highly non-linear cases. Hence, in this section, a non-linear metamodel, based on POD-PCE coupling, is proposed and results with the POD-PCE-3DVAR methodology presented in Subsection \ref{subsubsection:PODPCE3DVAR:DA}. This metamodel relies on POD, as introduced in Section \ref{subsubsection:PODEn3DVAR:POD}, but instead of relating the parameters and model response by a joint reduction, their dependency is modelled using a probabilistic non linear mapping called Polynomial Chaos Expansion (PCE). Theoretical elements for the latter are therefore briefly presented below, namely the probabilistic framework in Subsection \ref{subsubsection:PODPCE3DVAR:probabilistic} and PCE in Subsection \ref{subsubsection:PODPCE3DVAR:PCE}. Readers interested in details can refer for instance to \citep{Sudret2008,XiuKarniadakis2003_flow}. \\

\subsubsection{Probabilistic framework}
\label{subsubsection:PODPCE3DVAR:probabilistic}

For the following calculations, we define by $(\Omega,F,\mathbb{P})$ a probability space, where $\Omega$ is the event space (space of all the possible events $\omega$) equipped with $\sigma$-algebra $F$ (some events of $\Omega$) and its probability measure $\mathbb{P}$ (likelihood of a given event occurrence). A random variable defines an application $Y(\omega): \Omega \rightarrow D_Y \subseteq \mathbb{R}$, with realizations denoted by $y \in D_Y$. The PDF of $Y$ is a function $f_Y: D_Y \rightarrow \mathbb{R}$ that verifies $\mathbb{P}(Y \in E \subseteq D_Y) = \int_E f_Y(y) dy$. \\

The \textit{$k^{th}$ moments} of $Y$ are defined as $\mathbb{E}[Y^k] \coloneqq \int_{D_Y} y^kf_Y(y)dy$, the first being the expectation denoted $\mathbb{E}[Y]$. In the same manner, we define the \textit{$k^{th}$ central moments} of $Y$ as $\mathbb{E}[(Y-\mathbb{E}[Y])^k]$, the first being $0$ and the second the variance of $Y$ denoted by $\mathbb{V}[Y]$.  The covariance of two random variables is defined as $cov(X,Y)=\mathbb{E}[(X-\mathbb{E}[X])(Y-\mathbb{E}[Y])]$ and a resulting property is $\mathbb{V}[Y] = cov(Y,Y)$. For a multi-dimensional random denoted $\mathbf{Y}=[Y_1, \hdots, Y_m]^T$, the expectation is defined component-wise as $\mathbb{E}[\mathbf{Y}]=[\mathbb{E}[Y_1], \hdots, \mathbb{E}[Y_m]]^T$, and the variance is defined as $\mathbb{V}[\mathbf{Y}]=\mathbb{E}[(\mathbf{Y}-\mathbb{E}[\mathbf{Y}])(\mathbf{Y}-\mathbb{E}[\mathbf{Y}])^T]$. It is a matrix of size $m\times m$ called the \textit{covariance matrix}, and can also be developed as $\mathbb{V}[\mathbf{Y}] =\mathbb{E}[\mathbf{Y}\mathbf{Y}^T] - \mathbb{E}[\mathbf{Y}]\mathbb{E}[\mathbf{Y}]^T$. Component of line $i_1$ and column $i_2$ is exactly the covariance term $cov(Y_{i_1},Y_{i_2})$ and hence, diagonal terms $i$ correspond to $\mathbb{V}[Y_i]$ and the matrix is diagonal.\\

In this probabilistic framework, parameters and model response are considered to belong to the space of  random variables with finite variances, denoted $\mathcal{L}^2_{\mathbb{R}}$. They are referred to using capital letters $\mathbf{X}$ and $\mathbf{Y}$ respectively. A realization of these randoms for an event $\omega \in \Omega$, is denoted using lower case letters as $\mathbf{X}(\omega) \coloneqq \mathbf{x} = [x_1, \hdots, x_{m_x}]^T$ and $\mathbf{Y}(\omega) = \mathcal{G}(\mathbf{X}(\omega)) \coloneqq \mathbf{y} = [y_1, \hdots, y_{m_y}]^T$. \\

The space of real random variables with finite variances $\mathcal{L}^2_{\mathbb{R}}$ is a Hilbert equipped with an inner product $(Y,Z)_{\mathcal{L}^2_{\mathbb{R}}} \coloneqq \mathbb{E}[YZ] = \int_{\Omega}Y(\omega)Z(\omega)d\mathbb{P}(\omega)=\int_{\Omega} y z f_{Y,Z}(y,z) d\omega$ and its induced norm $||Y||_{\mathcal{L}^2_{\mathbb{R}}} \coloneqq \sqrt{\mathbb{E}[Y^2]}$. In particular,  $f_{Y,Z}$ defines the joint PDF of randoms $Y$ and $Z$, defined as  $\mathbb{P}(Y, Z \in E \subseteq D_Y \times D_Z) = \int_E f_{Y,Z}(y,z) dydz$. \\

A number $n$ of parameters realizations are used to produce $n$ realizations of the model response. Model responses are then stored in an ensemble matrix of size $m_y \times n$, and POD reduced as in Equation \ref{eq:materials:PODPCEDA:modelStatePOD}.
\begin{equation}
\label{eq:materials:PODPCEDA:modelStatePOD}
\mathbf{Y}(\omega) \approx \overline{\mathbf{Y}} + \boldsymbol{\Phi}^{(d)} \boldsymbol{\Sigma}^{(d)}\boldsymbol{\Upsilon}^{(d)}(\omega) ~ .
\end{equation}

where $\boldsymbol{\Upsilon}^{(d)}(\omega) \coloneqq \boldsymbol{\nu}^{(d)}$ is a realization of reduced variable associated to a realization of state $\mathbf{Y}(\omega)$. 
PCE is then used to formulate a non-linear model that links random reduced variable $\boldsymbol{\Upsilon}^{(d)}=[\Upsilon_1, \hdots, \Upsilon_d]^T$ to random parameters $\mathbf{X}$. 

\subsubsection{Polynomial Chaos Expansion}
\label{subsubsection:PODPCE3DVAR:PCE}
For a component $\Upsilon_k$ or random variable $\boldsymbol{\Upsilon}^{(d)}$, PCE approximation $\widetilde{\Upsilon}_k(\mathbf{X})$ is written in Equation \ref{eq:materials:PCE:polynomial},
\begin{equation}
  \label{eq:materials:PCE:polynomial}
\Upsilon_k \approx \widetilde{\Upsilon}_k(\mathbf{X}) \coloneqq \sum_{|\boldsymbol{\alpha}| \leq p} c_{\boldsymbol{\alpha}}^k \zeta_{\mathbf{X},\boldsymbol{\alpha}}(X_1, X_2, ..., X_{m_x}) \ ,
\end{equation}
where $c^k_{\boldsymbol{\alpha}} \in \mathbb{R}$ are deterministic coefficients, and $\left\{\zeta_{\mathbf{X},\boldsymbol{\alpha}}, \boldsymbol{\alpha} \coloneqq (\alpha_1, \hdots, \alpha_{m_x}) \in \mathbb{N}^{m_x}, |\boldsymbol{\alpha}| \coloneqq \sum_{i=1}^{m_x} \alpha_i \in [|0,p|]\right\}$ is an orthormal multivariate polynomial basis of maximum degree $p\in \mathbb{N}$. The orthonormality is defined with respect to the inner product $(.,~.)_{\mathcal{L}^2_{\mathbb{R}}}$. The multivariate basis is constructed for mutually independent parameters as $\zeta_{\mathbf{X},\boldsymbol{\alpha}} (\mathbf{X}) \coloneqq \prod_{i=1}^{m_x}\xi_{X_i,\alpha_i}(X_i)$, where $\left\{\xi_{X_i,\beta}, \beta \in [|0,p|] \right\}$ is an orthonormal univariate polynomial basis for each input variable $X_i$. This is written as $(\xi_{X_i,\beta_j},\xi_{X_i,\beta_k})_{\mathcal{L}^2_{\mathbb{R}}}=\int_{\Omega} \xi_{X_i,\beta_j}(x_i) \xi_{X_i,\beta_k}(x_i) f_{X_i}(x_i) d\omega = \delta_{\beta_j\beta_k}$, where $\delta_{\beta_j\beta_k}$ is the Kronecker symbol. The constructed multivariate basis is also orthonormal with respect to joint probability distribution of parameters denoted $f_{X_1, \hdots, X_{m_x}}$.  \\ 

The choice of the basis is therefore directly related to the choice of input variable marginals. It has been for example shown that Hermite polynomials are orthonormal with respect to Gaussian distributions, whereas Legendre polynomials are orthonormal with respect to Uniform densities \citep{XiuKarniadakis2002}.  \\

Coefficients $c_{\boldsymbol{\alpha}}^k$ can be estimated thanks to different methods, and is here performed using the Least Angle Regression Stagewise method (LARS) in order to construct an adaptive sparse PCE. This is an iterative procedure, where the algorithm begins by finding the polynomial pattern, denoted $\zeta_i$ for simplicity, that is the most correlated to the output. The latter is linearly approximated by $\epsilon_i\zeta_i$, where $\epsilon_i \in \mathbb{R}$. Coefficient $\epsilon_i$ is not set to its maximal value, but increased starting from 0, until another pattern $\zeta_j$ is found to be as correlated to $Y - \epsilon_i\zeta_i$, and so on. In this approach, a collection of possible PCE, ordered by sparsity, is provided and an optimum can be chosen with an accuracy estimate. It was performed in this study using corrected leave-one-out error. The reader can refer to the work of \citet{Blatman2011} for further details on LARS and more generally on sparse constructions.  \\

For multi-dimensional variables $\boldsymbol{\Upsilon}^{(d)}=[\Upsilon_1, \hdots, \Upsilon_d]$, PCE constructed component wise with the same inputs $\mathbf{X}$ can be written as in Equation \ref{eq:materials:PCE:MultidimensionalPolynomial}, where $\boldsymbol{\zeta}_{\mathbf{X}}$ is a vector containing the basis elements, and each line of matrix $\mathbf{C}$ contains expansion coefficients $c^k_{\boldsymbol{\alpha}}$ of component $\Upsilon_k$. 
\begin{equation}
  \label{eq:materials:PCE:MultidimensionalPolynomial}
\boldsymbol{\Upsilon}^{(d)} \approx \widetilde{\boldsymbol{\Upsilon}}^{(d)}(\mathbf{X}) = \mathbf{C} \boldsymbol{\zeta}_{\mathbf{X}}
\end{equation}

The orthonormality is a particularly convenient property for an efficient representation. It guarantees \textit{spectral decay} and therefore fast convergence of the approximation, as well as direct estimation of statistical moments \citep{Sudret2008}.

\subsubsection{Approximate cost function for the parametric case}
\label{subsubsection:PODPCE3DVAR:DA}

The model $\mathcal{G}$ in the parametric 3DVAR cost function \ref{eq:materials:DA:3DVARcostFunction} is now replaced by an approximate probabilistic POD-PCE metamodel denoted $\widetilde{\mathcal{G}}$ as in Equation \ref{eq:materials:PODPCEDA:metamodel} (here, $\mathbf{y}$ is written for given realization in lower case letters, with empirical estimate of the mean denoted $\overline{\mathbf{y}}$),

\begin{equation}
\label{eq:materials:PODPCEDA:metamodel}
\mathbf{y} = \mathcal{G}(\mathbf{x})  \approx \widetilde{\mathcal{G}}(\mathbf{x}) \coloneqq \overline{\mathbf{y}} +\boldsymbol{\Phi}^{(d)} \boldsymbol{\Sigma}^{(d)}\widetilde{\boldsymbol{\nu}}^{(d)}(\mathbf{x}) ~ .
\end{equation}
where $\widetilde{\boldsymbol{\nu}}^{(d)}$ corresponds to a realization using PCE models for the reduced POD coefficients, defined as in Equation \ref{eq:materials:PCE:MultidimensionalPolynomial}. A new approximate cost function is then written (for a given realization) as in Equation \ref{eq:materials:PODPCEDA:metamodelCostFunction},
\begin{equation}
\label{eq:materials:PODPCEDA:metamodelCostFunction}
J(\mathbf{x}) \approx \widetilde{J}(\mathbf{x})  =  \frac{1}{2} \left\| \mathbf{x} -  \mathbf{x}^{(b)}  \right\|^2_{\mathbf{B}^{-1}} + \frac{1}{2}\left\|\overline{\mathbf{y}} +\boldsymbol{\Phi}^{(d)} \boldsymbol{\Sigma}^{(d)}\widetilde{\boldsymbol{\nu}}^{(d)}(\mathbf{x}) - \mathbf{y}^{(o)}\right\|^2_{\mathbf{R}^{-1}} \ ,
\end{equation}

and its gradient is written,  using the matricial multivariate PCE form of Equation \ref{eq:materials:PCE:MultidimensionalPolynomial}, as in Equation \ref{eq:materials:PODPCEDA:CostFunctionGradient}.
\begin{equation}
\label{eq:materials:PODPCEDA:CostFunctionGradient}
\nabla_{\mathbf{x}} \widetilde{J}   =  \mathbf{B}^{-1}\left( \mathbf{x} - \mathbf{x}^{(b)} \right) +  \left(\boldsymbol{\Phi}^{(d)} \boldsymbol{\Sigma}^{(d)}  \mathbf{C} \nabla_{\mathbf{x}} \boldsymbol{\zeta}_{\mathbf{X}} \right)^T \mathbf{R}^{-1} \left(\overline{\mathbf{y}} + \boldsymbol{\Phi}^{(d)} \boldsymbol{\Sigma}^{(d)}\widetilde{\boldsymbol{\nu}}^{(d)}(\mathbf{x})  -  \mathbf{y}^{(o)} \right) \ ,
\end{equation}

where $\nabla_{\mathbf{x}} \boldsymbol{\zeta}_{\mathbf{X}}$ is the jacobian matrix of PCE basis elements $\boldsymbol{\zeta}_{\mathbf{X}}$ that can be calculated analytically in the case of independent variables. 

\subsection{POD-PCE-3DVAR metamodelling error covariance matrix}
\label{subsection:error}
Replacing the model $\mathbf{y}=\mathcal{G}(\mathbf{x})$ by a surrogate denoted $\widetilde{\mathcal{G}}(\mathbf{x})$ comes with an additional modelling error. This error can therefore be accounted for in the optimization process, as defined in the following for POD-PCE-3DVAR. \\

When considering the POD-PCE metamodel, a new cost function can be written as in Equation \ref{eq:materials:PODPCEDA:metamodelCostFunction}, which could more be formulated in weaker constraint form as in \ref{eq:materials:metamodelCostFunctionStrong}  
\begin{equation}
\label{eq:materials:metamodelCostFunctionStrong}
J(\mathbf{x}) \approx \widetilde{J}(\mathbf{x})  =  \frac{1}{2} \left\| \mathbf{x} -  \mathbf{x}^{(b)}  \right\|^2_{\mathbf{B}^{-1}} + \frac{1}{2}\left\|\widetilde{\mathcal{G}}(\mathbf{x}) - \mathbf{y}^{(o)}\right\|^2_{\widetilde{\mathbf{R}}^{-1}} \ ,
\end{equation}
where $\widetilde{\mathbf{R}}$ is a new error covariance matrix accounting for metamodelling error. Indeed, with this surrogate, the model can no longer be considered perfect, even for an originally strong constraint 3DVAR problem (with perfect model $\mathcal{G}$ assumption). A new error denoted $\boldsymbol{\epsilon}^{(meta,o)}$, combining the observation and metamodelling errors, can be defined, and gives the new error covariance matrix as $\widetilde{\mathbf{R}}=\mathbb{V} \left(\boldsymbol{\epsilon}^{(meta,o)},\boldsymbol{\epsilon}^{(meta,o)}\right)$. \\

The POD-PCE metamodel in Equation \ref{eq:materials:PODPCEDA:metamodel} is used to define the new cost function as in Equation \ref{eq:materials:metamodelCostFunctionStrong}, with new error $\boldsymbol{\epsilon}^{(meta,o)}$, resulting from combined POD reduction error $\boldsymbol{\epsilon}^{POD}$, PCE approximation error $\boldsymbol{\epsilon}^{PCE}$ and observation error $\boldsymbol{\epsilon}^{(o)}$. \\

Firstly, POD basis truncation results in an approximation written in Equation \ref{eq:materials:PODPCEDA:modelStatePODError}, 
where $\boldsymbol{\epsilon}^{POD}$ is calculated as in Equation \ref{eq:materials:PODPCEDA:PODError}, and $\underline{\boldsymbol{\nu}^{(d)}}$ designates the POD coefficients not considered in the approximate.
\begin{equation} 
\label{eq:materials:PODPCEDA:modelStatePODError}
\mathbf{y}  =  \overline{\mathbf{y}} +\boldsymbol{\Phi}^{(d)} \boldsymbol{\Sigma}^{(d)}\boldsymbol{\nu}^{(d)} + \boldsymbol{\epsilon}_{POD} \ ,
\end{equation}

\begin{equation} 
\label{eq:materials:PODPCEDA:PODError}
\boldsymbol{\epsilon}^{POD} \coloneqq  \underline{\boldsymbol{\Phi}^{(d)}}~\underline{\boldsymbol{\Sigma}^{(d)}}~\underline{\boldsymbol{\nu}^{(d)}} \ .
\end{equation}

Secondly, PCE approximation error, resulting from coefficients fitting and polynomial basis truncation to a given polynomial degree, is defined as in Equation \ref{eq:materials:PODPCEDA:PCEerror}.
\begin{equation} 
\label{eq:materials:PODPCEDA:PCEerror}
\boldsymbol{\epsilon}^{PCE} \coloneqq \boldsymbol{\nu}^{(d)} - \widetilde{\boldsymbol{\nu}}^{(d)}(\mathbf{x})  \ .
\end{equation}

Consequently, exact formulation of the metamodel in Equation \ref{eq:materials:PODPCEDA:metamodel} with error characterization is written in Equation \ref{eq:materials:PODPCEDA:exactPODPCEmetamodel}, where error $\boldsymbol{\epsilon}^{(meta)}$ is defined in Equation \ref{eq:materials:PODPCEDA:metamodelError}. 
\begin{equation}
\label{eq:materials:PODPCEDA:exactPODPCEmetamodel}
\mathbf{y}  = \widetilde{\mathcal{G}}(\mathbf{x}) + \boldsymbol{\epsilon}^{(meta)} \ .
\end{equation}
\begin{equation}
\label{eq:materials:PODPCEDA:metamodelError}
\boldsymbol{\epsilon}^{(meta)} = \boldsymbol{\Phi}^{(d)}\boldsymbol{\Sigma}^{(d)}\boldsymbol{\epsilon}^{PCE} + \boldsymbol{\epsilon}^{POD}  \ .
\end{equation}

Lastly, combined metamodel and observation error can be calculated as in Equation \ref{eq:materials:PODPCEDA:obsAndmetamodelError}.
\begin{equation}
\label{eq:materials:PODPCEDA:obsAndmetamodelError}
    \begin{matrix}
    \epsilon^{(meta,o)} & = & \mathbf{y}^{(o)} - \widetilde{\mathcal{G}}(\mathbf{x}^{(t)})\\
    & = &  \mathbf{y}^{(o)} - \left(\mathcal{G}(\mathbf{x}^{(t)}) -  \boldsymbol{\Phi}^{(d)} \boldsymbol{\Sigma}^{(d)}\boldsymbol{\epsilon}^{PCE} - \epsilon^{POD}\right) \\
    & =  & \boldsymbol{\epsilon}^{(o)} + \boldsymbol{\Phi}^{(d)}\boldsymbol{\Sigma}^{(d)}\epsilon^{PCE} + \epsilon^{POD} 
    \end{matrix}
\end{equation}

To estimate the new error defined in Equation \ref{eq:materials:PODPCEDA:obsAndmetamodelError}, observation errors, POD error and PCE errors for modes $1$ to $d$, are considered mutually independent and unbiased ($\mathbb{E}[\mathbf{\epsilon}] = 0$). Independence is assumed in the sense that both methods (patterns extraction and learning) are performed independently. \\

Independence and bilinearity allow to develop the error covariance matrix as in Equation \ref{eq:materials:PODPCEDA:metamodelErrorCovariance}.
\begin{equation}
\label{eq:materials:PODPCEDA:metamodelErrorCovariance}
    \begin{matrix}
    \widetilde{\mathbf{R}} & = & \mathbb{V}\left(\boldsymbol{\epsilon}^{(meta,o)}, \boldsymbol{\epsilon}^{(meta,o)}   \right) \\
    & = & \mathbb{V}\left(\boldsymbol{\epsilon}^{(o)} +  \boldsymbol{\Phi}^{(d)}\boldsymbol{\Sigma}^{(d)}\epsilon^{PCE} + \epsilon^{POD} ,  \epsilon^{(o)} +  \boldsymbol{\Phi}^{(d)}\boldsymbol{\Sigma}^{(d)}\epsilon^{PCE} + \epsilon^{POD} \right) \\
    & = & \mathbf{R} +  \mathbb{V} \left(\boldsymbol{\epsilon}^{POD}, \boldsymbol{\epsilon}^{POD}\right) +  \boldsymbol{\Phi}^{(d)} \boldsymbol{\Sigma}^{(d)} \mathbb{V}\left(\boldsymbol{\epsilon}^{PCE} , \boldsymbol{\epsilon}^{PCE} \right) \boldsymbol{\Sigma}^{(d)} \left(\boldsymbol{\Phi}^{(d)}\right)^T
    
    \end{matrix}
\end{equation}

Then, POD error covariance matrix is developed in Equation \ref{eq:materials:PODPCEDA:PODerrorCovariance}. 
\begin{equation}
\label{eq:materials:PODPCEDA:PODerrorCovariance}
\begin{matrix}
    \mathbb{V} \left(\boldsymbol{\epsilon}^{POD}, \boldsymbol{\epsilon}^{POD}\right) &  = & 
    \mathbb{V} \left( \underline{\boldsymbol{\Phi}^{(d)}}~\underline{\boldsymbol{\Sigma}^{(d)}}~\underline{\boldsymbol{\nu}^{(d)}}, \underline{\boldsymbol{\Phi}^{(d)}}~\underline{\boldsymbol{\Sigma}^{(d)}}~\underline{\boldsymbol{\nu}^{(d)}} \right) \\
    & = & \underline{\boldsymbol{\Phi}^{(d)}}~\underline{\boldsymbol{\Sigma}^{(d)}}~ \mathbb{V} \left(\underline{\boldsymbol{\nu}^{(d)}},\underline{\boldsymbol{\nu}^{(d)}}  \right) ~\underline{\boldsymbol{\Sigma}^{(d)}}~ \left(\underline{\boldsymbol{\Phi}^{(d)}}\right)^T
    \end{matrix} \ .
\end{equation}

  The covariance matrix $\mathbb{V} \left(\underline{\boldsymbol{\nu}^{(d)}},\underline{\boldsymbol{\nu}^{(d)}}\right)$ can be estimated as the \textit{ensemble-anomaly covariance matrix} \citep{Carrassi2018}, written in Equation \ref{eq:materials:PODPCEDA:PODerrorEnsembleAnomaly}. Indeed, POD was deduced from an ensemble of realizations of $\mathbf{y}$, which provides $n$ realizations for vector $\underline{\boldsymbol{\nu}}$, stored in the columns of matrix $(\underline{\mathbf{N}^{(d)}})^T \in \mathbb{R}^{(e-d)\times n}$, with $e=min(n,m)$ (details in Section \ref{subsubsection:PODEn3DVAR:POD}). Covariance is then developped in Equation \ref{eq:materials:PODPCEDA:PODerrorEnsembleAnomaly} with unbiased error that cancels $\mathbb{E}\left[\underline{\mathbf{N}^{(d)}}\right] = \left(\underline{\boldsymbol{\Sigma}^{(d)}}\right)^{-1}~ \left(\underline{\boldsymbol{\Phi}^{(d)}}\right)^T\mathbb{E}\left[\boldsymbol{\epsilon}^{POD}\right]$ and orthonormality of POD coefficients that gives $\left(\underline{\mathbf{N}^{(d)}}\right)^T\left(\underline{\mathbf{N}^{(d)}}\right)=\mathbf{I}_{e-d}$.
 \begin{equation}
\label{eq:materials:PODPCEDA:PODerrorEnsembleAnomaly}
\begin{matrix}
\mathbb{V} \left(\underline{\boldsymbol{\Upsilon}^{(d)}},\underline{\boldsymbol{\Upsilon}^{(d)}}  \right) & \approx &  \dfrac{1}{n-1}\left(\underline{\mathbf{N}^{(d)}}- \mathbb{E}\left[\underline{\mathbf{N}^{(d)}}\right] \right)^T\left(\underline{\mathbf{N}^{(d)}} - \mathbb{E}\left[\underline{\mathbf{N}^{(d)}}\right] \right) \\
& = & \dfrac{1}{n-1}\left(\underline{\mathbf{N}^{(d)}}\right)^T\left(\underline{\mathbf{N}^{(d)}} \right)  \\
& = & \dfrac{1}{n-1} \mathbf{I}_{e-d}
    
    \end{matrix} \ .
\end{equation}

PCE error covariance matrix, considering modes fitting errors are independent, can be estimated as: 
\begin{equation}
\begin{matrix}
   \mathbb{V} \left(\boldsymbol{\epsilon}^{PCE}, \boldsymbol{\epsilon}^{PCE}\right)  & = & \mathbb{V}\left( \boldsymbol{\nu}^{(d)} - \widetilde{\boldsymbol{\nu}}^{(d)}, \boldsymbol{\nu}^{(d)} - \widetilde{\boldsymbol{\nu}}^{(d)}\right) \\
    & & \\
    & = & \left( \begin{matrix}
        \mathbb{V}\left[\nu_1 - \widetilde{\nu_1}\right] & 0 & \hdots  & 0 \\
        0 & \mathbb{V}\left[\nu_2 - \widetilde{\nu_2}\right] & \hdots & 0 \\
        \vdots & \vdots & \ddots & \vdots\\
       0 &  0 & \hdots &  \mathbb{V}\left[\nu_d - \widetilde{\nu_d}\right]
    \end{matrix} \right)
    \end{matrix}
\end{equation}

This independence assumption, that allows to write a diagonal matrix, is possible since each POD mode is learned independently by corresponding PCE model. Then, each variance term can be written in terms of expectation with the Koening-Huygens development. As PCE errors are supposed unbiased (no average over or under estimation), this gives:
\begin{equation}
\label{eq:materials:PODPCEDA:PCEvariance}
\begin{matrix}
\mathbb{V}\left[\nu_k - \widetilde{\nu_k}\right] & = & \mathbb{E}\left[(\nu_k - \widetilde{\nu_k})^2\right] - \mathbb{E}\left[(\nu_k - \widetilde{\nu_k})\right]^2 \\
& = & \mathbb{E}\left[(\nu_k - \widetilde{\nu_k})^2\right]
\end{matrix}
\end{equation}

Quantity $\mathbb{E}\left[(\nu_k - \widetilde{\nu_k})^2\right]$ is called  \textit{generalization error} \citep{Blatman2009}. It can be approximated at PCE learning stage with the \textit{empirical error} denoted $\delta_{emp}$ defined in Equation \ref{eq:empiricalError}. This is performed by splitting the ensemble of realizations $(\mathbf{N}^{(d)})^T$ to a training set of size $n_t$ and a validation/prediction set of size $n_p = n - n_t$. The training ensemble is then used to fit the PCE models, and the validation ensemble, with reduced state realizations $\left[\nu_k(\mathbf{x}^1), \hdots,  \nu_k(\mathbf{x}^{n_p})\right]$ associated to parameters realizations $\left[\mathbf{x}^1, \hdots, \mathbf{x}^{n_p}\right]$, is used to estimate the learned model bias in Equation \ref{eq:empiricalError}.
\begin{equation}
  \label{eq:empiricalError}
  \mathbb{E}\left[(\nu_k - \widetilde{\nu_k})^2\right] \approx \delta_{emp}(\nu_k, \widetilde{\nu_k}) \coloneqq \dfrac{1}{n_p} \sum_{j=1}^n \left( \nu_k(\mathbf{x}^j) - \widetilde{\nu_k}(\mathbf{x}^j) \right)^2\ .
  \end{equation}

As explained above, generalization error is used in this study to fit an optimal PCE model for each component $\nu_k$ independently (choice of polynomial degree), by minimization on the prediction set. Consequently, the resulting covariance estimation comes with no additional cost than fitting. \\

As a result, error $\widetilde{\mathbf{R}}$ can be estimated as in Equation  \ref{eq:metamodelCov}. 
\begin{equation}
\label{eq:metamodelCov}
    \widetilde{\mathbf{R}} \approx \mathbf{R} +  \dfrac{1}{n-1} ~ \underline{\boldsymbol{\Phi}^{(d)}} ~ \underline{\boldsymbol{\Lambda}^{(d)}}  \left(\underline{\boldsymbol{\Phi}^{(d)}}\right)^T +  \boldsymbol{\Phi}^{(d)} \left( \begin{matrix}
         \lambda_1\delta_{emp}(\nu_1 - \widetilde{\nu_1}) & 0 & \hdots  & 0 \\
        0 &  \lambda_2\delta_{emp}(\nu_2 - \widetilde{\nu_2}) & \hdots & 0 \\
        \vdots & \vdots & \ddots & \vdots\\
       0 &  0 & \hdots &  \lambda_d\delta_{emp}(\nu_d - \widetilde{\nu_d})
    \end{matrix} \right) \left(\boldsymbol{\Phi}^{(d)}\right)^T \ .
\end{equation}

It is directly calculated from: \begin{itemize}
    \item[$\bullet$] the eigenvalue matrix associated to unused modes $\underline{\boldsymbol{\Lambda}^{(d)}}=\left(\underline{\boldsymbol{\Sigma}^{(d)}}\right)^2$ representing an Unexplained Variance Rate (UVR, oppositely to EVR) due to POD truncation;
    \item[$\bullet$] the approximation error due to PCE learning of each mode $\nu_k$, enhanced by the importance of the latter, represented by its eigenvalue $\lambda_k$. \\
\end{itemize}
For its estimation, no additional computation is necessary, besides POD calculation and PCE fitting by generalization error minimization. New error covariance is therefore directly estimated by transforming reduction and learning errors to the output's space through adequate matrix product using the full POD basis $\boldsymbol{\Phi} = \left[\boldsymbol{\Phi}^{(d)}, \underline{\boldsymbol{\Phi}^{(d)}}\right]$. \\

It is worth mentioning that the same analysis can be performed for PODEn3DVAR, simply by skipping the calculation of error portion that is due to PCE. Its impact was however not investigated in the following work, and focus was rather put on POD-PCE-3DVAR optimization, as the latter has proven, through presented application, to be more efficient and robust to noise (details and comparisons in following Section \ref{section:application}).



\shorthandoff{:}


\shorthandoff{:}

\section{Numerical experiments with the Shallow Water Equations}

\label{section:application}
The accuracy of the proposed hybrid DA approaches, presented in Section \ref{section:materials}, is now assessed on practical examples. A model $\mathcal{M}$, with uncertain parameters $\mathbf{x}$ and associated state $\mathbf{y}$, and observations $\mathbf{y}^{(o)}$, should be specified for cost function defined in Equation \ref{eq:materials:DA:3DVARcostFunction}. The latter are presented in Section \ref{subsection:application:problem}. Then, a twin DA experiment, and a measurement-based DA experiment are presented in Sections \ref{subsection:application:twin} and  \ref{subsection:application:measurements} respectively. In particular in Section \ref{subsection:application:measurements}, results with the proposed algorithms are compared to classical 3DVAR. It should be noted that the following results are presented for PODEn3DVAR using the analytical gradient derivation as in Section \ref{subsection:PODEn3DVAR}, and for POD-PCE-3DVAR using an iterative gradient descent (that would be more convenient in case of dependent parameters). More precisely,  the advanced constrained
Broyden-Fletcher-Goldfarb-Shanno Quasi-Newton (c-BFGS-QN) method is used to impose variation bounds for the parameters to optimize \citep{Zhu1997}.

\subsection{Problem setup}
\label{subsection:application:problem}
\begin{figure}[H]
  \centering
  \subfloat[][Intake scheme and observation points]{\includegraphics[trim={0cm 0cm 0cm 0.cm},clip,width=0.25\textwidth]{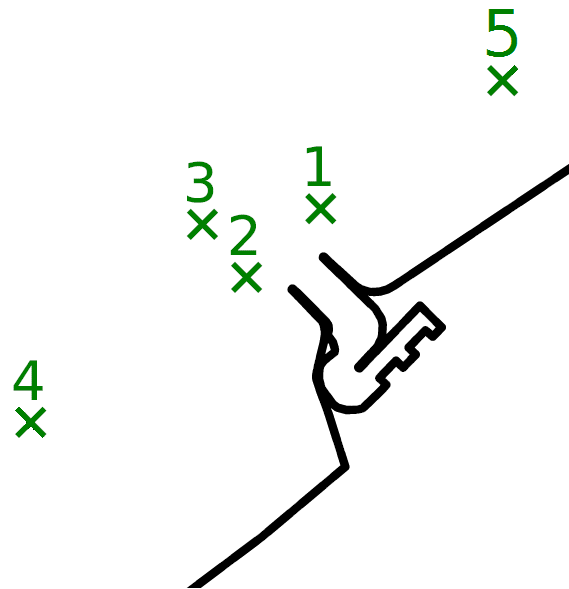}}
  \subfloat[][Computation domain and bathymetry]{\hspace{1cm} \includegraphics[trim={0cm 0cm 0cm 0.cm},clip,width=0.4\textwidth]{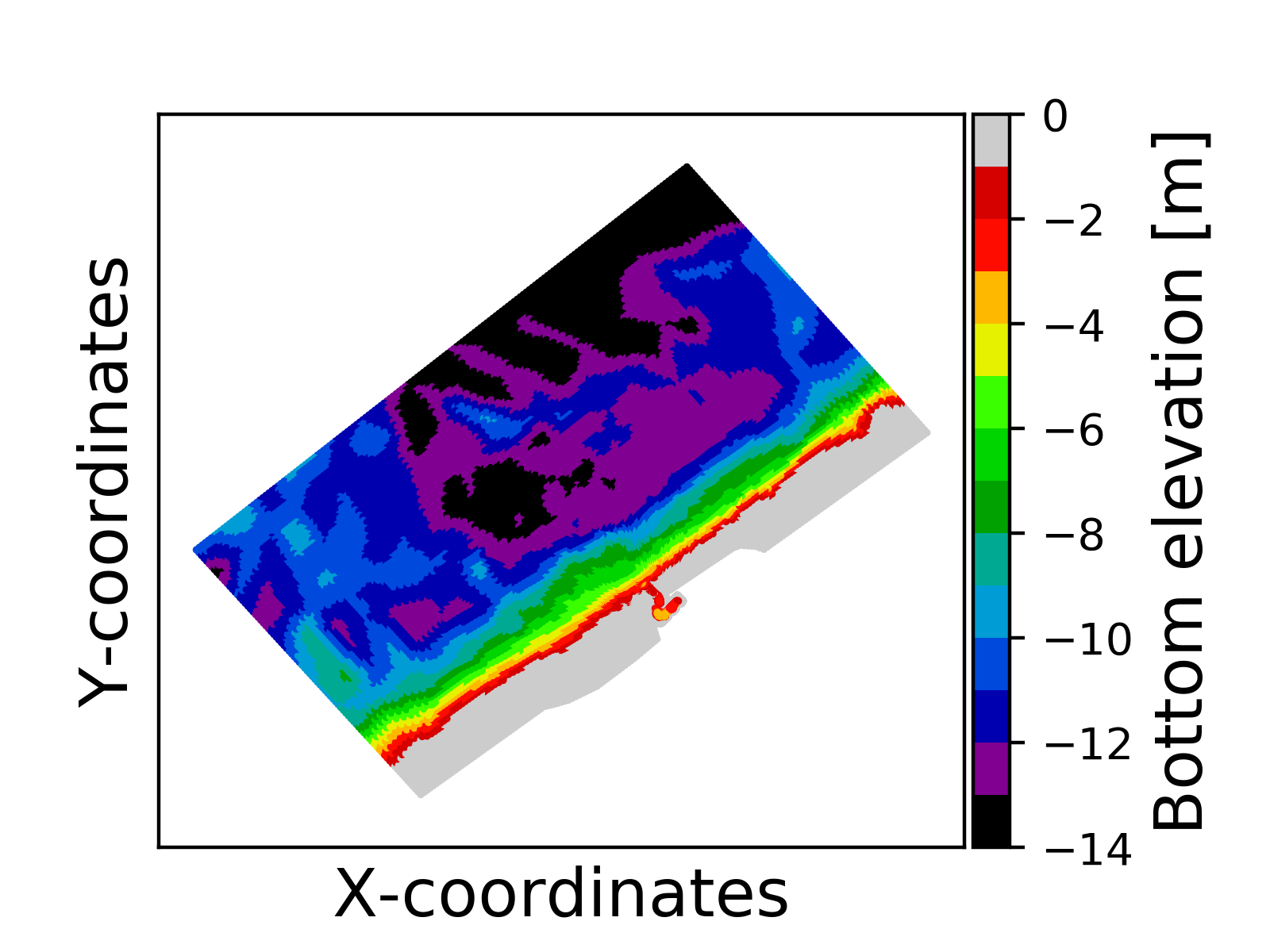}}
  \caption{Locations of measurement points for the two-months survey at intake's vicinity}
  \label{fig:application:case}
\end{figure}

The modelling of tidal flow at the vicinity of a power plant's cooling intake is here targeted. The studied intake is located on the eastern English Channel coast in northern France, which is a macro/mega-tidal zone dominated by a semi-diurnal circulation. Few field information about the currents are available, namely a survey of depth-averaged velocity components denoted $(u,v)^T$ and free-surface elevation denoted $\eta$ at five measurement points, indicated with a schematic drawing of the intake in Figure \ref{fig:application:case}-a. 

\begin{figure}[H]
  \centering
  \subfloat[][Free surface]{\includegraphics[trim={0cm 0cm 0cm 0cm},clip,scale=0.5]{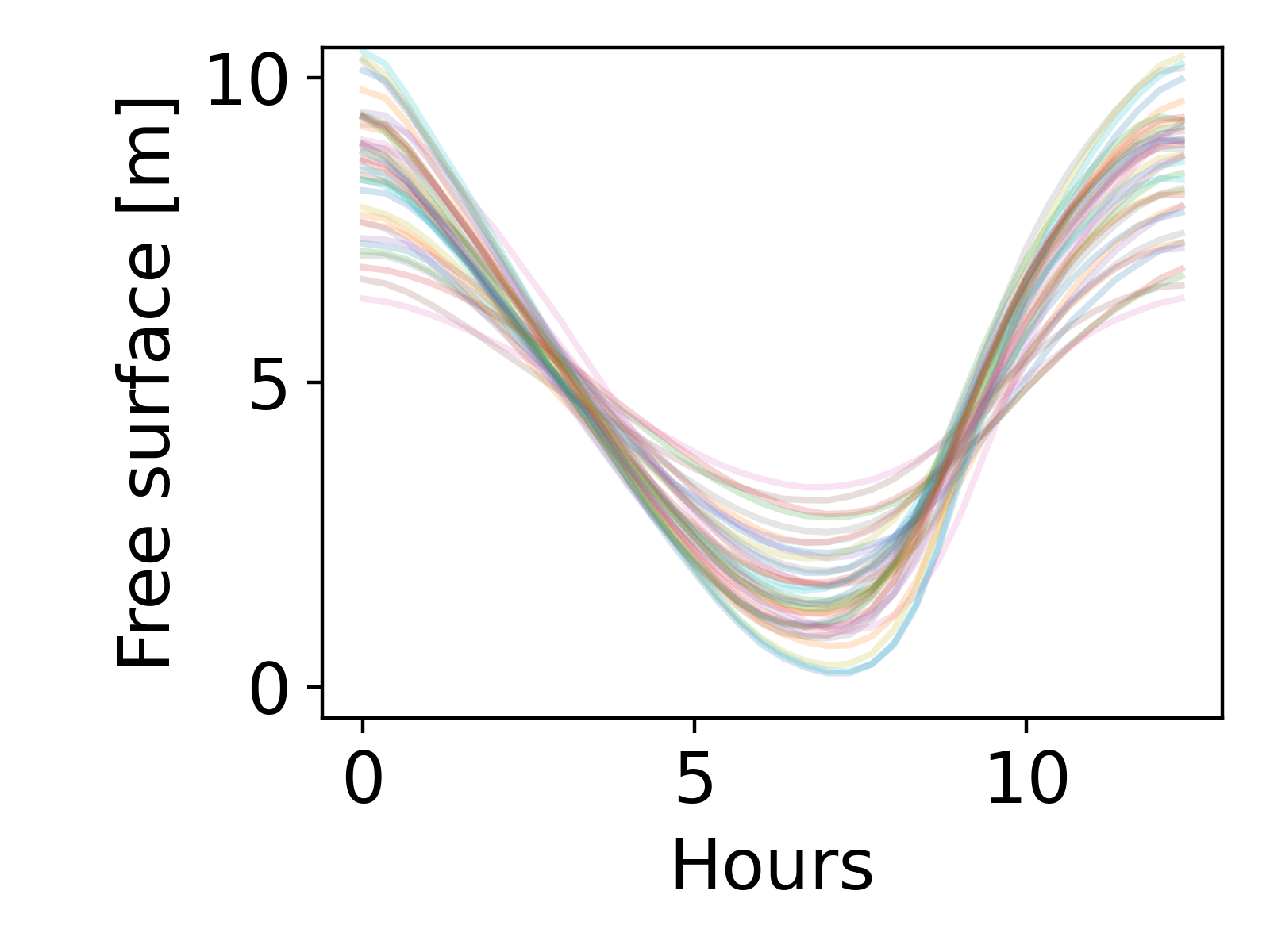}}
  \subfloat[][X-velocity $u$]{\includegraphics[trim={0cm 0cm 0cm 0cm},clip,scale=0.5]{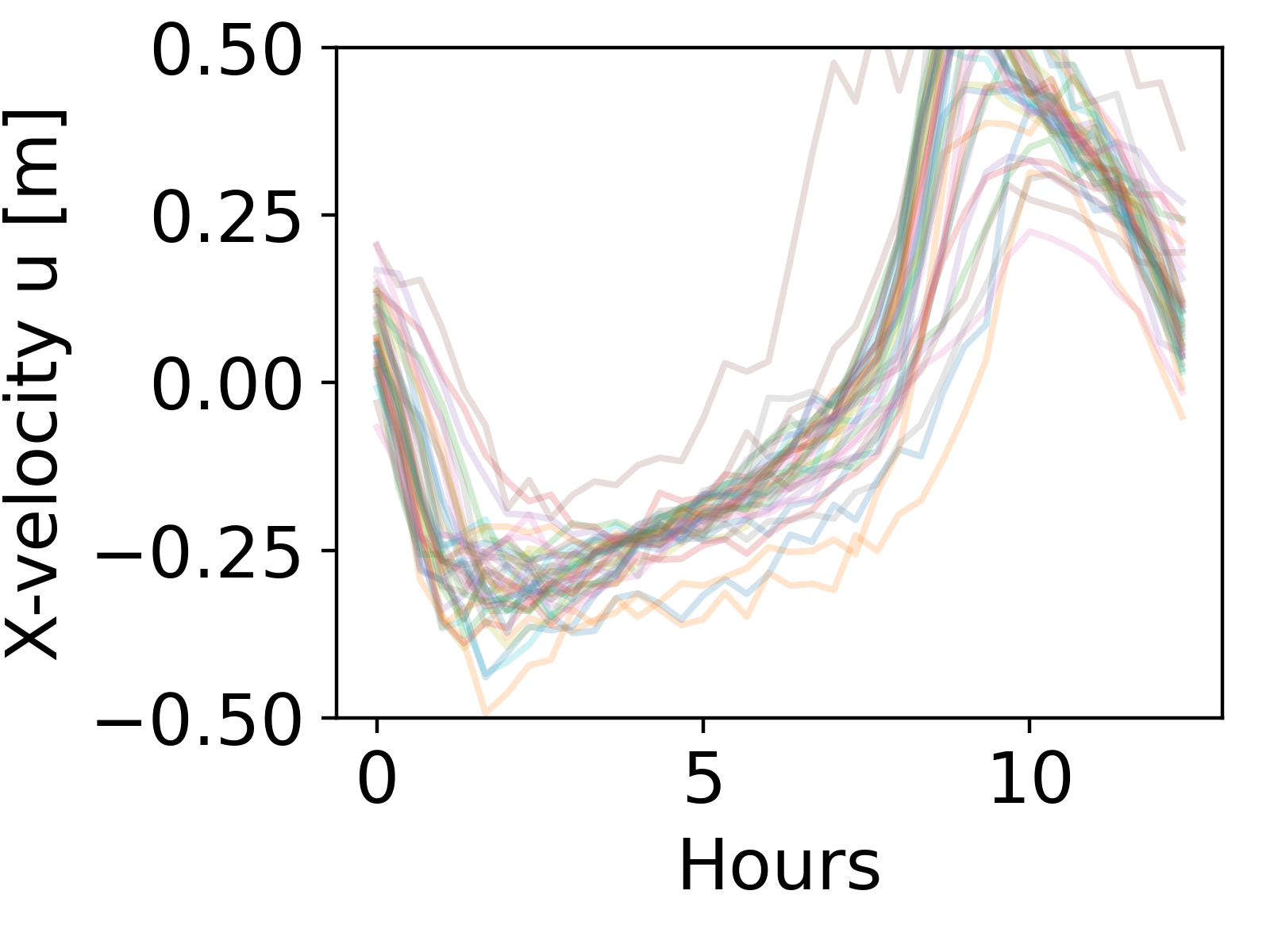}}
  \subfloat[][Y-velocity $v$]{\includegraphics[trim={0cm 0cm 0cm 0cm},clip,scale=0.5]{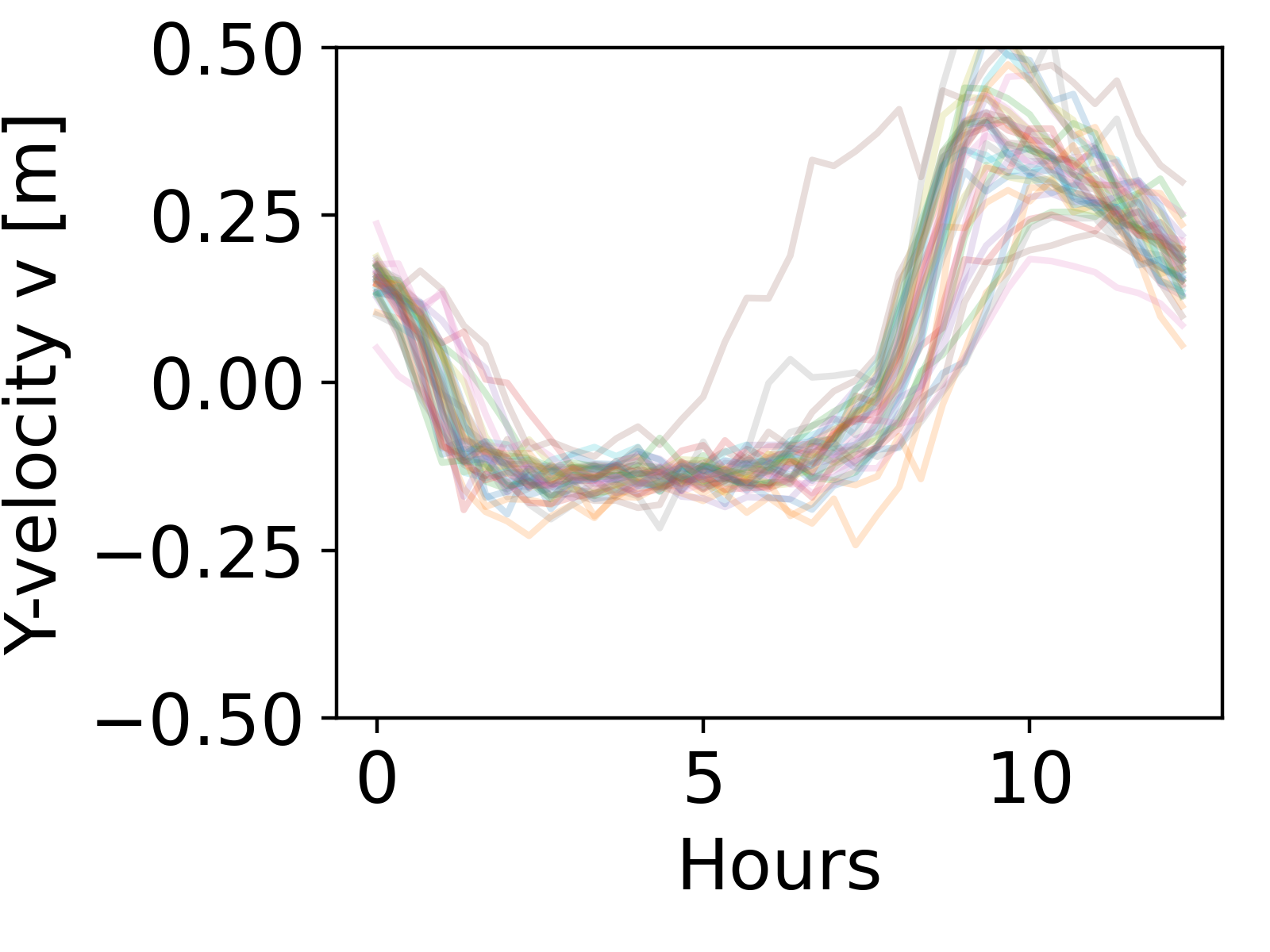}}
  \caption{Superposition of measurements for the hydrodynamic variables on Point 1}
  \label{fig:manuscript:context:data:2010}
\end{figure}

A superposition of the measured hydrodynamic variables on tidal periods is shown in Figure \ref{fig:manuscript:context:data:2010}. The field campaign period is characterized with possible storms and surges. 

These data are subject to measurement errors. In particular, Acoustic Wave And Current (AWAC) meters ($1$ MHz Nortek, Doppler technology) were used for offshore tidal velocities and water depths (through pressure records). Measurements are characterized with absolute errors $1$ cm/s for and $5$ cm for tidal velocities and free-surface elevations respectively.  \\

To obtain a detailed spatio-temporal current distribution in the study area, the Shallow Water Equations (SWE) are used. The SWE are a set of non-linear hyperbolic Partial Differential Equations (PDE), obtained by depth-averaging the three-dimensional Reynolds-averaged free-surface Navier-Stokes equations, allowing the representation of almost-horizontal, two-dimensional (2D), shallow flows \citep{GerbeauPertham2000}. The resulting mass and momentum conservation equations are defined in Equation \ref{eq:application:SWE}.
\begin{equation}\label{eq:application:SWE}
\left\{
	\begin{array}{l}
	  \dfrac{\partial h}{\partial t} + \dfrac{\partial (hu)}{\partial x}  + \dfrac{\partial (hv)}{\partial y} = 0\\
          \dfrac{\partial(hu)}{\partial t} + \dfrac{\partial(hu^2)}{\partial x} + \dfrac{\partial(huv)}{\partial y} = -gh\dfrac{\partial \eta}{\partial x} -  \dfrac{1}{\rho} {\tau_b}_x + \dfrac{h}{\rho} F_x + \boldsymbol{\nabla} \cdot (h \nu_e \nabla u) \\
          \dfrac{\partial(hv)}{\partial t} + \dfrac{\partial(huv)}{\partial x} + \dfrac{\partial(hv^2)}{\partial y} = -gh\dfrac{\partial \eta}{\partial y} -  \dfrac{1}{\rho} {\tau_b}_y +  \dfrac{h}{\rho} F_y + \boldsymbol{\nabla} \cdot (h \nu_e \nabla v)          
          
	\end{array}
\right.
\end{equation}
where the system unknowns are the depth-averaged velocity components $\mathbf{u}=(u,v)^T$ along the Cartesian coordinates $(x,y)$ and the free surface elevation $\eta \coloneqq h+b$, with h the water depth and b the bottom elevation. The gravitational acceleration $g$ and the water density $\rho$ are considered as constant values. Vector $\boldsymbol{\tau_b}=({\tau_b}_x,{\tau_b}_y)$ denotes the bottom shear stress, with components ${\tau_b}_x$ and ${\tau_b}_y$ along the Cartesian coordinates $(x,y)$. Vector $\mathbf{F}$ represents external forces (in the presented study, only Coriolis effect is considered as external force), and $\nu_e$ is the effective viscosity, accounting for kinematic, eddy and “dispersion” viscosity resulting from vertical integration, and here set equal to water’s kinematic viscosity. \\

The Digital Elevation Model from \citep{Shom2015} provides the bottom elevation $b$, represented in Figure \ref{fig:application:case}-b. This model set-up, although characterized with possible approximations (e.g. bottom interpolations), is considered as trustworthy for the ongoing study. However, bottom friction and tidal Boundary Conditions (BC), which are essential elements for tidal modelling with SWE, are unknown, and their uncertainty should be quantified for optimal calibration using DA. In the following, attention is therefore focused on parametric calibration of friction and BC using appropriate variables. However, it should be noted that the developed framework could be used to calibrate other control vectors, as the bathymetry, Initial Conditions (IC), etc. \\

Firstly, bed shear stress is capital for environmental applications, as it has considerable influence on the flow because of the energy dissipation it induces \citep{Morvan2008}. Its exact formulation remains unknown, but parametrization can be done using one of the many formulas that can be found in literature, with specific calibration parameters. Sensitivity analysis performed in \citep{Mouradi2020} on the study case showed that formula choice does not impact the model response. Consequently, the empirical Strickler’s formula \citep{Morvan2008} is used as in Equation \ref{eq:application:StricklerbottomFriction}, where $K$ is called Strickler coefficient. The latter is an uncertain calibration parameter, that ranges in $[21.02 ,90.66]$ m$^{1/3}$ s$^{-1}$, as explained in \citep{Mouradi2020}. Two values are set: $K_1$ inside the intake and $K_2$ outside. Model outputs at measurement points of Figure \ref{fig:application:case} showed no sensitivity to $K_1$ in \citep{Mouradi2020}. Therefore, it is fixed to average interval value, and only $K_2$ is calibrated using DA.
\begin{equation}
  \label{eq:application:StricklerbottomFriction}
  \boldsymbol{\tau_b} = \frac{\rho g}{K^2}\left(\frac{1}{h}\right)^{1/3} |\mathbf{u}|\mathbf{u}
\end{equation}

Secondly, tidal BC are parametrized with the TPXO data-base \citep{Egbert2002}, particularly the European Shelf (ES) local model implemented in the TELEMAC-2D module \citep{Pham2012} of the open-source TELEMAC-MASCARET modelling System (TMS) (\url{https://www.opentelemac.org/}) used in this study. The hydrodynamic unknowns at the boundary are modelled as a superposition of harmonic components, as in Equation \ref{eq:application:harmonics},
\begin{equation}
\label{eq:application:harmonics}
F(x,y,t) = \sum F_i(x,y,t)  = \sum  f_i(t)A_{F_i}(x,y)cos\left(2 \pi t/T_i - \phi_{F_i}(x,y) + u_i^0 + v_i(t)\right) \ ,
\end{equation}
where the term $F$ at point of coordinates $(x,y)$ and time $t$ represents velocity components or water depth, $F_i$ a harmonic component with constant period $T_i$, amplitude $A_{F_i}$, phase $\phi_{F_i}$, phase at origin of times $u_i^0$, and temporal nodal factors $f_i(t)$ and $v_i(t)$. Thompson’s method is then used to prescribe BC \citep{Hervouet2007}, and three parameters, denoted $CTL$ (Coefficient of Tidal Level), $MTL$ (Mean Tidal Level) and $CTV$ (Coefficient of Tidal Velocity), can be used to calibrate the BC on measurements, as in Equation \ref{eq:application:correctionBC}. For example, $MTL$ allows to account for seasonal variability (effect of thermal expansion, salinity variations, air pressure, etc.) in addition to long-term sea level rise resulting from climate change \citep{Idier2019}, and all three parameters can be used to compensate the effects of storm and surge (atmospheric and wave setup) \citep{Idier2019}, as the latter are not modelled in the TPXO data-base. These calibration coefficients additionally allow to correct the approximations resulting from scale difference between the TPXO data-base and the model geometry, as the used domain size is smaller than a TPXO element size. Hence, a linear interpolation is used to impose tidal conditions on the studied geometry's boundary elements. Variation interval for $MTL$ is deduced from measurements as $[4.0, 6.0]$ m CM, whereas the non-dimensional parameters $CTL$ and $CTV$ are expertly set to $[0.8, 1.2]$ and $[0.8, 3.0]$ respectively, so that the measurements fall within the simulated min-max interval \citep{Mouradi2020}. 
\begin{equation}
\label{eq:application:correctionBC}
\begin{matrix}
h(x,y,t) & = & CTL \times \sum h_i(x,y,t) - z_f(x,y) + MTL \\
u(x,y,t) & = & CTV \times \sum u_i(x,y,t) \\
v(x,y,t) & = & CTV \times \sum v_i(x,y,t)
\end{matrix}
\end{equation}

To conclude, four uncertain parameters are defined for DA calibration: one input for sea friction denoted $K_2$, and three BC parameters denoted $MTL$, $CTL$ and $CTV$. As the only available information about these variables uncertainties are their variation supports, Uniform PDFs should be used to represent their uncertainty (\textit{Maximum Entropy Principle} \citep{Soize2017}). These PDFs are used to produce random ensembles of inputs using Monte Carlo (MC) sampling. Each input configuration is then propagated through the studied model by running a direct simulation. A corresponding ensemble of output values is therefore obtained, which then allows to learn the joint parameter-state POD on the one hand, and the POD-PCE model on the other hand. For the latter, Legendre polynomials are used, and PCE is fitted for each POD pattern independently, by choosing the polynomial degree that minimizes the empirical error calculated in Equation \ref{eq:empiricalError}. For the variational algorithms however, all variables are characterized by Gaussian errors. In the following, the background parameters are set to the average value from the variation interval, and standard deviation empirically estimated from the bounds, as summarized in Table \ref{table:application:case:inputsBounds}. A large discrepancy is therefore considered for both twin and measurement-based experiments, using the full variation range of parameters, as the latter are generally unknown.
\begin{table}[H]
  \centering
  \begin{tabular}{|M{3cm}|M{2.5cm}|M{2.5cm}|M{2.5cm}|M{2.5cm}|M{2.5cm}|}
    \hline
    Variable $x_i$ & $x_i^{min}$ & $x_i^{max}$ & $\mathbb{E}[x_i]$ & $\sqrt{\mathbb{V}[x_i]}$ \\
\hline
 $K_2$ [m$^{1/3}$s$^{-1}$] & 21.02 & 90.66 & 55.84 & 34.82 \\
\hline
 $MTL$ [m CM] & 4.0 & 6.0 & 5.0 & 1.0 \\
\hline
 $CTL$ [-] & 0.8 & 1.3 & 1.05 & 0.25 \\
\hline
 $CTV$ [-] & 0.8 & 3.0 & 1.9 & 1.1 \\
\hline
  \end{tabular}
  \caption{Uncertain parameters bounds and considered standard deviation for the variational DA cost function}
  \label{table:application:case:inputsBounds}
\end{table}

Lastly, previously described governing equations are solved by the module TELEMAC-2D \citep{Hervouet2007} of TMS, that also incorporates the used closures (friction, tidal BC). A numerical domain of size $8\times 16$ km (offshore $\times$ longshore distances) is chosen, resulting from previous sensitivity analysis \citep{Mouradi2020}. After mesh convergence study, elements of maximal size $50$ m at the sea and $2.5$ m in the intake are selected.


\shorthandoff{:}

\subsection{Robustness investigations on twin experiment}
\label{subsection:application:twin}

Before attempting model calibration on complex measurements, one of the randomly produced model states is used to assess the performance of previously presented DA approaches in Section \ref{section:materials} on twin experiments. The model state consists in temporal series of the hydrodynamic variables, registered with a $20$ minutes time step, and measured at the five observation points  represented in Figure \ref{fig:application:case}-a, over a semi-diurnal tidal period. Model state is arranged in vector form as $\mathbf{y} \coloneqq [u_{P1}(t_1),\hdots,u_{P1}(t_{k}),\hdots,u_{P5}(t_{k}), v_{P1}(t_1),\hdots, v_{P5}(t_{k}), \eta_{P1}(t_1), \hdots, \eta_{P5}(t_{k})]$, where $Pi \in \{P1,\hdots,P5\}$ refer to record points and $t_j \in \{t_1,\hdots,t_{k}\}$ with $k \in \mathbb{N}^*$ the record times. The studied twin experiment model output is considered to be the "true" observation field, denoted $\mathbf{y}^{(t)}$. Perturbed versions denoted $\mathbf{y}^{(o)}$, with different noise levels ranging from $1~\%$ to $40~\%$, are used to emulate imperfect observations. Each component of $\mathbf{y}^{(t)}$ is perturbed using a white Gaussian noise around the true value, with a variance corresponding to chosen noise percentage. The variance employed for each component is also used to fill the observation error covariance matrix $\mathbf{R}$. Parameter prior values denoted $\mathbf{x}^{(b)}$ are set to average, as reported in Table \ref{table:application:case:inputsBounds}, and corresponding error covariances stored in $\mathbf{B}$ are calculated as deviation from truth $\mathbf{x}^{(t)}$, corresponding to the unperturbed twin experiment. It should also be noted that all states, observations and parameters components are centered and reduced (normalized by their standard deviations) in order to give the same weight to all. Corresponding error covariance matrices are normalized accordingly.   \\

Comparisons of PODEn3DVAR and POD-PCE-3DVAR are given, with three elements of interest: (i) their performance and convergence with ensemble size (ii) their sensitivity to noise and (iii) the influence of mode number (or EVR) choice. Confrontations are therefore performed by successively setting a parameter (noise, mode number) to an arbitrary fixed value, in order to make the analysis easier. \\ 

\begin{figure}[H]
  \centering
  \includegraphics[trim={0cm 0cm 0cm 0cm},clip,width=0.5\textwidth]{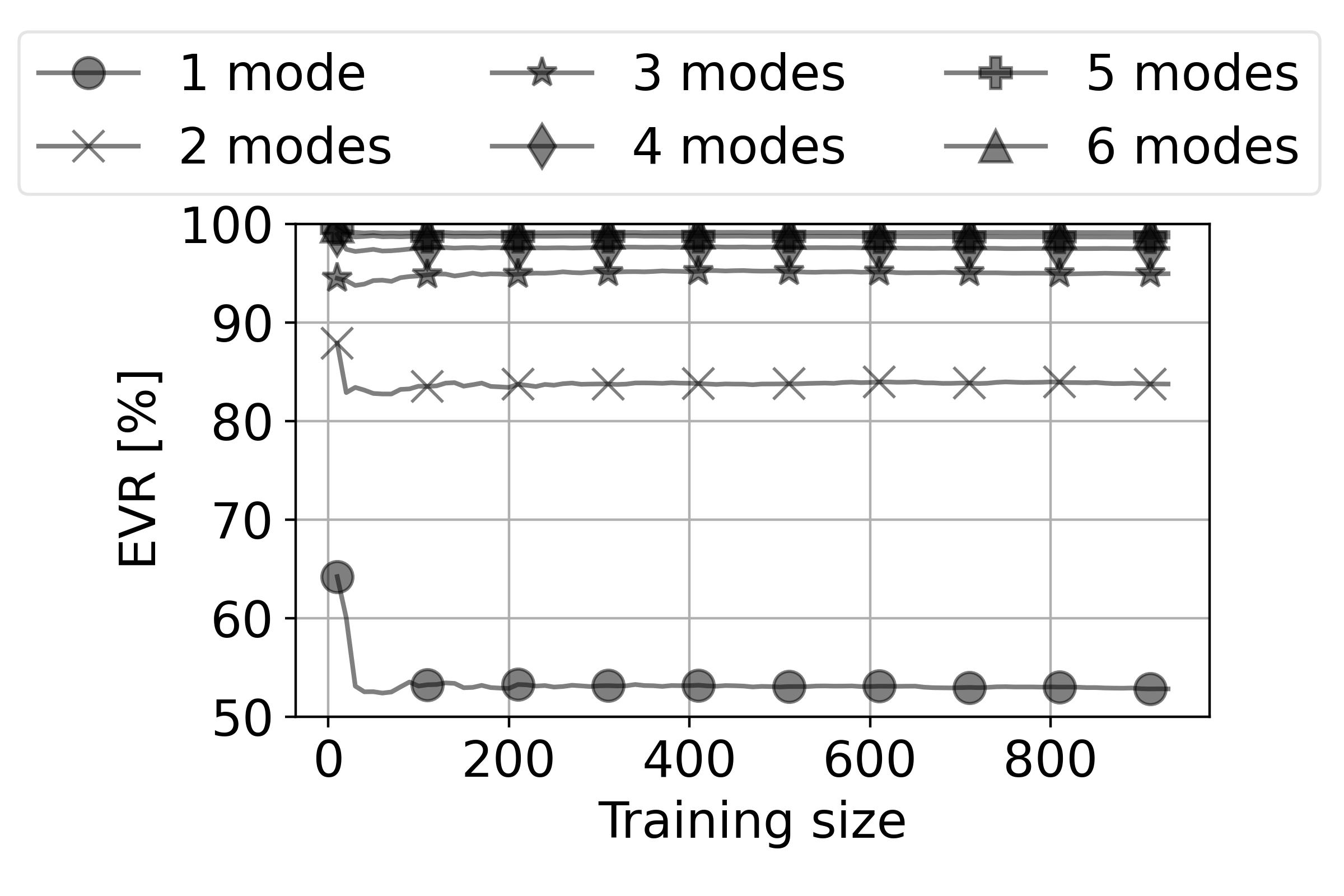}
    \caption{Convergence of the EVR with training size for different mode numbers. POD is here applied to the snapshot matrix containing model states $\mathbf{y}$}
    \label{fig:application:EVR}
\end{figure}

The evolution of EVR with mode number, and its convergence with training size, is for example shown in Figure \ref{fig:application:EVR} for POD applied to the snapshot matrix containing model states $\mathbf{y}$. For example, when $2$ modes are selected, over $80\%$ of the EVR is captured with the metamodel. The EVR of POD applied to joint parameter-state snapshot matrix shows substantially equivalent convergence and values. This proves that the relation between the model's degrees of freedom, represented by forcing parameters, and output states is well captured. The EVR of joint parameter-state matrix compared to state-only matrix are barely lower, not visible on a plot.

To perform comparisons between PODEn3DVAR and POD-PCE-3DVAR, the relative Root Mean Square Error (RMSE) measuring the distance between true state $\mathbf{y}^{(t)}$ and analyzed model output $\mathbf{y}^{(a)}$, and the relative RMSE between observation $\mathbf{y}^{(o)}$ and analyzed model output $\mathbf{y}^{(a)}$, are computed. These are calculated either on the full vector, called global RMSE, or for each point $P1$ to $P5$ and each output variable $z_s$, $u$ and $v$ independently. For RMSE computation on the full vector, each variable is standardized beforehand, i.e. centered around mean and divided by standard deviation, in order to give equivalent importance to all model outputs. \\

Lastly, the results detailed below show promising efficiency and robustness to noise for POD-PCE-3DVAR, therefore selected as best approach, for further investigation and optimization. More precisely, the impact of error covariance matrices choice is studied, as specification of the latter is known to have significant impact on DA algorithms performance \citep{Senegas2001}. The question is raised by the use of covariance matrix $\mathbf{R}$ to account for model-observation error, which is based upon perfect model assumption. This would have been valid if $\mathcal{M}$ was used, as the twin observation, before being perturbed with noise, originates from the model. The model is therefore theoretically able to recover true state with optimal definition of the cost function (true error covariance matrices) and perfect resolution of the latter. However, an approximated POD-PCE metamodel is used instead, characterized with errors, which can be considered through the use of $\widetilde{\mathbf{R}}$ instead of $\mathbf{R}$, as explained in Section \ref{section:materials}. Detailed study about the impact of error covariance matrix choice, regarding the quality of POD-PCE-3DVAR analysis, is hence finally proposed.

\subsubsection*{Noise impact}

\begin{figure}[H]
  \centering
  \subfloat{\hspace{-0.3cm} \includegraphics[trim={1cm 6.5cm 0cm 0.3cm},clip,width=0.5\textwidth]{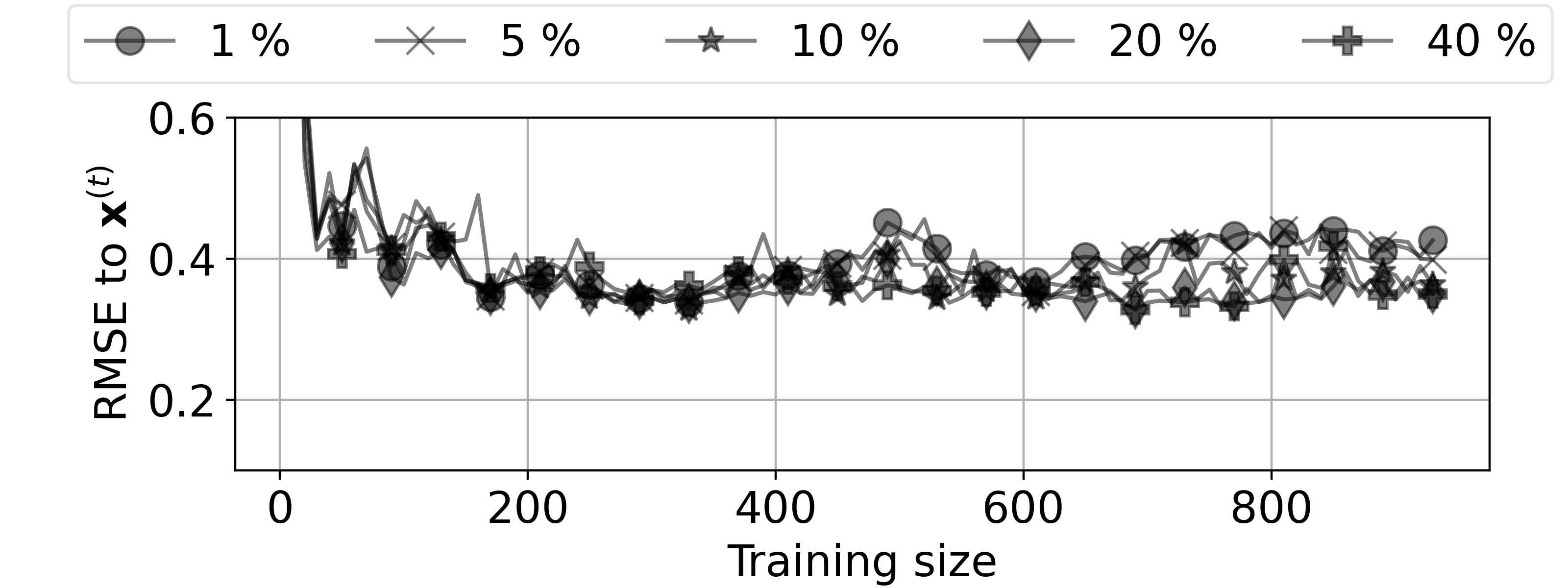}} \\
  \vspace{-0.4cm}
	\setcounter{subfigure}{0}
  \subfloat[][Distance to observation]{\includegraphics[trim={0.5cm 0cm 0cm 0cm},clip,width=0.5\textwidth]{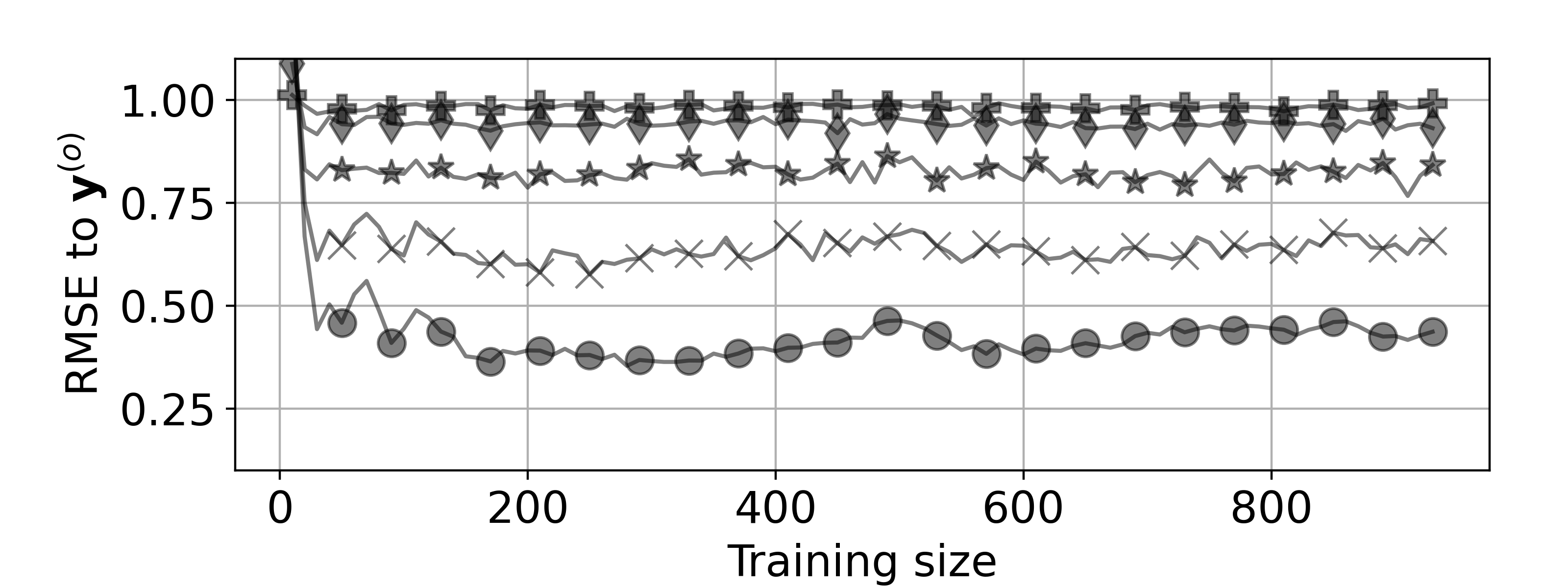}}
  \subfloat[][Distance to truth]{\includegraphics[trim={0.5cm 0cm 0cm 0cm},clip,width=0.5\textwidth]{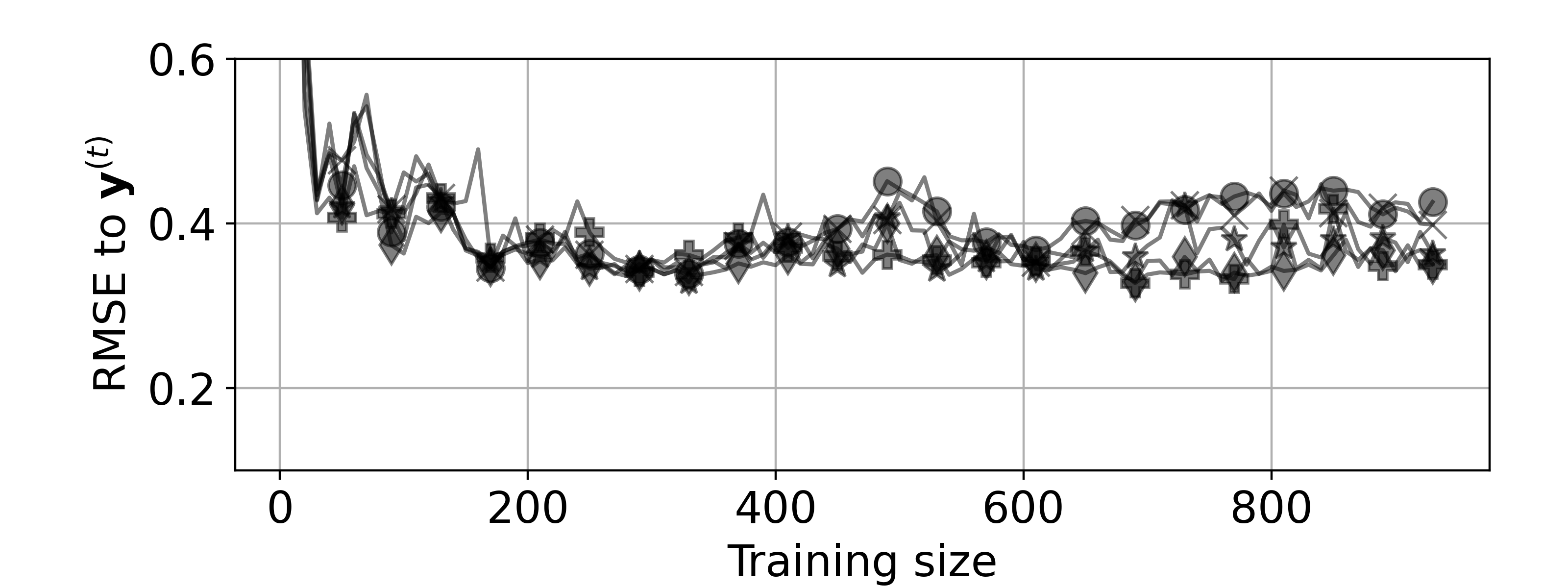}}
    \caption{Global relative RMSE comparison for different noise levels and 3 POD modes using POD-PCE-3DVAR}
    \label{fig:application:twin:PODPCE_noiseCompare_3modes}
\end{figure}

Firstly, noise impact is analyzed by fixing an arbitrary mode number of 3, corresponding to EVR $95\%$, for both POD-PCE and PODEn3DVAR methods. Using different noise levels (1, 5, 10, 20 and 40 $\%$), the relative RMSE between observed state $\mathbf{y}^{(o)}$ and analyzed model output $\mathbf{y}^{(a)}$ on the one hand, and between the truth $\mathbf{y}^{(t)}$ and analysis $\mathbf{y}^{(a)}$ on the other hand, are shown in Figure \ref{fig:application:twin:PODPCE_noiseCompare_3modes} with the POD-PCE-3DVAR method. It can be noticed that relative RMSE between assimilated state and observations increases with the noise injected in data. Oppositely, distance to truth stabilizes with the training size, whatever the noise level in the data, showing robustness of the used POD-PCE metamodel. \\

\begin{figure}[H]
  \centering
  \subfloat{\hspace{-0.3cm} \includegraphics[trim={1cm 6.5cm 0cm 0.3cm},clip,width=0.5\textwidth]{figs/noiseCompare_legend.png}} \\
  \vspace{-0.2cm}
	\setcounter{subfigure}{0}
  \includegraphics[trim={0.5cm 0cm 0cm 0cm},clip,width=0.5\textwidth]{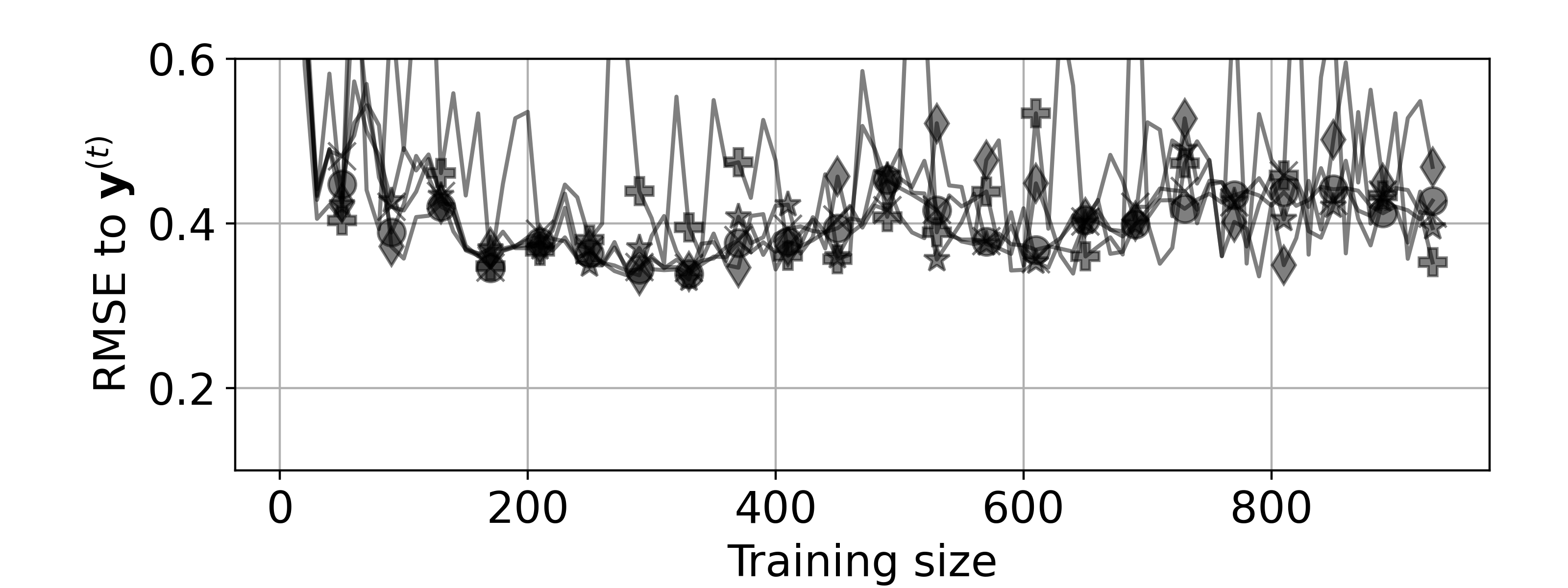}
    \caption{Global relative RMSE between analysis and truth for different noise levels and 3 POD modes using PODEn3DVAR}
    \label{fig:application:twin:PODEn3DVAR_noiseCompare_3modes}
\end{figure}

Results with PODEn3DVAR are similar to Figure \ref{fig:application:twin:PODPCE_noiseCompare_3modes}-a regarding distance to observation. However, it is shown in Figure \ref{fig:application:twin:PODEn3DVAR_noiseCompare_3modes} that the obtained analysis using PODEn3DVAR is less robust to noise in the observation, as relative RMSE between analysis and truth is less stable than for POD-PCE-3DVAR in Figure \ref{fig:application:twin:PODPCE_noiseCompare_3modes}-b. Additionally, PODEn3DVAR struggles to converge, which is in particular true for the highest noise levels.  

Comparison between POD-PCE-3DVAR and PODEn3DVAR can also be made for each output variable and observation point independently. The same behavior to noise as for global RMSE is then observed. For example, free surface error with POD-PCE-3DVAR stabilized around $2.5 \%$ and varies by $\pm 1 \%$ depending on the noise level, which is much smaller than the variation interval of noise.

\subsubsection*{Influence of the mode number or EVR}

\begin{figure}[H]
  \centering
  \subfloat{\hspace{-0.3cm} \includegraphics[trim={1cm 5.6cm 0cm 0.3cm},clip,width=0.5\textwidth]{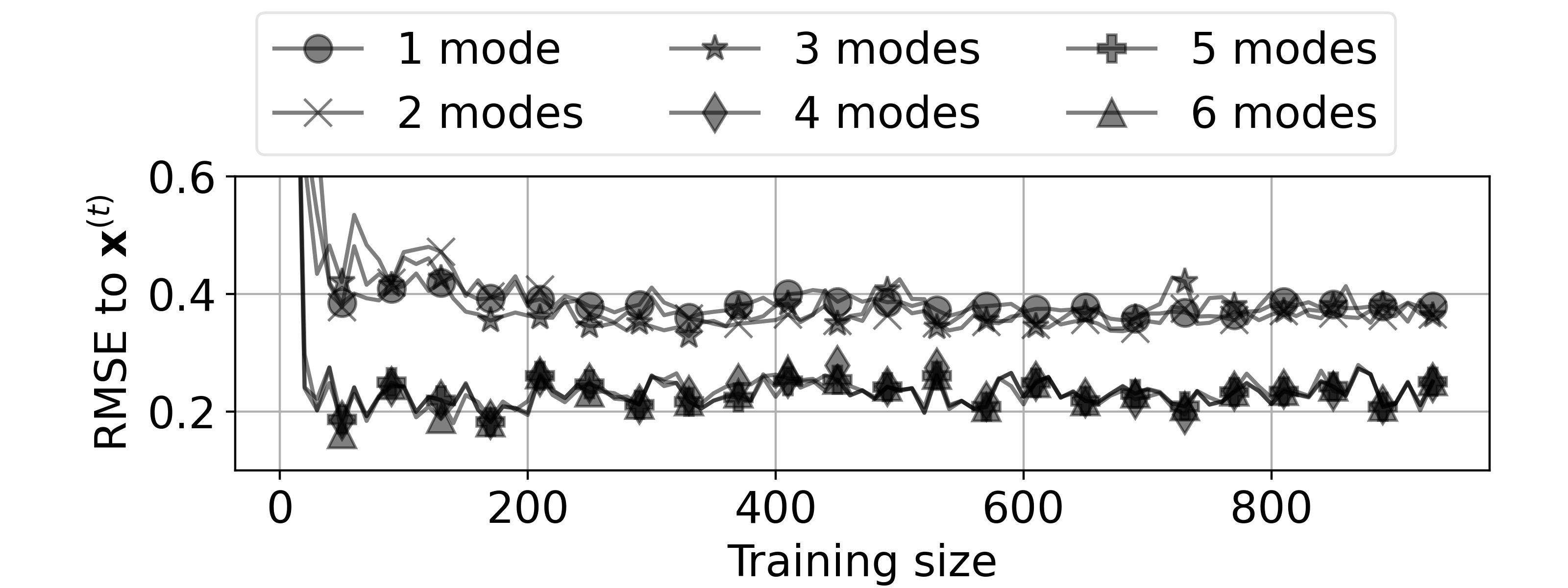}} \\
  \vspace{-0.4cm}
	\setcounter{subfigure}{0}
  \subfloat[][POD-PCE-3DVAR]{\includegraphics[trim={0.5cm 0cm 0cm 0cm},clip,width=0.5\textwidth]{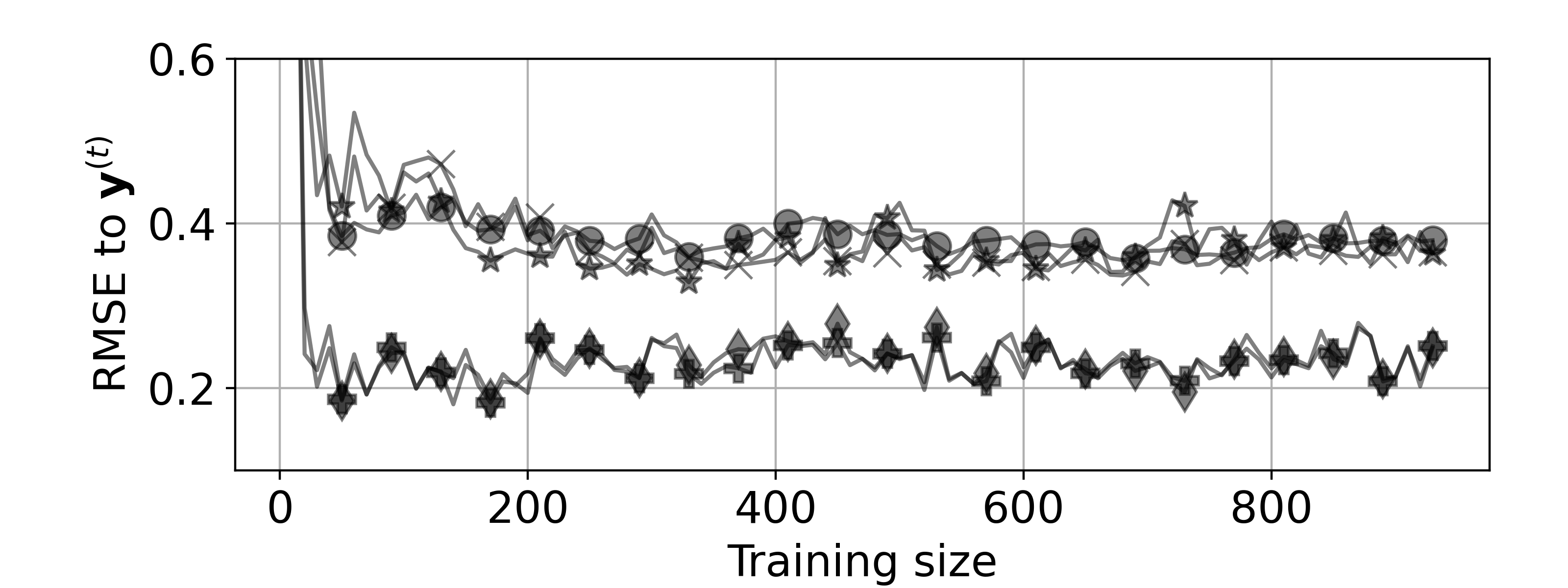}}
 \subfloat[][PODEn3DVAR]{\includegraphics[trim={0.5cm 0cm 0cm 0cm},clip,width=0.5\textwidth]{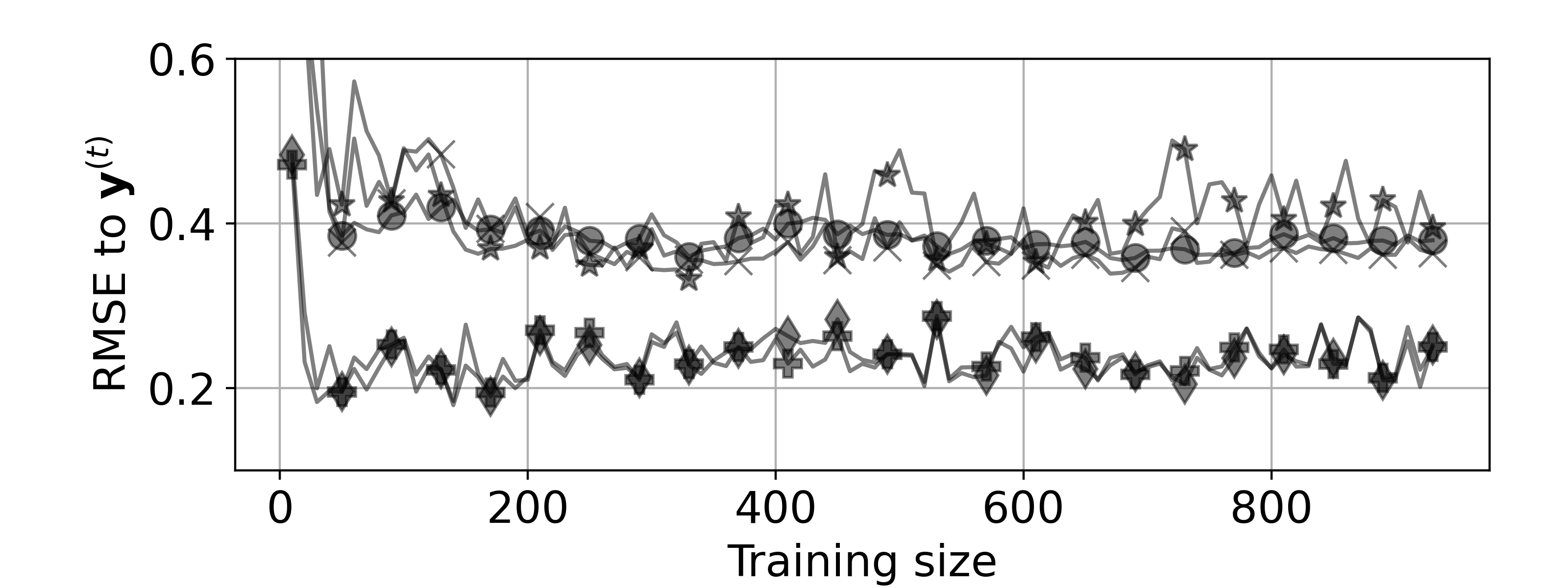}}
    \caption{Global relative RMSE between analysis and truth for different POD mode numbers, for the experiment with $10~\%$ noisy data, using POD-PCE-3DVAR and PODEn3DVAR}
    \label{fig:application:twin:modeCompare_Noise10}
\end{figure}

Sensitivity of the proposed algorithms to EVR or POD mode number choice is here studied. An experiment example with $10~\%$ of noise added to the observation is selected. Comparison of global RMSE, calculated between analyzed model output $\mathbf{y}^{(a)}$ and true state $\mathbf{y}^{(t)}$, is shown in Figure \ref{fig:application:twin:modeCompare_Noise10}.

For both methods, an RMSE step between metamodels using 3 modes or less and 4 modes or more is noticed. Use of 4 modes here seems optimal, as increasing the number beyond does not decrease the RMSE any longer. Comparison between POD-PCE-3DVAR and PODEn3DVAR methods in Figures \ref{fig:application:twin:modeCompare_Noise10}-a and \ref{fig:application:twin:modeCompare_Noise10}-b respectively shows that PODEn3DVAR may be less convergent than POD-PCE-3DVAR at higher mode numbers, for example here when 3 modes are used. This difference is also observed on variables errors, for example in the case of free surface elevation, where convergence is better guaranteed with lower number of modes (1 or 2) for both methods, although a gain of $1\%$ relative RMSE can be achieved by selecting 4 modes. The gain in accuracy is higher for the velocity components when selecting 4 modes, for example with POD-PCE-3DVAR, where average error decreases from around $4\%$ to $2\%$. Error decrease with the EVR is more significant for small ensembles.

\subsubsection*{Influence of the training members}
\begin{figure}[H]
  \centering
  \subfloat[][POD-PCE-3DVAR]{\includegraphics[trim={0.5cm 0cm 0cm 0cm},clip,width=0.25\textwidth]{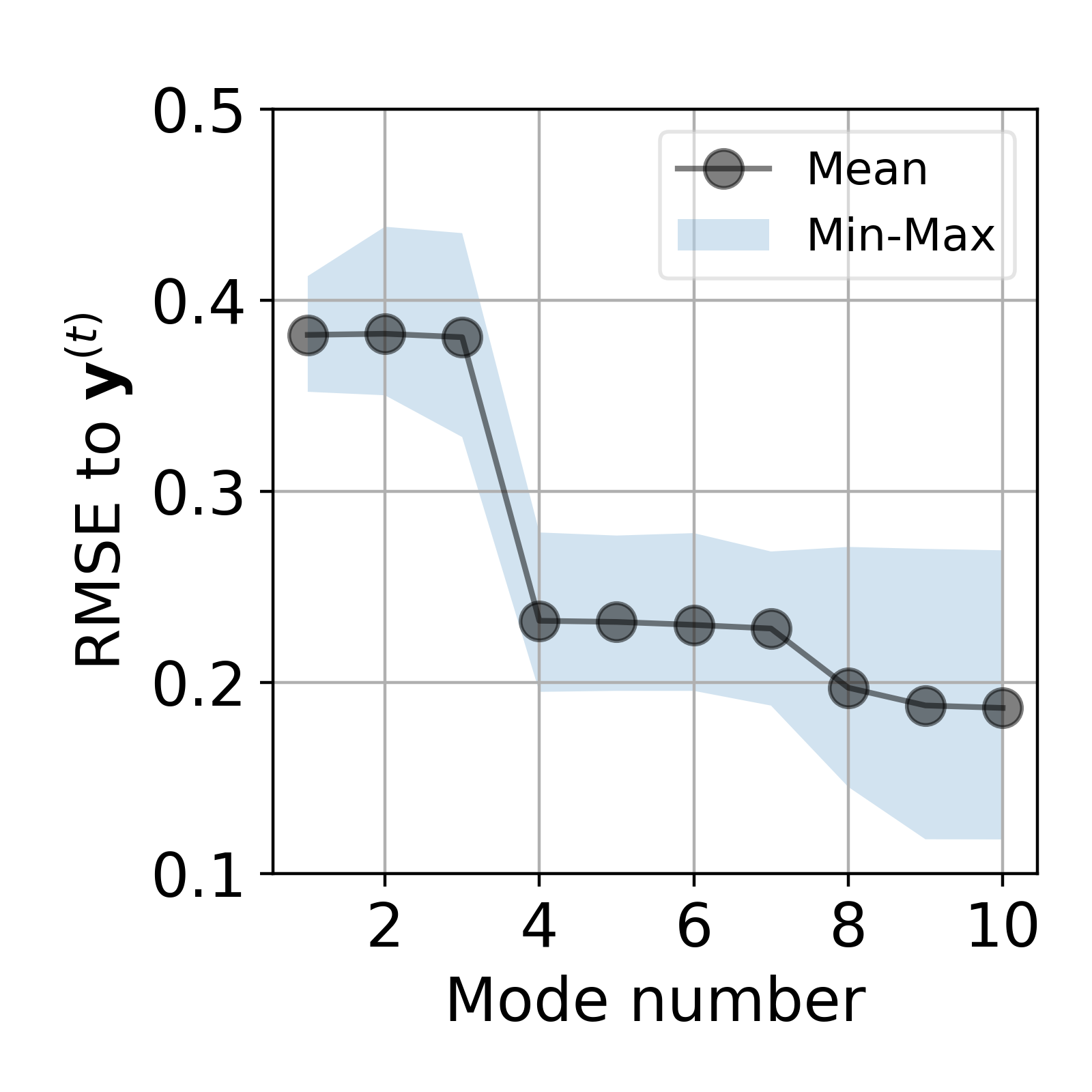}}
 \subfloat[][PODEn3DVAR]{\includegraphics[trim={0.5cm 0cm 0cm 0cm},clip,width=0.25\textwidth]{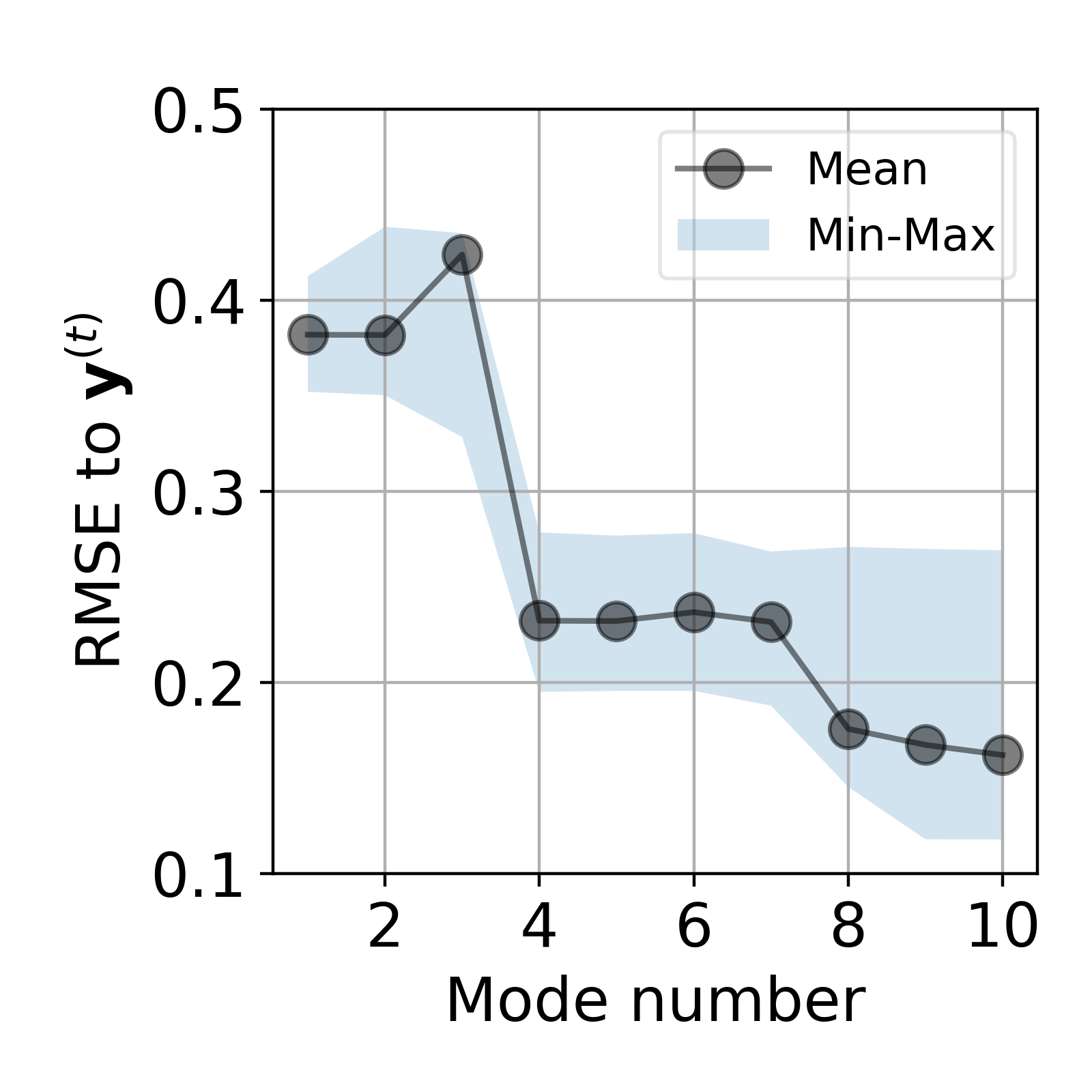}}
    \caption{Relative RMSE statistics with a bootstrap analysis on training members, for an ensemble of size $n=800$. Relative global RMSE is computed between assimilated and true states, for different POD modes numbers, using the experiment with $10~\%$ noisy data, with POD-PCE-3DVAR and PODEn3DVAR}
    \label{fig:application:twin:modeCompare_bootstrap_Noise10}
\end{figure}

Until now, the training set size was increased inclusively. This means that the ensemble of size $n=500$ for example, contains the ensemble of size $n=400$. The influence of training members was therefore not assessed. To do so, a bootstrap analysis is performed with size $n=800$, where $50$ realizations of training members are performed. At this ensemble size, both PODEn3DVAR and POD-PCE are considered to have converged. Sensitivity of the analysis relative global RMSE to truth is shown in Figure \ref{fig:application:twin:modeCompare_bootstrap_Noise10}. 

It can be seen in Figure \ref{fig:application:twin:modeCompare_bootstrap_Noise10} that POD-PCE-3DVAR and PODEn3DVAR globally follow the same trend, except for the 3-Mode metamodels where POD-PCE performs better in average. Additionnaly, for both methods, a widening of the min-max interval is noticed with the increasing mode number, which indicates greater sensitivity of the higher rank modes to training members.

\subsubsection*{Further investigations with POD-PCE-3DVAR: impact of error covariance matrices}

Previously shown comparisons shed light on robustness of POD-PCE-3DVAR to noise. It is therefore considered as best approach and selected for further investigation. For example, the impact of error covariance matrices choice on the quality of POD-PCE-3DVAR analysis is of interest, especially as used matrix $\mathbf{R}$ for model-observation uncertainties does not account for metamodelling biases. Hence, results with the POD-PCE-3DVAR method are here shown with assimilation experiments using $\widetilde{\mathbf{R}}$ instead of $\mathbf{R}$, as defined in Section \ref{section:materials}, allowing to consider metamodelling errors in the assimilation algorithm. \\

The example of $10\%$ noisy data is here chosen as a twin experiment. In order to assess the sensitivity of the method to error covariance matrix choice, matrices $\widetilde{\mathbf{R}}$ and $\mathbf{R}$, considered as optimal estimations, are multiplied by coefficients as $\alpha_R~\widetilde{\mathbf{R}}$ and $\alpha_B~\mathbf{B}$, where $\alpha_R~, \alpha_B \in \{0.01,0.1,1.0,10.0,100.0\}$, following the example shown in \citep{Tandeo2018}. Then, different assimilations are performed with each couple $(\alpha_B~\mathbf{B},\alpha_R~\widetilde{\mathbf{R}})$, and associated RMSE is computed. \\

\begin{figure}[H]
  \centering
  \includegraphics[trim={0cm 1.5cm 0cm 1.5cm},clip,width=0.4\textwidth]{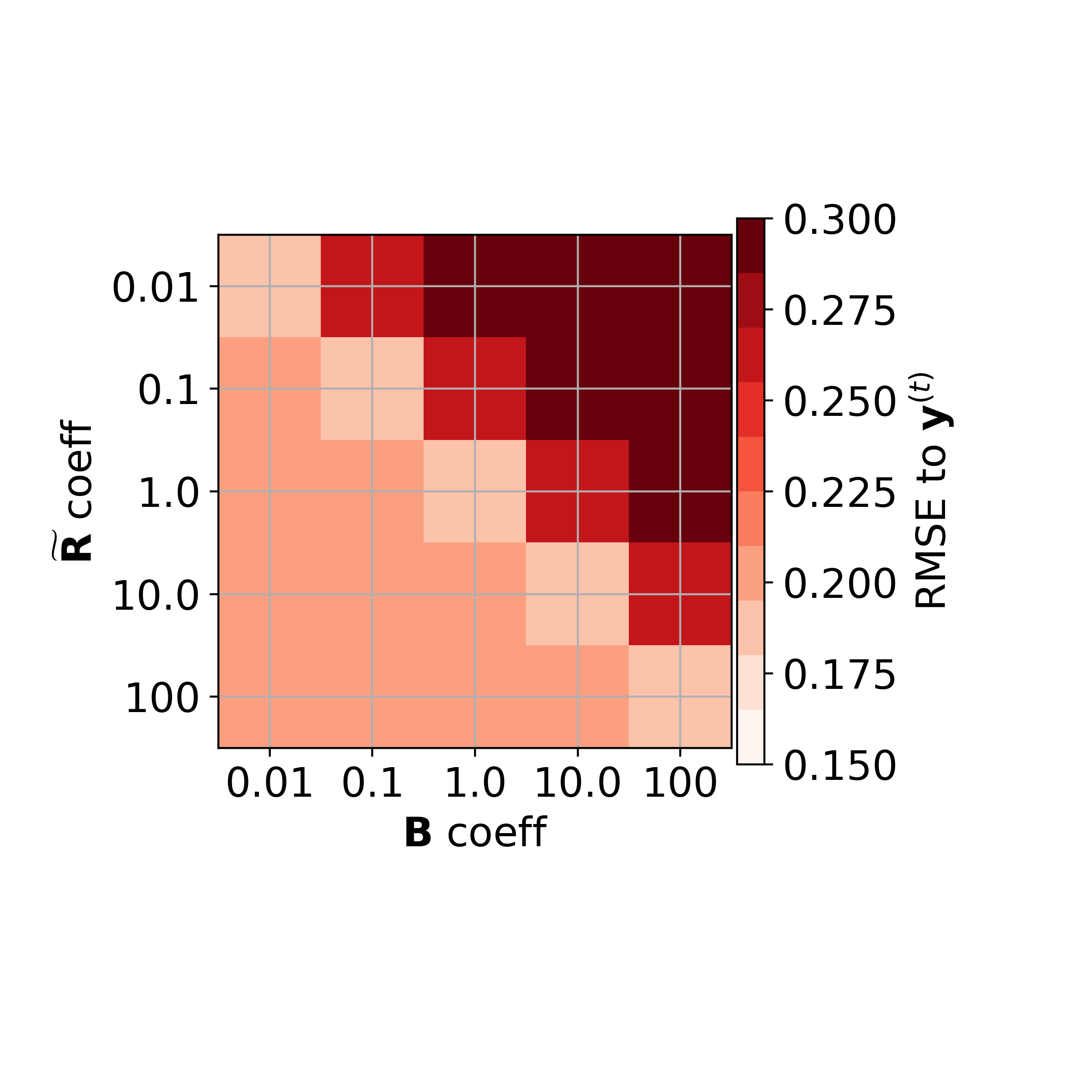}
    \caption{Global relative RMSE between analysis and truth for different error covariance matrices and a 5-Mode POD-PCE metamodel, for the experiment with $10~\%$ noisy data using POD-PCE-3DVAR}
    \label{fig:application:twin:covarianceCompare_Noise10}
\end{figure}

Relative global RMSE of analysis to truth is shown in Figure  \ref{fig:application:twin:covarianceCompare_Noise10}, for example for a 5-Mode POD-PCE metamodel. Firstly, it can be seen in Figure \ref{fig:application:twin:covarianceCompare_Noise10} that minimal RMSE corresponds to the diagonal, where same coefficients $\alpha_R = \alpha_B$ are applied, which allows to conserve the $\widetilde{\mathbf{R}}/\mathbf{B}$ ratio. Using the true definition of error covariance matrix, corresponding to $\alpha_B=1$, the error covariance matrix that gives minimal RMSE is $\alpha_R~\widetilde{\mathbf{R}}$ where $\alpha_R=1$. This means that $\widetilde{\mathbf{R}}$ is an optimal choice.  Additionally, underestimating $\widetilde{\mathbf{R}}$ with exact $\mathbf{B}$ ($\alpha_R<1$ and $\alpha_B=1$) or overestimating $\mathbf{B}$ with exact $\widetilde{\mathbf{R}}$ ($\alpha_R=1$ and $\alpha_B>1$) results with the greatest RMSE. These two correspond to cases where observation is given more weight compared to the background, in other words where the observation error compared to background error is undervalued. This behaviour is reported by other works, and attempts to correct the background error covariance matrix (which is often difficult to specify for DA on operational cases) in order to avoir its overestimation can for example be found in \citep{Cheng2019,Cheng2020error} (iterative process).
 
\begin{figure}[H]
  \centering
  \includegraphics[trim={0cm 0cm 0cm 0cm},clip,width=0.5\textwidth]{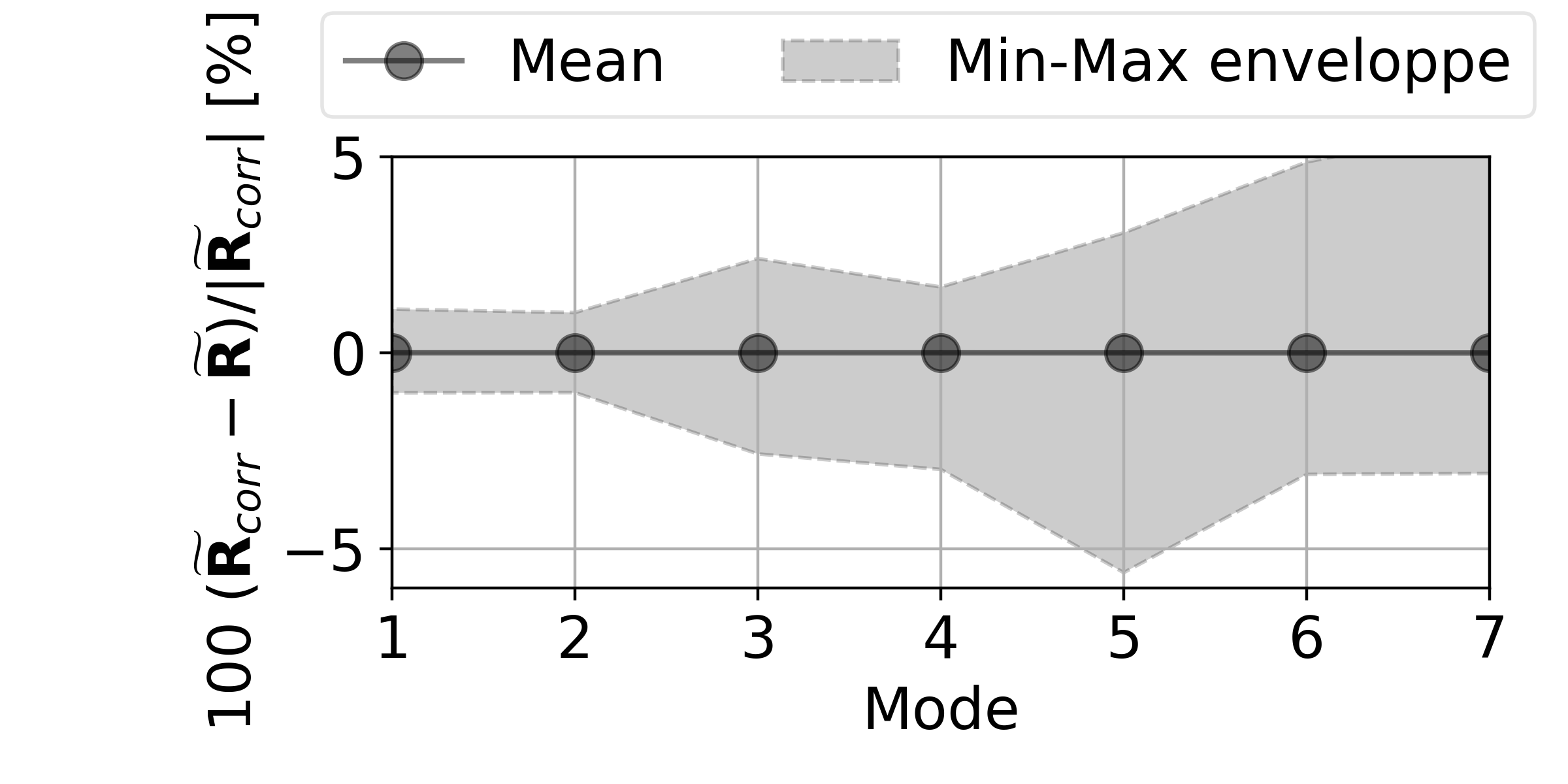}
    \caption{Evolution of telative error between $\widetilde{\mathbf{R}}_{corr}$ and $\widetilde{\mathbf{R}}$ with the POD mode number, in the POD-PCE-3DVAR case with the $10\%$ noisy twin experiment}
    \label{fig:application:twin:correctedCovarianceError}
\end{figure}

  It can also be noted that bias is sometimes present in the PCE learning. This results with an overestimation of PCE learning variances for each mode, as the expectation of model bias, denoted $\mathbb{E}\left[(\nu_k - \widetilde{\nu_k})\right]$, was neglected in Equation \ref{eq:materials:PODPCEDA:PCEvariance}. If $\mathbb{E}\left[(\nu_k - \widetilde{\nu_k})\right]$ is estimated using the test set for PCE learning, this gives a corrected error covariance matrix that is here denoted $\widetilde{\mathbf{R}}_{corr}$. The used error covariance matrix $\widetilde{\mathbf{R}}$ can either overestimate $\widetilde{\mathbf{R}}_{corr}$ or underestimate it, due to multiplication of PCE variances in Equation \ref{eq:metamodelCov} by POD basis elements $\boldsymbol{\Phi}^{(d)}$ that may contain negative values. Statistics of the relative error of $\widetilde{\mathbf{R}}$ to  $\widetilde{\mathbf{R}}_{corr}$, calculated over the matrix elements, is shown in Figure \ref{fig:application:twin:correctedCovarianceError}. It can be noticed that over and underestimation of the covariances become increasingly significant with the complexity of the model. This means that a bias in the PCE learning is recorded, and increases with the POD mode rank. Additionally, a widening of the min-max envelope is noticed for mode 3 compared to modes 1 and 2. 


\shorthandoff{:}

\subsection{Confrontation to classical 3DVAR on measurements}
\label{subsection:application:measurements}

In this subsection, the proposed PODEn3DVAR and POD-PCE algorithms are used for field measurements based calibration, and confronted to classical 3DVAR using the SWE. Measurements on a tidal period duration are used for fitting. The results are first shown with algorithms using $\mathbf{R}$ for different mode numbers, then compared in the POD-PCE-3DVAR case to results from algorithms using $\widetilde{\mathbf{R}}$. \\

\subsubsection*{Impact of POD mode number}
\begin{figure}[H]
  \centering
  \subfloat{\hspace{-0.3cm} \includegraphics[trim={1cm 5.6cm 0cm 0.3cm},clip,width=0.5\textwidth]{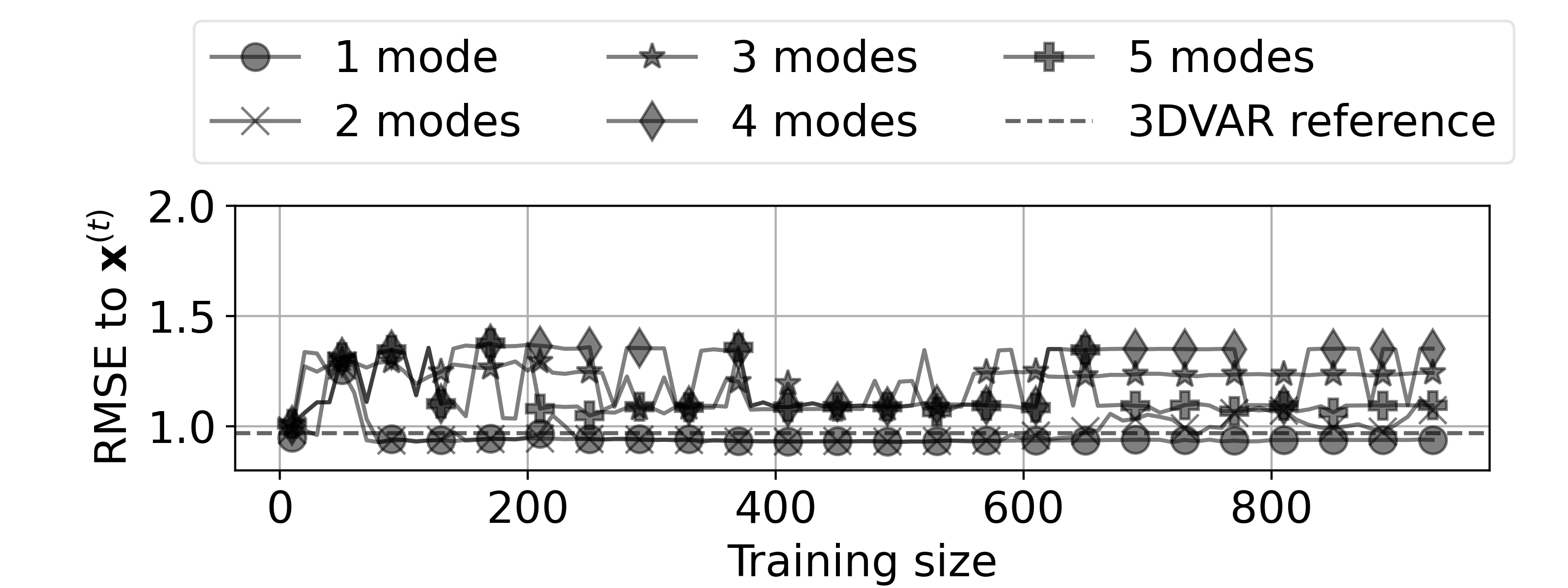}} \\
  \vspace{-0.4cm}
	\setcounter{subfigure}{0}
  \subfloat[][POD-PCE-3DVAR]{\includegraphics[trim={0.5cm 0cm 0cm 0cm},clip,width=0.5\textwidth]{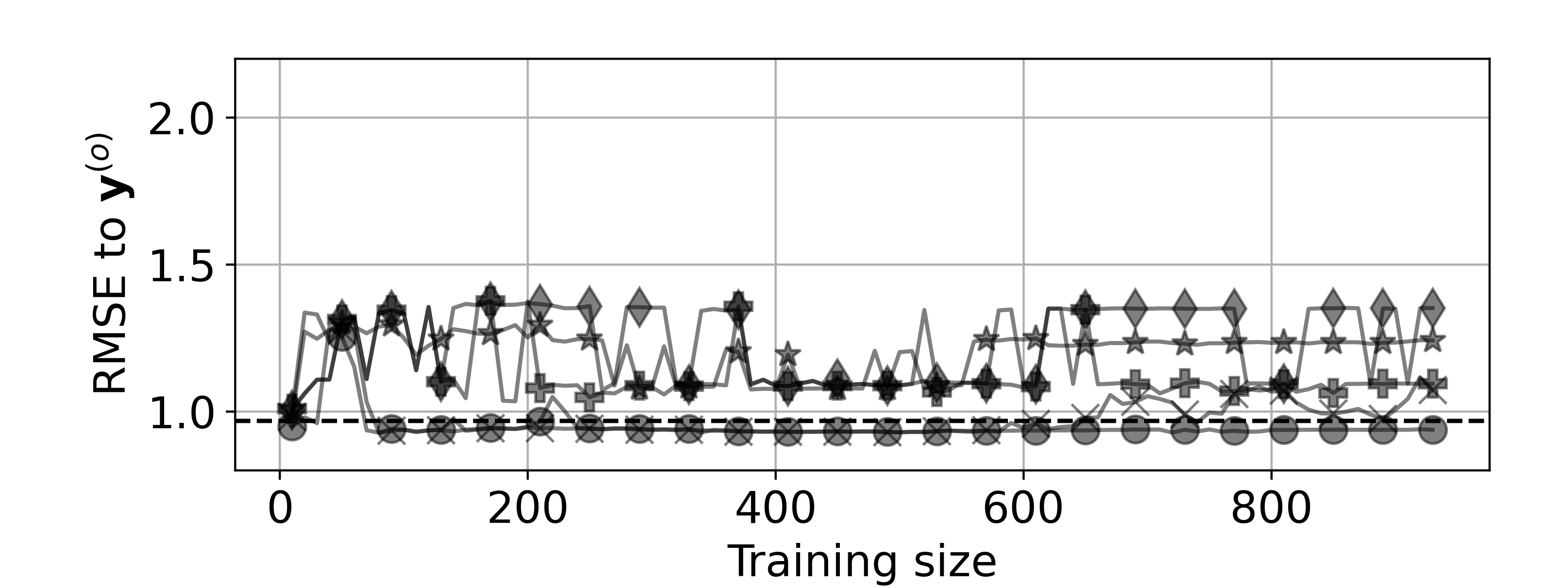}}
 \subfloat[][PODEn3DVAR]{\includegraphics[trim={0.5cm 0cm 0cm 0cm},clip,width=0.5\textwidth]{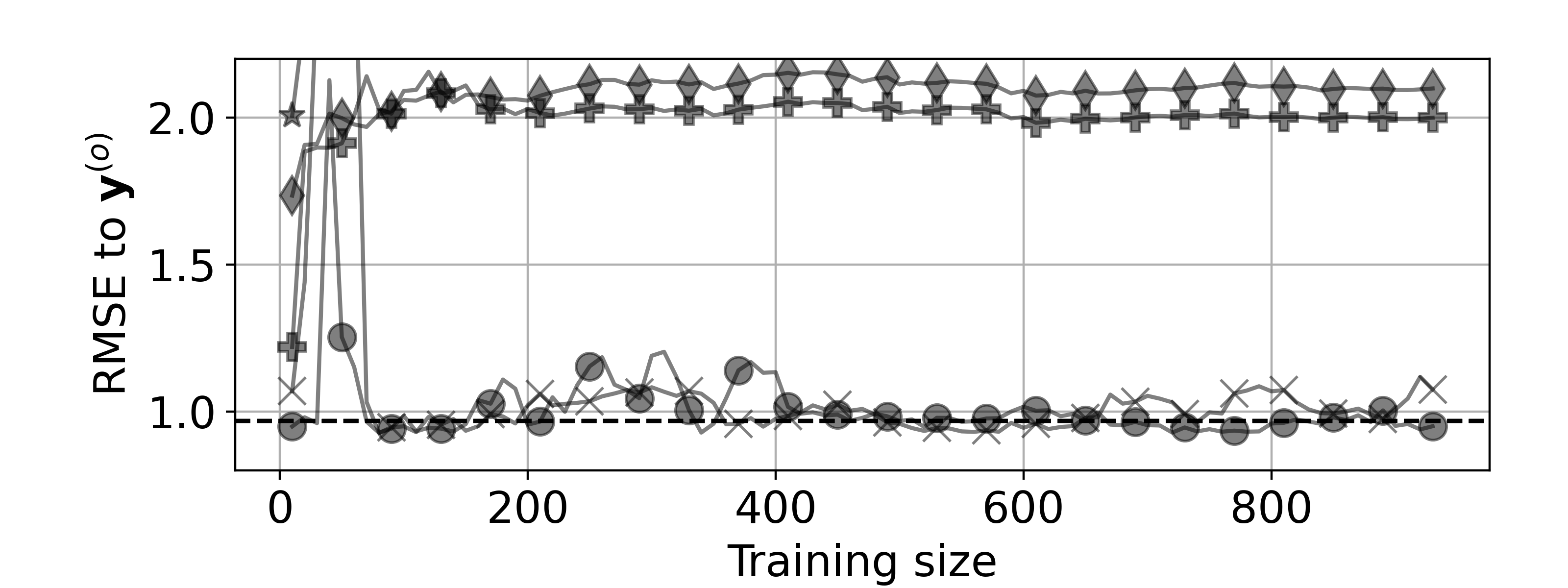}}
    \caption{Global relative RMSE between analysis and observation for different POD modes numbers, using POD-PCE-3DVAR and PODEn3DVAR applied to measurements, and comparison to classical 3DVAR results}
    \label{fig:application:measuerment:modeComparisonGlobalRMSE}
\end{figure}

Comparison of global relative RMSE with the used approaches for different POD mode numbers is shown in Figure \ref{fig:application:measuerment:modeComparisonGlobalRMSE}.

Firstly, increasing the mode number does not systematically come with decrease in RMSE, be it for POD-PCE-3DVAR in Figure \ref{fig:application:measuerment:modeComparisonGlobalRMSE}-a or PODEn3DVAR in Figure \ref{fig:application:measuerment:modeComparisonGlobalRMSE}-b. The most optimal configurations for both methods correspond to 1 or 2 POD modes. Additionally, a loss in convergence is noticed for high mode numbers with the POD-PCE-3DVAR method. Secondly, while convergence to reference classical 3DVAR result is fast with POD-PCE-3DVAR in Figure \ref{fig:application:measuerment:modeComparisonGlobalRMSE}-a, oscillations around the reference value are still noticed with PODEn3DVAR in Figure \ref{fig:application:measuerment:modeComparisonGlobalRMSE}-b, even for great ensemble sizes. It can be noted however that both methods reach their target. \\

\begin{figure}[H]
  \centering
  \subfloat{\hspace{-0.3cm} \includegraphics[trim={1cm 5.6cm 0cm 0.3cm},clip,width=0.5\textwidth]{figs/modesCompare_3DVAR_legend.png}} \\
  \vspace{-0.4cm}
	\setcounter{subfigure}{0}
  \subfloat[][POD-PCE-3DVAR]{\includegraphics[trim={0.5cm 0cm 0cm 0cm},clip,width=0.5\textwidth]{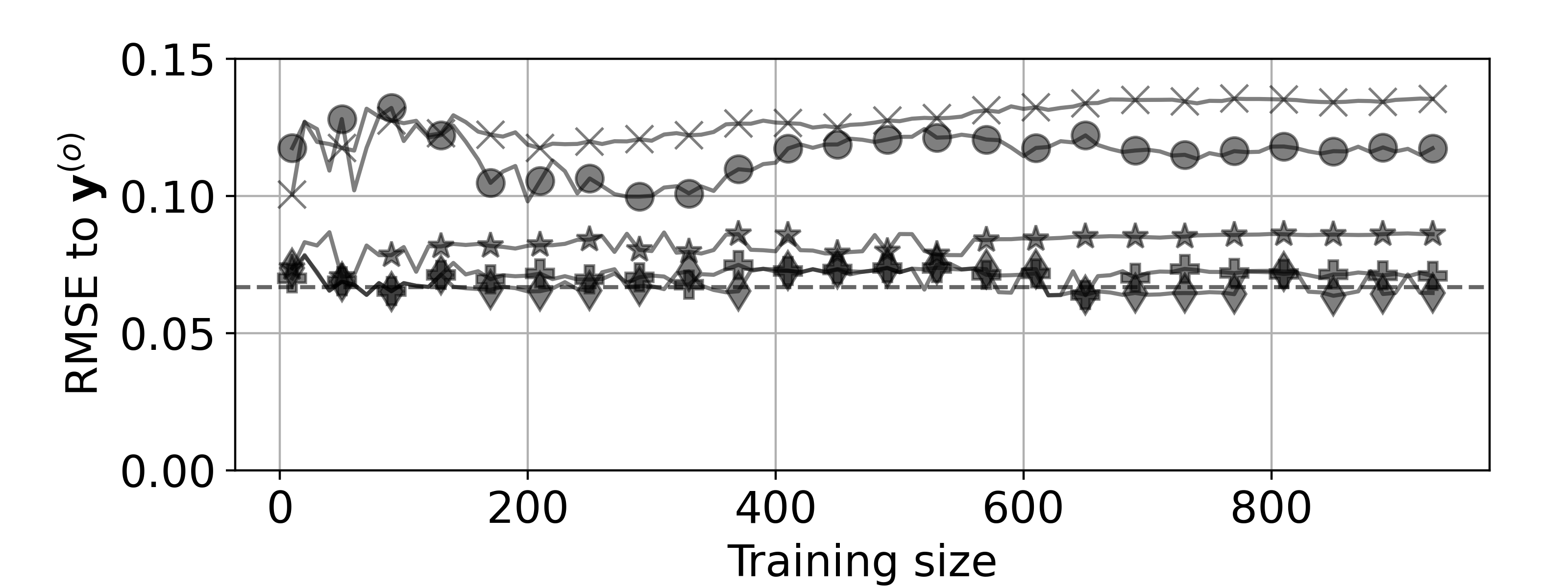}}
 \subfloat[][PODEn3DVAR]{\includegraphics[trim={0.5cm 0cm 0cm 0cm},clip,width=0.5\textwidth]{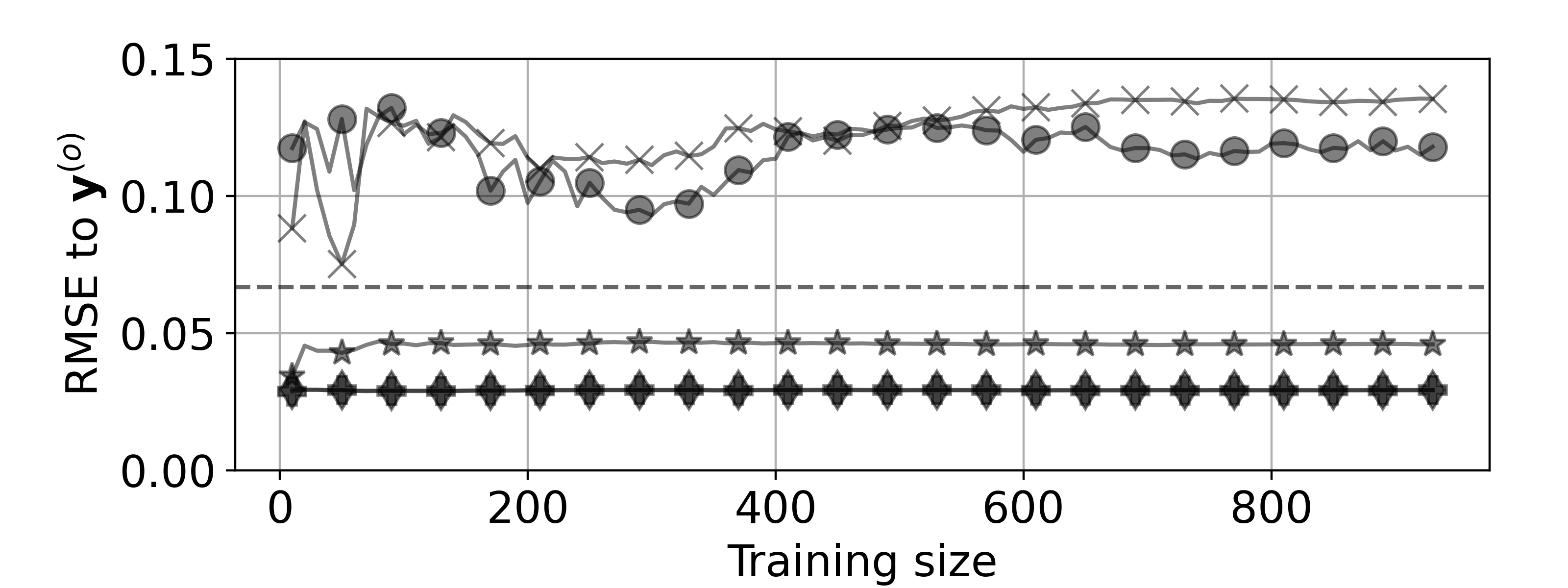}}
    \caption{Relative RMSE between assimilated and observed free surface elevation on Point 1, for different POD modes numbers, using POD-PCE-3DVAR and PODEn3DVAR applied to measurements, and comparison to classical 3DVAR results}
    \label{fig:application:measuerment:modeComparison_FREESURFACE_RMSE}
\end{figure}

\begin{figure}[H]
  \centering
  \subfloat{\hspace{-0.3cm} \includegraphics[trim={1cm 5.6cm 0cm 0.3cm},clip,width=0.5\textwidth]{figs/modesCompare_3DVAR_legend.png}} \\
  \vspace{-0.4cm}
	\setcounter{subfigure}{0}
  \subfloat[][POD-PCE-3DVAR]{\includegraphics[trim={0.5cm 0cm 0cm 0cm},clip,width=0.5\textwidth]{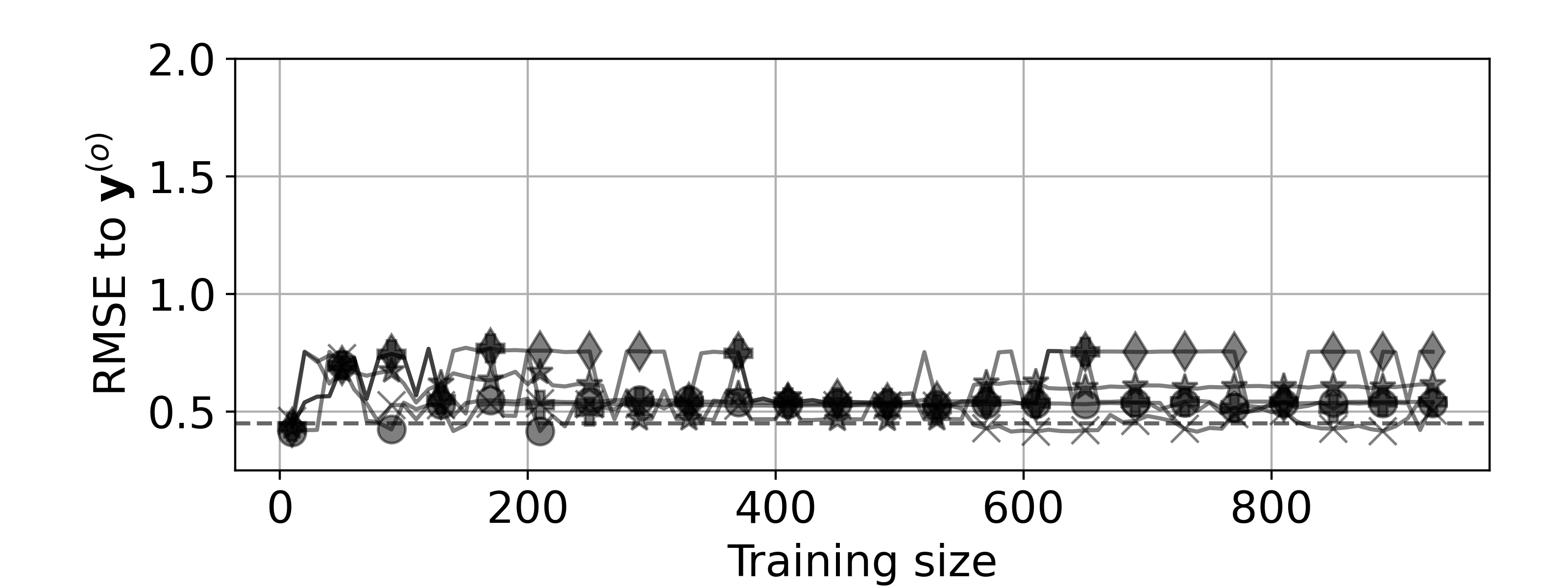}}
 \subfloat[][PODEn3DVAR]{\includegraphics[trim={0.5cm 0cm 0cm 0cm},clip,width=0.5\textwidth]{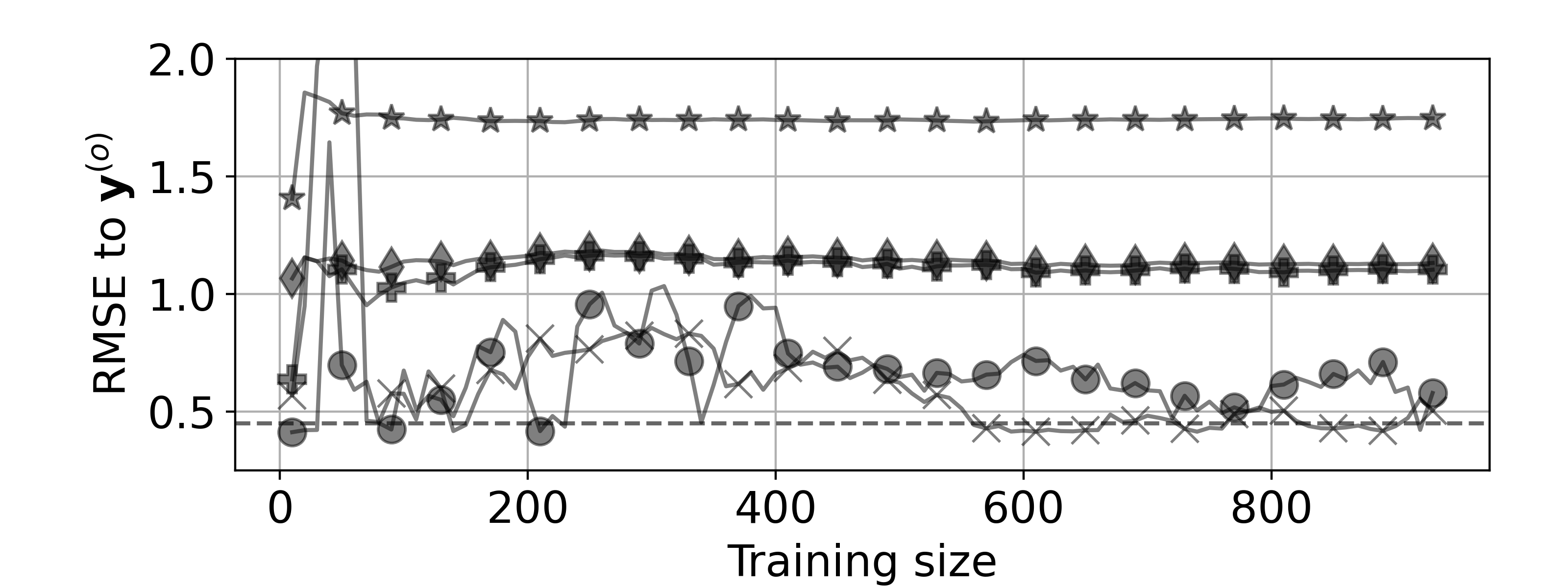}}
    \caption{Relative RMSE between assimilated and observed x-velocity $u$ elevation on Point 1, for different POD modes numbers, using POD-PCE-3DVAR and PODEn3DVAR applied to measurements, and comparison to classical 3DVAR results}
    \label{fig:application:measuerment:modeComparison_VELOCITYU_RMSE}
\end{figure}

Relative RMSE comparisons for specific variables are also studied. Free surface results on Point 1 using different mode numbers are for example shown in Figure \ref{fig:application:measuerment:modeComparison_FREESURFACE_RMSE}, while x-velocity results are shown in Figure \ref{fig:application:measuerment:modeComparison_VELOCITYU_RMSE}. For free surface in Figure \ref{fig:application:measuerment:modeComparison_FREESURFACE_RMSE}, a 4-Mode approximation seems optimal with both methods. However, this configuration, although interesting for free surface, gives higher RMSE for velocity as can be seen in Figure \ref{fig:application:measuerment:modeComparison_VELOCITYU_RMSE}, in particular for PODEn3DVAR in Figure \ref{fig:application:measuerment:modeComparison_VELOCITYU_RMSE}-b, and suffers from convergence issues with POD-PCE-3DVAR in Figure \ref{fig:application:measuerment:modeComparison_VELOCITYU_RMSE}-a. For the more challenging velocity variable, it can be noticed that while POD-PCE-3DVAR results fastly converged to reference 3DVAR in Figure \ref{fig:application:measuerment:modeComparison_VELOCITYU_RMSE}-a, PODEn3DVAR in Figure \ref{fig:application:measuerment:modeComparison_VELOCITYU_RMSE}-b needs greater ensemble sizes. \\

POD-PCE-3DVAR outperforms PODEn3DVAR for velocity analysis in Figure \ref{fig:application:measuerment:modeComparison_VELOCITYU_RMSE}, but the inverse is noticed free-surface analysis in Figure \ref{fig:application:measuerment:modeComparison_FREESURFACE_RMSE}, with $2.5\%$ of RMSE decrease. However, analyzing its parameters fitting shows that it may result with non-physical conclusions (parameters values negative when they should be positive), particularly with high mode number and low ensemble sizes. This difference with POD-PCE-3DVAR can be due to the use of a constrained descent algorithm for the latter (imposed bounds for parameters with c-BFGS-QN \citep{Zhu1997}), or to a poor learning of parameters-modes causality with a completely linear model in PODEn3DVAR. \\

Lastly, as previously highlighted, both methods result with good approximates of classical 3DVAR analysis when convergence is reached. They can additionally represent a run time advantage with the adequate computational resources.  Indeed, an optimized gradient algorithm can for example investigate the influence of change in all parameters simultaneously by running $v+1$ simulations, where $v$ is the number of uncertain parameters to calibrate. If $Nc$ is the number of cores necessary for a single run, then $(v+1)\times Nc$ cores are necessary for an optimal gradient descent. When then number of available processors greatly exceeds the latter (e.g. computer cluster), it could be more interesting to perform a larger batch of simulations to optimally benefit from the available computational resources, then learn a metamodel that maps the computed variations, and use it to estimate the gradient at lower cost. This is exactly the principle of introduced POD-PCE-3DVAR and PODEn3DVAR approaches. In the studied case for example with $4$ uncertain parameters, $25$ processors were necessary for each simulation in order to optimize the run time (approx. $8$ hours). Hence, a maximum of $(4+1) \times 25 = 125$ cores can be used simultaneously by an optimized gradient descent, while $1000$ were available for use. This makes possible to simultaneously run an ensemble of $1000/25 = 40$ scenarios instead of only $5$, hence allowing to learn a metamodel that runs in seconds to perform better 3DVAR analysis for the same run time (complements in \hyperref[Appendix:A]{Appendix A}). However, if computational resources are a concern, then classical 3DVAR should be preferred as it reaches an optimum with less model runs, although the minimization process takes more time (\hyperref[Appendix:A]{Appendix A}). In consequence, choice between classical 3DVAR and metamodel based 3DVAR could be led by the importance of computational time vs. computational resources.

\subsubsection*{Further investigations with POD-PCE-3DVAR: impact of the error covariance matrices}

As with the twin experiment in Section \ref{subsection:application:twin}, POD-PCE-3DVAR outperforms PODEn3DVAR: it is characterized with faster convergence to reference 3DVAR analysis, and it better estimates the most challenging state variable (velocity). Hence, it can be selected as most promising algorithm, for further investigation and optimization. Here, once again, the impact of model-observation error covariance matrix choice is studied. Results are now analyzed for POD-PCE-3DVAR with a cost function using $\widetilde{\mathbf{R}}$ instead of $\mathbf{R}$, defined in Section \ref{section:materials}, in order to account for metamodelling error in the assimilation algorithm. Two elements should however be noted at this step: 
\begin{itemize}
    \item[$\bullet$] conversely to the twin experiments, the model can not be considered perfect here. Although $\widetilde{\mathbf{R}}$ is supposed to better account for metamodelling errors compared to $\mathbf{R}$, it does not however recover bad consideration of model errors;
    \item[$\bullet$] the "true" field is not available here, as the experiment concerns field measurements. 
\end{itemize}

\begin{figure}[H]
  \centering
  \vspace{-0.3cm}
  \subfloat{\hspace{-2.5cm} \includegraphics[trim={0cm 5.5cm 0cm 0.5cm},clip,width=0.7\textwidth]{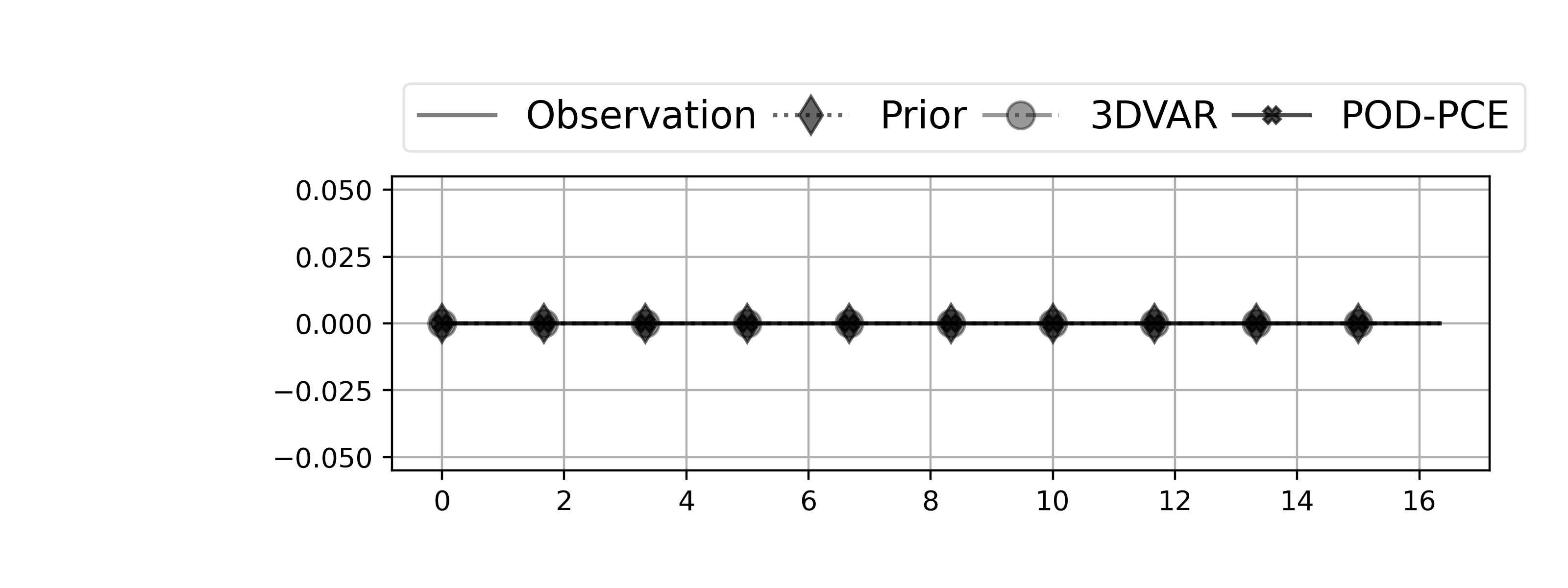}} \\
	\setcounter{subfigure}{0}
  \subfloat[][Free surface]{\includegraphics[trim={0.5cm 0cm 0cm 0.5cm},clip,width=0.33\textwidth]{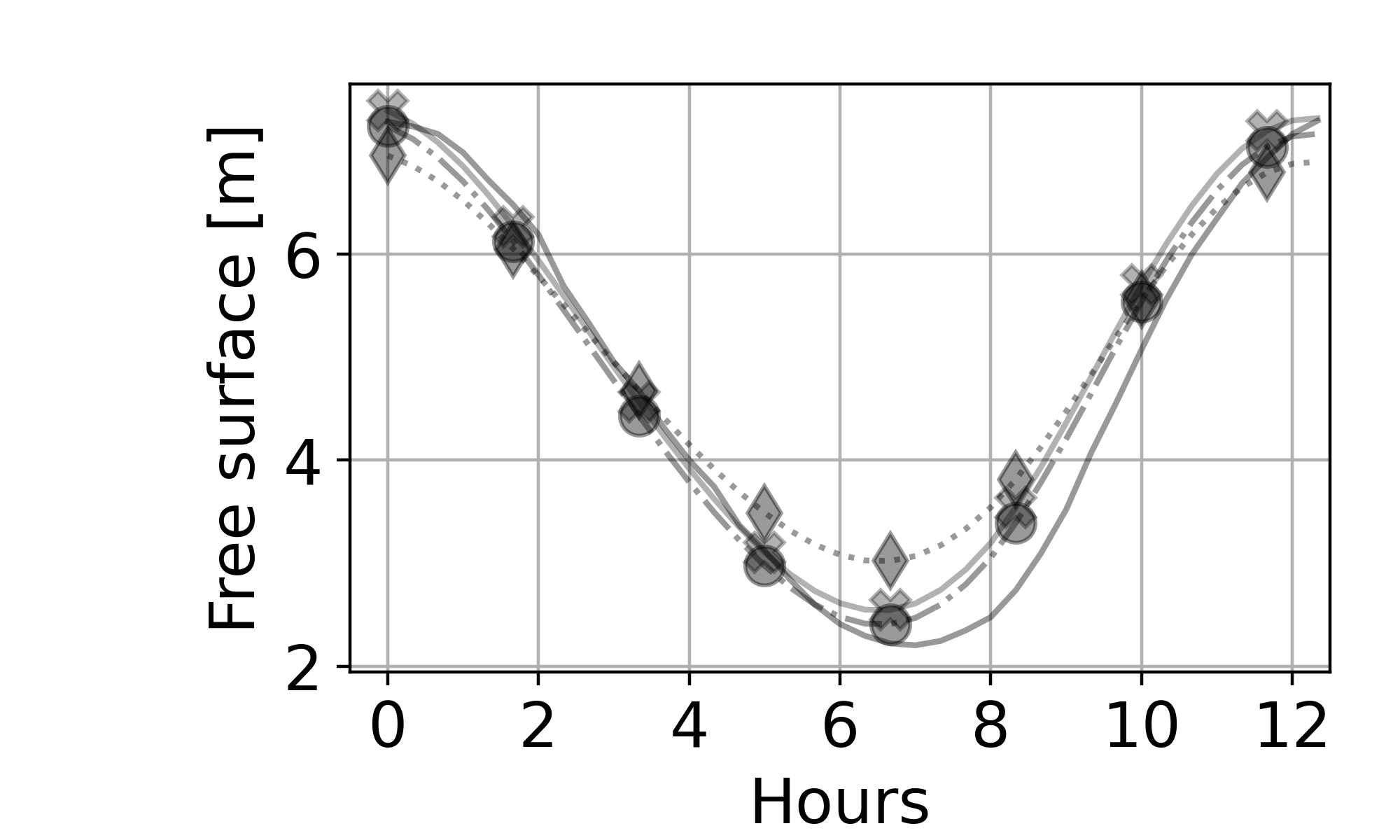}}
 \subfloat[][X-velocity]{\includegraphics[trim={0.5cm 0cm 0cm 0.5cm},clip,width=0.33\textwidth]{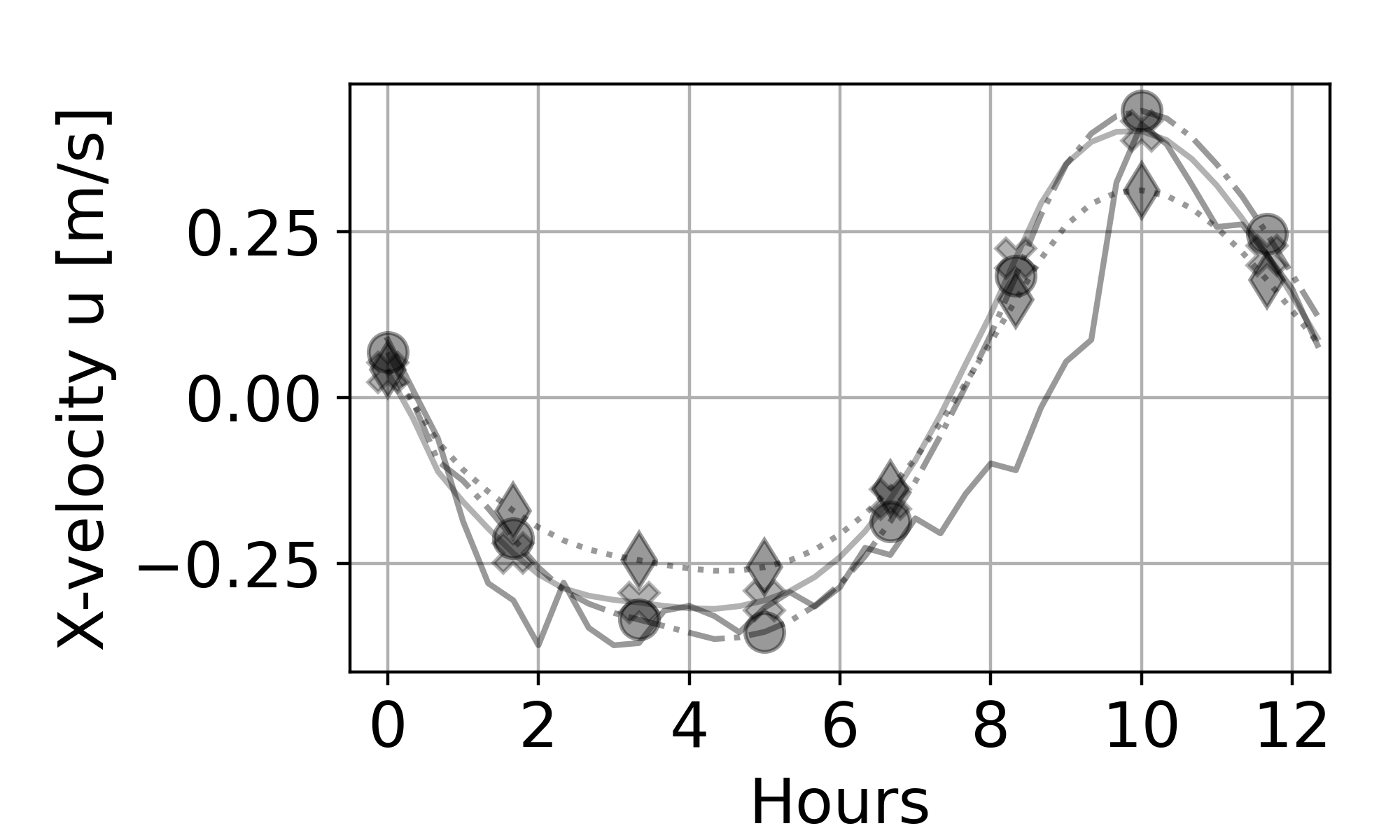}}
 \subfloat[Y-velocity]{\includegraphics[trim={0.5cm 0cm 0cm 0.5cm},clip,width=0.33\textwidth]{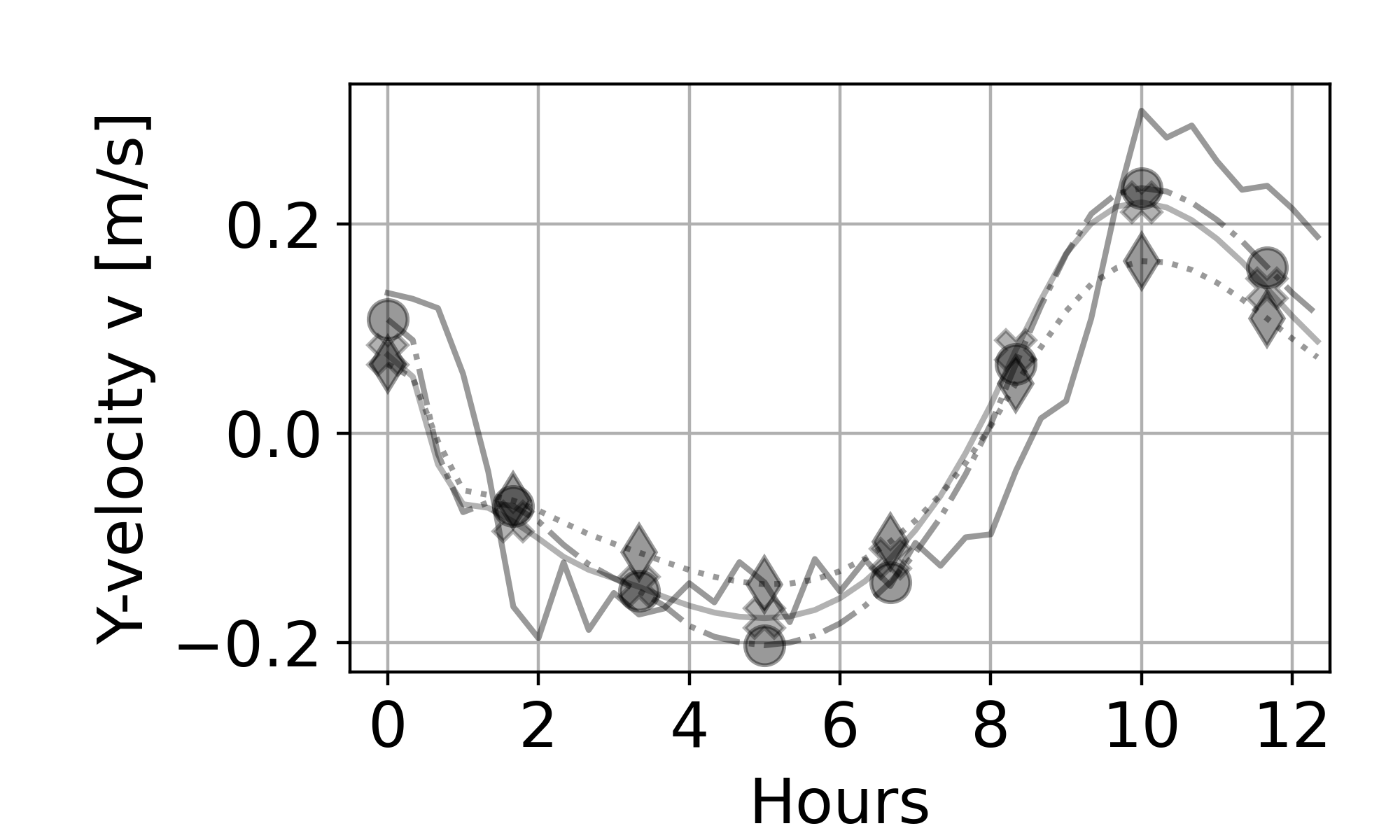}}
    \caption{Examples of fitting using the POD-PCE algorithm with $90~\%$ EVR}
    \label{fig:application:measuerment:fitting}
\end{figure}

 Comparison of measurements to optimal fitting can help shed light on the model incapacity to perfectly fit observations. As can be seen with the fitting example in Figure \ref{fig:application:measuerment:fitting}, a phase shift can be noticed in the free-surface measurements compared to model, as well as an asymmetry in the velocity evolution. This might for example be due to the imposed BC through the TPXO data-base, implying interpolation bias due to the use of coarsely defined data-base points compared to model scale, and incorporating significant simplifications as the effects of storm and surge (atmospheric and wave setup) are not modelled, although known to impact currents at local scale \citep{Idier2019}. Estimating model errors is however not an easy task, and perfect model assumption is used in practice. It can be noted however that even without error covariance matrix correction, the POD-PCE results in Figure \ref{fig:application:measuerment:fitting} are almost identical to classical 3DVAR solution.

For the reasons above, reduction of the distance to observation, i.e. decrease in RMSE, is not targeted in absolute terms. However, reduction in RMSE differences between POD-PCE-3DVAR analysis and classical 3DVAR analysis must in principle be achieved, as use of $\widetilde{\mathbf{R}}$ instead of $\mathbf{R}$ is supposed to compensate for the metamodel approximations, compared to full model $\mathcal{G}$. Changes in convergence and mode number sensitivity are therefore explored below. \\

\begin{figure}[H]
  \centering
  \subfloat{\hspace{-0.3cm} \includegraphics[trim={1cm 5.6cm 0cm 0.3cm},clip,width=0.5\textwidth]{figs/modesCompare_3DVAR_legend.png}} \\
  \vspace{-0.4cm}
	\setcounter{subfigure}{0}
  \subfloat{\includegraphics[trim={0.5cm 0cm 0cm 0cm},clip,width=0.5\textwidth]{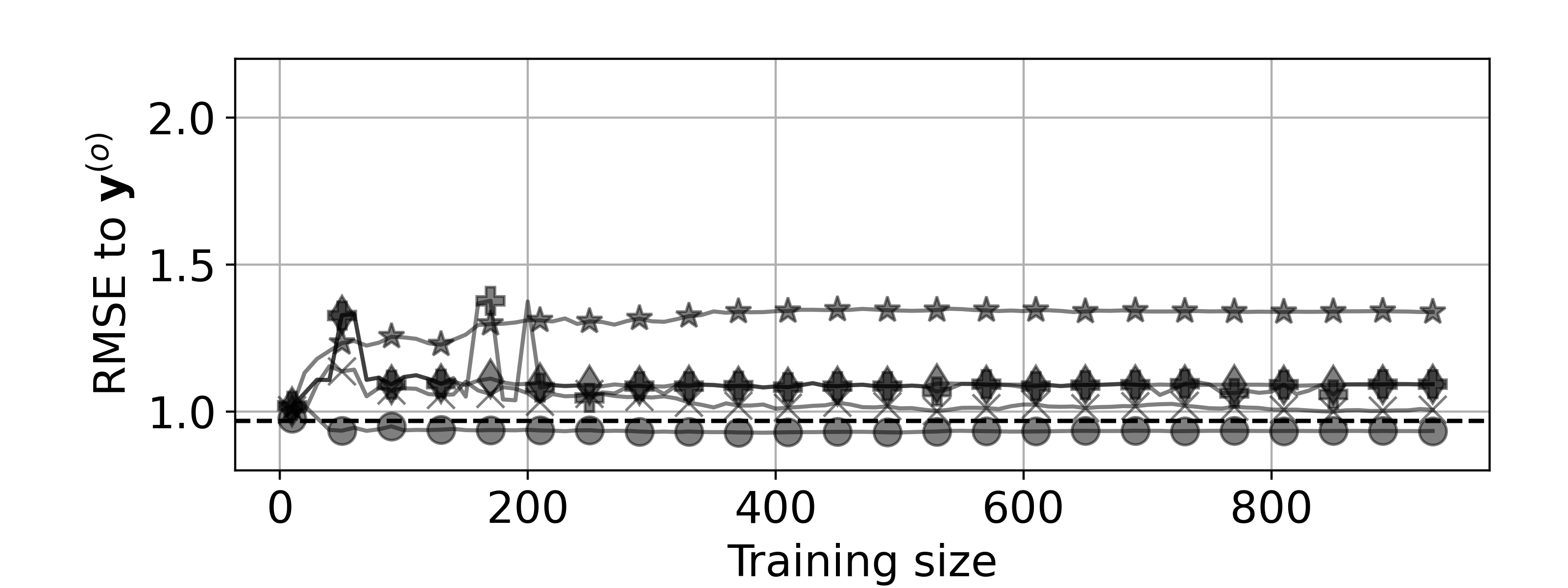}}
    \caption{Global relative RMSE between analysis and observation for different POD modes numbers, using POD-PCE-3DVAR with defined metamodelling error $\widetilde{\mathbf{R}}$ applied to measurements, and comparison to classical 3DVAR results}
    \label{fig:application:measuerment:modeComparisonGlobalRMSE_metamodelR}
\end{figure}

Global relative RMSE is shown in Figure \ref{fig:application:measuerment:modeComparisonGlobalRMSE_metamodelR}. Compared to the results of Figure \ref{fig:application:measuerment:modeComparisonGlobalRMSE}-a, a gain in convergence speed is noted, in particular when using high mode number (over 4). The metamodels reach the reference 3DVAR results much faster, and stabilize around it at lower ensemble sizes. A different behavior is however noted with the 3 mode metamodel, where no improvement of RMSE is noticed. This can be compared to the results of Section \ref{subsection:application:twin} with the twin experiments, interpreted in terms of overestimated PCE models variances (neglected bias). Perhaps, for some high rank modes, strong non-linearity may be observed, which demands finer discretization of the parameter space for better quality metamodelling. This can for example be fulfilled with moderate ensemble sizes by using an iterative DA algorithm with analysis perturbation. Relative RMSE was also analyzed for free surface and velocity components independently (not displayed here), showing comparable results where use of $\widetilde{\mathbf{R}}$ comes with faster convergence.



\shorthandoff{:}

\section{Discussion and conclusion}
\label{subsection:application:discussion}

In the results Sections \ref{subsection:application:twin} and \ref{subsection:application:measurements}, the PODEn3DVAR and POD-PCE-3DVAR methods were confronted on a twin and a measurement-driven DA experiments respectively. In the following, the main conclusions from previously presented results are discussed in Subsection \ref{subsubsection:application:discussion:discussion} and summarized in \ref{subsubsection:application:discussion:summary}.

\subsection{Discussion}
\label{subsubsection:application:discussion:discussion}
\subsubsection*{POD-PCE-3DVAR vs. PODEn3DVAR}
It was shown that both methods reach optimal results if a sufficient ensemble is used for metamodel learning. However, the analysis provided by POD-PCE-3DVAR in the measurement-based experiment, converges much faster to reference 3DVAR than the analysis by PODEn3DVAR. This better convergence was also observed with the twin experiments when using high noise levels ($20$ or $40\%$).  \\

On the field measurements experiment, the particular case of free-surface elevation showed lower distance between observation and analysis with PODEn3DVAR compared to POD-PCE-3DVAR. However, it can be highlighted that decreasing the RMSE to observation is not a target in DA, since measurements are uncertain. This low distance to observation with PODEn3DVAR was even lower than with classical 3DVAR using the full model. Furthermore, it resulted with unrealistic estimations of parameters, that can be avoided by selecting a lower number of POD components (1 or 2), which in turn decreases the accuracy. These unrealistic estimations were avoided in the case of POD-PCE-3DVAR by using a constrained gradient descent algorithm. Indeed, POD-PCE-3DVAR gave better estimations, and has shown to be nearly as efficient as classical 3DVAR for all output variables, with reduced computational cost. PODEn3DVAR can also be tested using a descent instead of analytical gradient calculation, in order to limit the parameters bounds. Additionally, this would not increase the computational cost required to perform the optimization step. \\

Another interpretation of PODEn3DVAR inadequacy could be the difficulty of learning with the linear joint parameter-state POD metamodel, particularly for non-linear relationships. Moreover, the great discrepancies of the background (wide variation interval) make linearization difficult. Perhaps, if small perturbations around the background were used to generate the ensemble, then convergence would be faster. However, this does not assure a good coverage of parameters space in case of great uncertainty about prior parameters. It should be here noted that an improved version called adaptive PODEn4DVAR was proposed in \citep{Cao2007}, consisting in POD basis update using an outer loop, by iterative perturbation of the sample around the analysis, also performed in \citep{Mons2017}. This allows to generate an ensemble based on a small perturbation around the background (analysis of previous iteration), and better succeed in linearizing the relationships. Furthermore the adaptive PODEn4DVAR is said to be more stable to the ensemble size than other ensemble DA methods \citep{Mons2017}, and a limited sample of size $n=20$ is used to generate the POD basis at each step, said sufficient up to $10^5$ control variables \citep{Mons2017}. An interesting perspective would be to confront PODEn3DVAR and ensemble-POD-PCE-3DVAR, both with an iterative ensemble update. \\

Additionally, it was suggested in \citep{Altaf2013} that an improvement is possible if the cost function is estimated using the original model, while the adjoint is approximated in the reduced space \citep{Altaf2013}. Similarly, one could use the proposed PODEn3DVAR and POD-PCE-3DVAR approaches to perform faster calculation of descent directions, while still estimating the cost function and the analysis with the full model. \\
 
\subsubsection*{Robustness to noise}

Robustness tests to noise were conducted using the twin experiments. Five noise levels (1, 5, 10, 20 and 40 $\%$) were incorporated to the truth (perfect observation, equal to twin simulation result) in order to simulate measured observation. The tested POD-PCE-3DVAR and PODEn3DVAR are both robust to noise in the observations. It is shown that RMSE increase does not exceed $2\%$ even with a noise increase of $10~\%$. PODEn3DVAR shows however greater sensitivity than POD-PCE-3DVAR in case of high noise levels (20 and 40 $\%$). \\

POD in that sense plays a major role. Indeed, given that the states are simulated with few POD patterns, the resulting analysis is smooth, not characterized with local fluctuations. For example in \citep{Shenefelt2002}, a Singular Value Decomposition (SVD), on which POD extraction in discrete framework is based, is used as a smoother to delete noise from data, and therefore provide better estimations on a surface heat transfer inverse problem. Additionally, PCE was learned on smooth numerical model results. However, even in the case of perturbed training set,  robustness of PCE to noise was demonstrated in \citep{Torre2019} in pure Machine Learning setups, and confirmed in \citep{Mouradi2021} in the case of POD-PCE coupling. \\

\subsubsection*{POD modes number influence}

For both twin and measurement-based experiments, sensitivity of the results to the choice of POD components number was investigated. \\

It was shown that POD mode number increase may contribute in decreasing the analysis error, for both PODEn3DVAR and POD-PCE-3DVAR algorithms. This impact of chosen POD modes number on the performance of POD-based ensemble-variational methods was also highlighted by \citet{Cao2007}.  \\

However, the twin experiment case showed that increasing the mode number beyond 4 was useless. Therefore, an optimum it to be sought depending on the treated case. Furthermore, the analysis with higher number of POD modes is more sensitive to the ensemble members choice. \\

In the measurement-based experiment, it was shown that POD mode number increase does not systematically result in better analysis. Firstly, convergence issues can be met, particularly with ensembles of low size. The latter may indeed not be sufficient to learn the small variances patterns (high rank modes) efficiently. Adding them to the metamodel may therefore result in uncertain analysis. This difficulty was also highlighted in \citep{Mouradi2021} regarding POD expansion coefficients learning using PCE. Secondly, increasing the number of selected patterns has lead to unrealistic parameter estimations using PODEn3DVAR. The algorithm, by controlling too many POD expansion coefficients, results in over-fitting of the analysis on measurements, with the cost of unphysical parameter estimations. 
\subsubsection*{Impact of metamodel error covariance matrix}

An adapted estimation of metamodel error covariance matrix, resulting in an update of observation error covariance matrix $\mathbf{R}$ denoted $\widetilde{\mathbf{R}}$, was proposed and tested for POD-PCE-3DVAR. \\

In the twin experiment case, multiplicative coefficients in front of $\widetilde{\mathbf{R}}$ and $\mathbf{B}$ were used to test the sensitivity of the analysis to error covariance matrices choice. This helped identify the impact of error over- or under- estimation on the algorithm performance. The results are unequivocal: use of $\widetilde{\mathbf{R}}$ provides the most optimal analysis, compared to over- or under-estimation. \\

In the measurement-based application, use of the corrected error covariance matrix $\widetilde{\mathbf{R}}$ compared to $\mathbf{R}$ globally improved the analysis, making it closer to analysis by classical 3DVAR using the full model. Furthermore, convergence was greatly improved, in particular at high mode number. 

\subsubsection*{POD-PCE-3DVAR limitations and outlook}
 Firstly, direct POD-PCE coupling may not always be adequate for arbitrary physical problems, for example when dealing with discontinuity or when facing considerable non-linearities. For discontinuity treatment, RePOD (Registration POD) proposed by \citet{Taddei2020}, consisting in parametric smoothing of the discontinuous field prior to POD, can be used. Furthermore, adaptive Multi-Element PCE in \cite{WanKarniadakis2005} can be attempted, in which sub-learning problems are performed independently on subsets of the inputs space. In a Bayesian DA problem characterized with high non-linearity, \citet{Birolleau2014} show that representation problems may occur, with the Karhunen Loève Transform and PCE coupling proposed by \citet{Marzouk2009} for functional outputs. An iterative PCE method is therefore proposed as a solution, where interest field is first expanded using classical PCE, the resulting approximation used as a transform for the generation of a new PCE orthonormal basis, and so on. Remarkably, the expansion resulting from this iterative process performs much better in the presence of abrupt variations.  \\
 
 Secondly, use of PCE as metamodel may sometimes result with instabilities, which is highlighted by \citet{Despres2013} where PCE based reduction on the 1D SWE hyperbolic system generates non-physical oscillations that may grow in time, resulting from a loss of hyperbolicity.\\
 
  Lastly, authors in \citep{Cao2007} also mentioned the possible impact of snapshot sampling method on the performance of POD-based ensemble-variational methods. For more efficient PCE learning, in order to reduce the required ensemble size, Gauss-Hermite quadrature rule can be used to learn PCE coefficients instead of random MC sampling, as performed in \citep{Rochoux2014a,ElMocayd2017} to accelerate the Ensemble-based Kalman Filter. This however may not be optimal when dealing with high dimensional input spaces. \\ 

\subsection{Conclusion}
\label{subsubsection:application:discussion:summary}

In the presented study, two approaches for hybrid ensemble-variational
DA were compared, both relying on an ensemble-based surrogate, used to replace a time consuming numerical model in a variational cost function. Practical interest would be either to achieve complex non-linear DA in competitive times, or run multiple DA problems on a same case at lower cost. \\

To this end, a linear surrogate deduced by applying POD to joint control-state variables results with the PODEn3DVAR methodology, whereas a non-linear POD-PCE coupled metamodel results with POD-PCE-3DVAR. A new error covariance matrix, accounting for the POD-PCE metamodel approximation errors, was additionally estimated for POD-PCE-3DVAR, in order to improve the optimization process. Proposed approaches are convenient for a broad range of DA problems, for example inverse variables estimation, shape optimization, state reanalysis, etc. PODEn3DVAR and POD-PCE-3DVAR were here assessed for parametric calibration, on a tidal currents modelling case, in a coastal area. The case was challenging, because the fitting of three interest spatio-temporal outputs was targeted, and due to the presence of asymmetries in the tidal velocity components that are not captured by the model, making the DA process difficult. A twin and a measurement-based DA experiment were attempted and allowed investigating different properties. \\

Firstly, robustness of both methodologies to noise was demonstrated on the twin experiment. In particular, POD in both approaches helps providing a smooth analysis. POD-PCE-3DVAR performed better than PODEn3DVAR for high noise levels (20 and 40 $\%$) in the observation, and should be preferred in such cases. Secondly, analysis provided by POD-PCE-3DVAR is characterized by better convergence than PODEn3DVAR, compared to the truth for twin experiment, at higher noise levels. It also provides an analysis closer to reference classical 3DVAR results for the measurement-based experiment, with much faster convergence that would allow DA at lower ensemble sizes. This convergence was further improved using the updated error covariance matrix in particular at higher mode numbers. \\

Sensitivity of the results to the choice of POD components number was assessed for both POD-PCE-3DVAR and PODEn3DVAR. While increasing the complexity can improve the analysis, it may also, beyond a certain degree, result with non-converged metamodel learning and therefore uncertain analysis. Additionally, in the particular case of PODEn3DVAR with analytical gradient and analysis calculation, increasing the number of selected patterns has lead to unrealistic parameter estimations. \\

This work showed POD-PCE to be the most relevant choice for surrogate-based DA. It is both accurate, more robust to noise, and computationally efficient. Convergence is achieved faster than with PODEn3DVAR. Additionally, proposed metamodel error covariance calculation significantly increases the accuracy. It gives satisfactory results compared to classical 3DVAR, while requiring less run time. However, it should be kept in mind that classical 3DVAR is more interesting in terms of needed number of simulations for convergence. Hence, computational time vs. computational resources could lead the choice between classical 3DVAR and metamodel-based 3DVAR. \\

Proposed perspectives for the POD-PCE-3DVAR methodology are also promising and could result with better accuracy. As an example, an adaptive POD basis update by iterative perturbation around the analysis could be attempted as done in \citep{Cao2007} with PODEn4DVAR, which may both further reduce the uncertainties and work with smaller ensembles \citep{Mons2017}. Additionally, the POD-PCE metamodel could be adapted for particular cases of discontinuity \citep{WanKarniadakis2005}, or considerable non-linearities \citep{Birolleau2014}, which are bottlenecks of metamodelling of complex physical cases, and would help applying the POD-PCE-3DVAR methodology to a great variety of challenging DA problems.


\section*{Acknowledgements}
This work was funded by the French National Association of Research and Technology (ANRT) and EDF R\&D with the Industrial Conventions for Training through REsearch (CIFRE grant agreement 2017/1452). The authors acknowledge their support, and are grateful for data collection and feedback from EDF operators. The authors gratefully acknowledge the TELEMAC-MASCARET (environmental numerical modelling) and OpenTURNS (Uncertainties treatment python library) open source communities, in particular within EDF R\&D, as well as the ADAO (Data Assimilation python library) developper teams, more precisely J.P. Argaud and A. Ponçot from EDF R\&D.

\shorthandoff{:}

\section*{Appendix A. Simulation times}
\label{Appendix:A}

\begin{figure}[H]
  \centering
 \subfloat[][Metamodel learning time]{\includegraphics[trim={0.8cm 0cm 0.5cm 0.5cm},clip,width=0.4\textwidth]{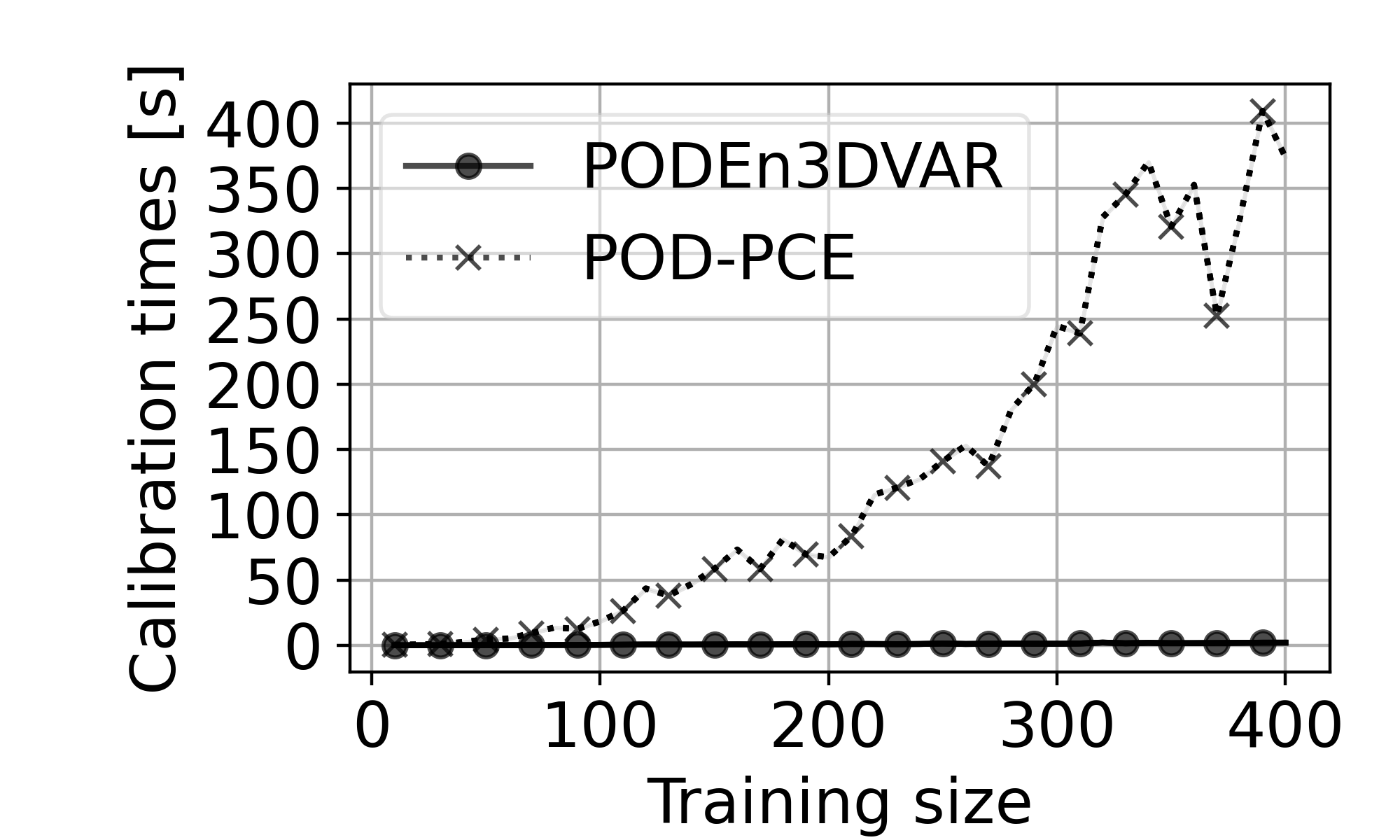}}
 \subfloat[][Metamodel-based 3DVAR time]{\includegraphics[trim={0.8cm 0cm 0.5cm 0.5cm},clip,width=0.4\textwidth]{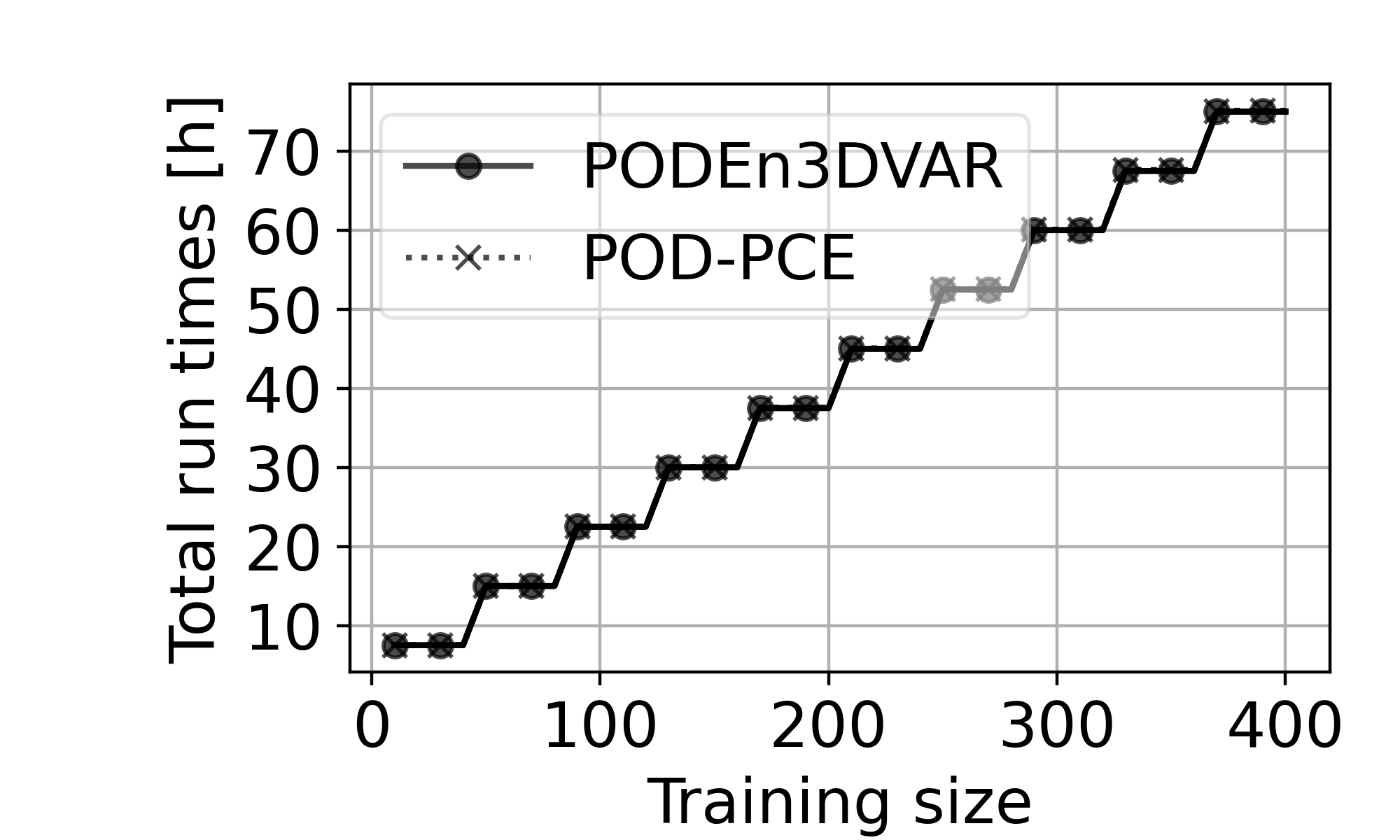}}
    \caption{Evolution of calibration and total time (including simulations time) with training size by BLUE-POD and POD-PCE algorithms}
    \label{fig:application:measuerment:runTimes}
\end{figure}

Examples of calculation times are given in Figure \ref{fig:application:measuerment:runTimes}.  It can be seen in Figure \ref{fig:application:measuerment:runTimes}-a that the POD-PCE model calibration time, including POD construction (eigenvalue problem) and PCE fitting for each POD pattern (choice of optimal degree and learning), is at most $10$ minutes for the considered ensemble sizes. Effective calibration time for metamodel-based 3DVAR, showed in Figure \ref{fig:application:measuerment:runTimes}-b, therefore principally consists in computational time required for model runs. As these can be performed $40$ at a time, run time increases in stairs. 

\begin{figure}[H]
  \centering
 \subfloat[][Free surface $z_s$]{\includegraphics[trim={0.8cm 0cm 0.5cm 0.6cm},clip,width=0.35\textwidth]{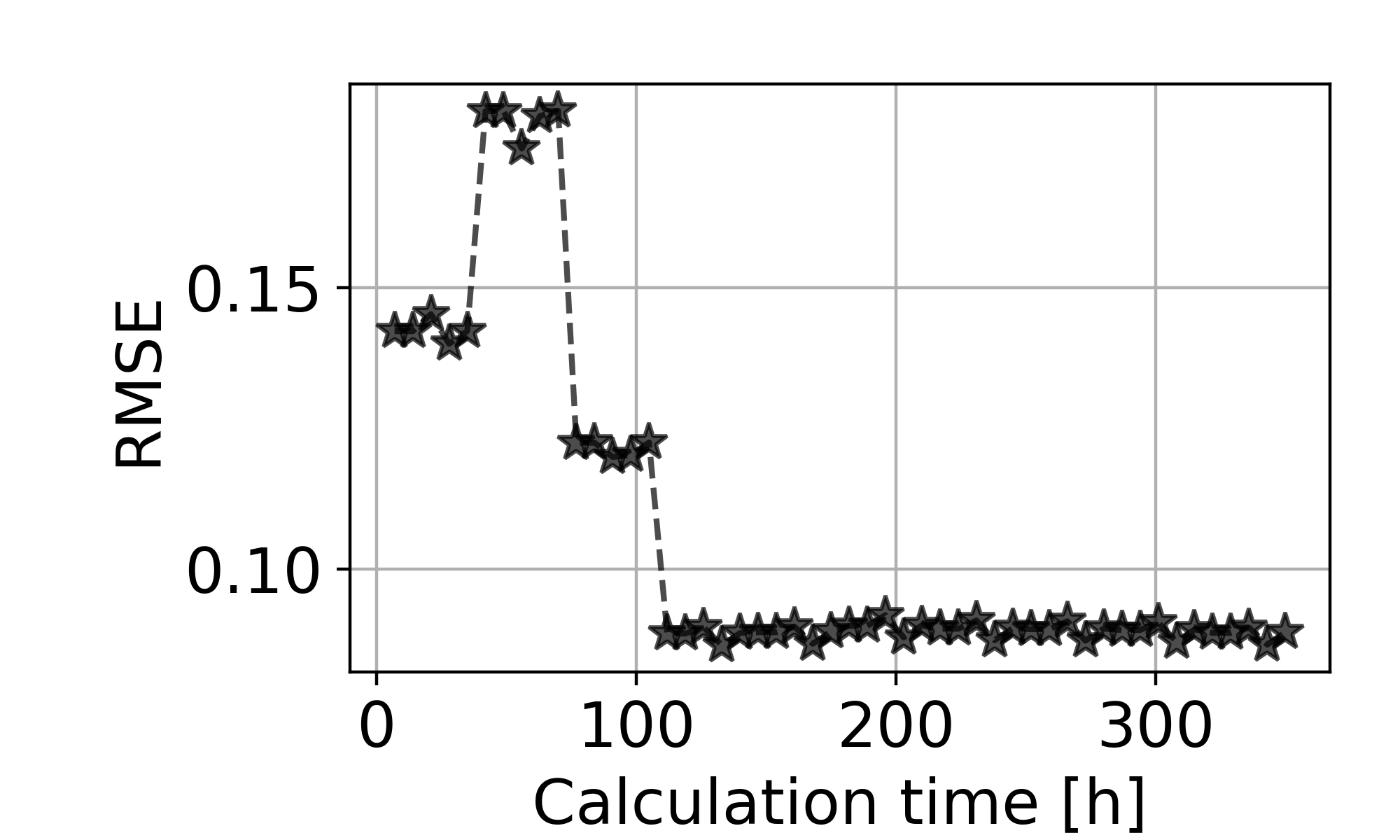}}
 \subfloat[][X-velocity $u$]{\includegraphics[trim={0.8cm 0cm 0.5cm 0.6cm},clip,width=0.35\textwidth]{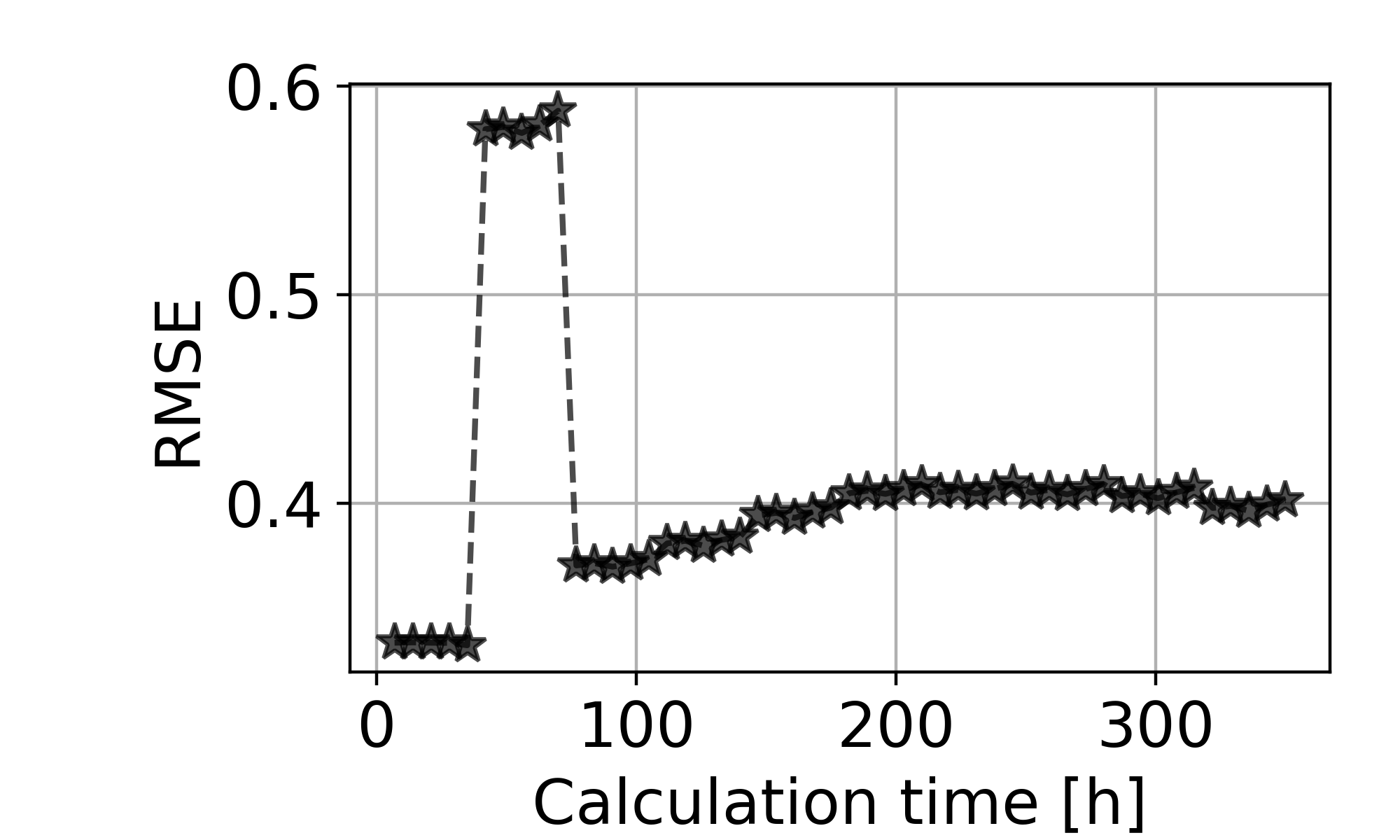}}
 \subfloat[][Y-velocity $v$]{\includegraphics[trim={0.8cm 0cm 0.5cm 0.6cm},clip,width=0.35\textwidth]{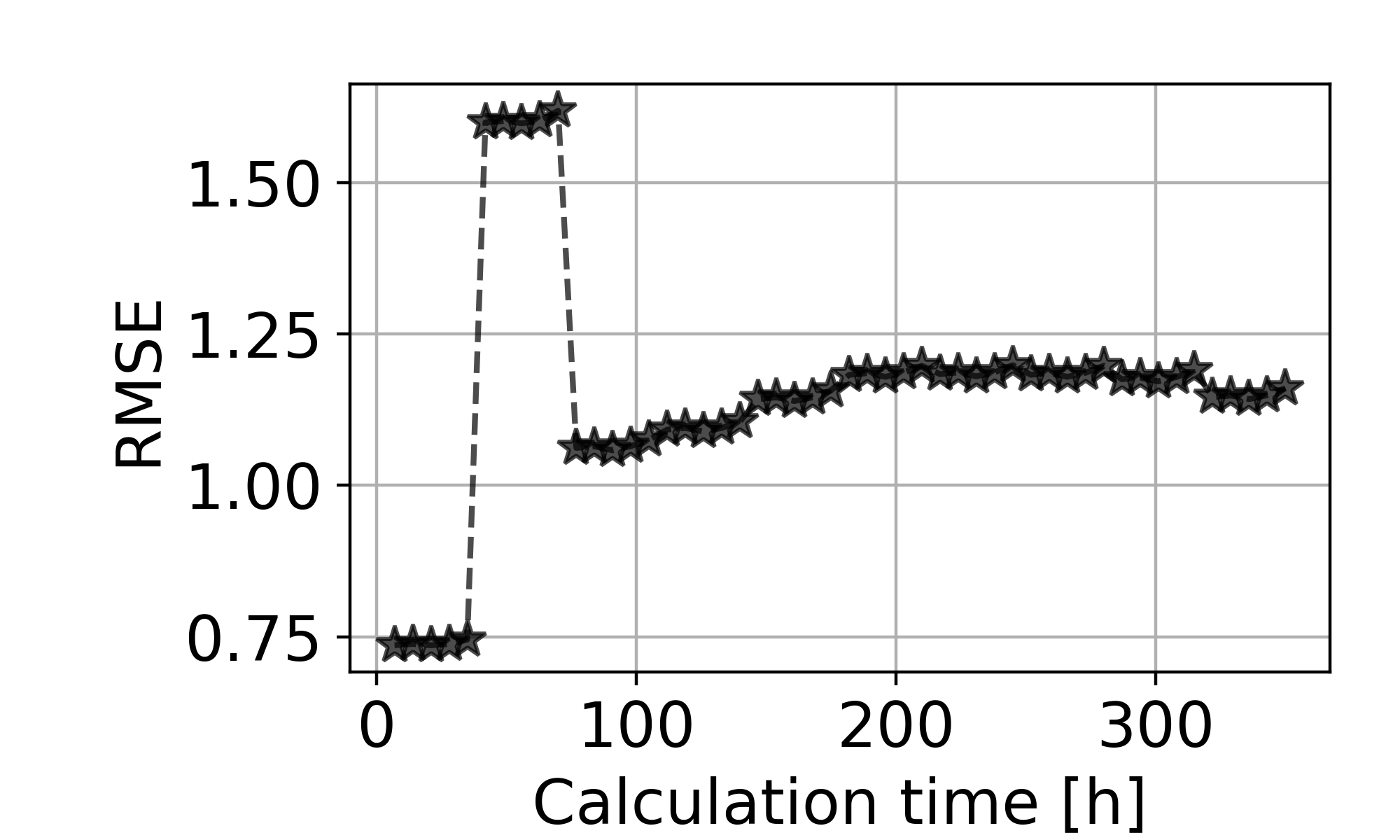}}
    \caption{Evolution of RMSE with calibration time for the three state variables on point 5 using classical 3DVAR}
    \label{fig:application:measuerment:3DVAR}
\end{figure}

In the end, around $80$ hours are necessary for a POD-PCE or PODEn3DVAR calibration, with an ensemble of size $400$. In comparison, evolution of RMSE using classical 3DVAR is shown in Figure \ref{fig:application:measuerment:3DVAR}. It can be noticed for example that convergence is reached after 200 hours of descent time, which is already much higher than the used run time with the proposed algorithms. 

However, in terms of needed number of simulations to reach an optimal analysis, classical 3DVAR is superior, as $40$ simulations were already sufficient in this particular case, compared to $300$ simulations for POD-PCE. As a conclusion, choice between classical 3DVAR and POD-PCE-3DVAR could be led by the importance of computational time vs. computational resources. 


\bibliographystyle{abbrvnat}
\bibliography{Refs}

\begin{thebibliography}{65}
\providecommand{\natexlab}[1]{#1}
\providecommand{\url}[1]{\texttt{#1}}
\expandafter\ifx\csname urlstyle\endcsname\relax
  \providecommand{\doi}[1]{doi: #1}\else
  \providecommand{\doi}{doi: \begingroup \urlstyle{rm}\Url}\fi

\bibitem[Altaf et~al.(2013)Altaf, Gharamti, Heemink, and Hoteit]{Altaf2013}
M.~Altaf, M.~E. Gharamti, A.~Heemink, and I.~Hoteit.
\newblock A reduced adjoint approach to variational data assimilation.
\newblock \emph{Computer Methods in Applied Mechanics and Engineering},
  254:\penalty0 1 -- 13, 2013.

\bibitem[Altaf et~al.(2009)Altaf, Heemink, and Verlaan]{Altaf2009}
M.~U. Altaf, A.~W. Heemink, and M.~Verlaan.
\newblock Inverse shallow-water flow modeling using model reduction.
\newblock \emph{International journal for multiscale computational
  engineering}, 7\penalty0 (6), 2009.

\bibitem[Asch et~al.(2016)Asch, Bocquet, and Nodet]{Asch2016}
M.~Asch, M.~Bocquet, and M.~Nodet.
\newblock \emph{Data assimilation: methods, algorithms, and applications}.
\newblock SIAM, 2016.

\bibitem[Bannister(2017)]{Bannister2017}
R.~Bannister.
\newblock A review of operational methods of variational and
  ensemble-variational data assimilation.
\newblock \emph{Quarterly Journal of the Royal Meteorological Society},
  143\penalty0 (703):\penalty0 607--633, 2017.

\bibitem[Birolleau et~al.(2014)Birolleau, Po{\"e}tte, and Lucor]{Birolleau2014}
A.~Birolleau, G.~Po{\"e}tte, and D.~Lucor.
\newblock Adaptive bayesian inference for discontinuous inverse problems,
  application to hyperbolic conservation laws.
\newblock \emph{Communications in Computational Physics}, 16\penalty0
  (1):\penalty0 1--34, 2014.

\bibitem[Blatman(2009)]{Blatman2009}
G.~Blatman.
\newblock \emph{Adaptive sparse polynomial chaos expansions for uncertainty
  propagation and sensitivity analysis}.
\newblock PhD thesis, 2009.

\bibitem[Blatman and Sudret(2011)]{Blatman2011}
G.~Blatman and B.~Sudret.
\newblock Adaptive sparse polynomial chaos expansion based on least angle
  regression.
\newblock \emph{Journal of Computational Physics}, 230\penalty0 (6):\penalty0
  2345 -- 2367, 2011.

\bibitem[Blayo et~al.(2011)Blayo, Cosme, Nodet, and Vidart]{Blayo2011}
E.~Blayo, E.~Cosme, M.~Nodet, and A.~Vidart.
\newblock Introduction to data assimilation, 2011.

\bibitem[Cao et~al.(2007)Cao, Zhu, Navon, and Luo]{Cao2007}
Y.~Cao, J.~Zhu, I.~M. Navon, and Z.~Luo.
\newblock A reduced-order approach to four-dimensional variational data
  assimilation using proper orthogonal decomposition.
\newblock \emph{International Journal for Numerical Methods in Fluids},
  53\penalty0 (10):\penalty0 1571--1583, 2007.

\bibitem[Carrassi et~al.(2018)Carrassi, Bocquet, Bertino, and
  Evensen]{Carrassi2018}
A.~Carrassi, M.~Bocquet, L.~Bertino, and G.~Evensen.
\newblock Data assimilation in the geosciences: An overview of methods, issues,
  and perspectives.
\newblock \emph{Wiley Interdisciplinary Reviews: Climate Change}, 9\penalty0
  (5):\penalty0 e535, 2018.

\bibitem[Cheng et~al.(2019)Cheng, Argaud, Iooss, Lucor, and
  Pon{\c{c}}ot]{Cheng2019}
S.~Cheng, J.-P. Argaud, B.~Iooss, D.~Lucor, and A.~Pon{\c{c}}ot.
\newblock Background error covariance iterative updating with invariant
  observation measures for data assimilation.
\newblock \emph{Stochastic Environmental Research and Risk Assessment},
  33\penalty0 (11-12):\penalty0 2033--2051, 2019.

\bibitem[Cheng et~al.(2020)Cheng, Argaud, Iooss, Lucor, and
  Pon{\c{c}}ot]{Cheng2020error}
S.~Cheng, J.-P. Argaud, B.~Iooss, D.~Lucor, and A.~Pon{\c{c}}ot.
\newblock Error covariance tuning in variational data assimilation: application
  to an operating hydrological model.
\newblock \emph{Stochastic Environmental Research and Risk Assessment}, pages
  1--20, 2020.

\bibitem[Despr{\'e}s et~al.(2013)Despr{\'e}s, Po{\"e}tte, and
  Lucor]{Despres2013}
B.~Despr{\'e}s, G.~Po{\"e}tte, and D.~Lucor.
\newblock Robust uncertainty propagation in systems of conservation laws with
  the entropy closure method.
\newblock In \emph{Uncertainty quantification in computational fluid dynamics},
  pages 105--149. Springer, 2013.

\bibitem[Durbiano(2001)]{Durbiano2001}
S.~Durbiano.
\newblock \emph{Vecteurs caractéristiques de modèles océaniques pour la
  réduction d'ordre en assimilation de données}.
\newblock PhD thesis, 2001.

\bibitem[Egbert and Erofeeva(2002)]{Egbert2002}
G.~D. Egbert and S.~Y. Erofeeva.
\newblock {Efficient Inverse Modeling of Barotropic Ocean Tides}.
\newblock \emph{Journal of Atmospheric and Oceanic Technology}, 19\penalty0
  (2):\penalty0 183--204, 02 2002.
\newblock ISSN 0739-0572.

\bibitem[El~Moçayd(2017)]{ElMocayd2017}
N.~El~Moçayd.
\newblock \emph{La décomposition en polynôme du chaos pour l'amélioration de
  l'assimilation de données ensembliste en hydraulique fluviale}.
\newblock PhD thesis, 2017.

\bibitem[Evangelista et~al.(2017)Evangelista, Giovinco, and
  Kocaman]{Evangelista2017}
S.~Evangelista, G.~Giovinco, and S.~Kocaman.
\newblock A multi-parameter calibration method for the numerical simulation of
  morphodynamic problems.
\newblock \emph{Journal of Hydrology and Hydromechanics}, 65:\penalty0
  175--182, 06 2017.

\bibitem[Evensen(2009)]{Evensen2009}
G.~Evensen.
\newblock \emph{Data assimilation: the ensemble Kalman filter}.
\newblock Springer Science \& Business Media, 2009.

\bibitem[Garcia et~al.(2013)Garcia, El~Serafy, Heemink, and
  Schuttelaars]{Garcia2013}
T.~Garcia, G.~El~Serafy, A.~Heemink, and H.~Schuttelaars.
\newblock Towards a data assimilation system for morphodynamic modeling:
  Bathymetric data assimilation for wave property estimation.
\newblock \emph{Ocean Dynamics}, 63, 05 2013.

\bibitem[Gerbeau and Perthame(2000)]{GerbeauPertham2000}
J.-F. Gerbeau and B.~Perthame.
\newblock {Derivation of Viscous Saint-Venant System for Laminar Shallow Water;
  Numerical Validation}.
\newblock Research Report RR-4084, {INRIA}, 2000.
\newblock Projet M3N.

\bibitem[Hervouet(2007)]{Hervouet2007}
J.-M. Hervouet.
\newblock \emph{Hydrodynamics of Free Surface Flows: Modelling with the Finite
  Element Method}.
\newblock John Wiley \& Sons, Ltd.: Hoboken, NJ, USA, 2007.

\bibitem[Idier et~al.(2019)Idier, Bertin, Thompson, and Pickering]{Idier2019}
D.~Idier, X.~Bertin, P.~Thompson, and M.~D. Pickering.
\newblock Interactions between mean sea level, tide, surge, waves and flooding:
  mechanisms and contributions to sea level variations at the coast.
\newblock \emph{Surveys in Geophysics}, 40\penalty0 (6):\penalty0 1603--1630,
  2019.

\bibitem[{Karpatne} et~al.(2019){Karpatne}, {Ebert-Uphoff}, {Ravela}, {Babaie},
  and {Kumar}]{Karpatne2019}
A.~{Karpatne}, I.~{Ebert-Uphoff}, S.~{Ravela}, H.~A. {Babaie}, and V.~{Kumar}.
\newblock Machine learning for the geosciences: Challenges and opportunities.
\newblock \emph{IEEE Transactions on Knowledge and Data Engineering},
  31\penalty0 (8):\penalty0 1544--1554, Aug 2019.

\bibitem[Larnier et~al.(2020)Larnier, Monnier, Garambois, and
  Verley]{Larnier2020}
K.~Larnier, J.~Monnier, P.-A. Garambois, and J.~Verley.
\newblock River discharge and bathymetry estimation from swot altimetry
  measurements.
\newblock \emph{Inverse Problems in Science and Engineering}, pages 1--31,
  2020.

\bibitem[{Le Maitre} et~al.(2001){Le Maitre}, Knio, Najm, and
  Ghanem]{Lemaitre2001_a}
O.~P. {Le Maitre}, O.~M. Knio, H.~N. Najm, and R.~G. Ghanem.
\newblock A stochastic projection method for fluid flow: I. basic formulation.
\newblock \emph{Journal of Computational Physics}, 173\penalty0 (2):\penalty0
  481 -- 511, 2001.

\bibitem[{Le Maitre} et~al.(2002){Le Maitre}, Reagan, Najm, Ghanem, and
  Knio]{Lemaitre2002_b}
O.~P. {Le Maitre}, M.~T. Reagan, H.~N. Najm, R.~G. Ghanem, and O.~M. Knio.
\newblock {A Stochastic Projection Method for Fluid Flow: II. Random Process}.
\newblock \emph{Journal of Computational Physics}, 181\penalty0 (1):\penalty0 9
  -- 44, 2002.

\bibitem[Li and Xiu(2008)]{Li2008}
J.~Li and D.~Xiu.
\newblock On numerical properties of the ensemble kalman filter for data
  assimilation.
\newblock \emph{Computer Methods in Applied Mechanics and Engineering},
  197\penalty0 (43-44):\penalty0 3574--3583, 2008.

\bibitem[Li and Xiu(2009)]{Li2009}
J.~Li and D.~Xiu.
\newblock A generalized polynomial chaos based ensemble kalman filter with high
  accuracy.
\newblock \emph{Journal of computational physics}, 228\penalty0 (15):\penalty0
  5454--5469, 2009.

\bibitem[Lumley(1967)]{Lumley1967}
J.~L. Lumley.
\newblock The structure of inhomogeneous turbulent flows.
\newblock \emph{Atmospheric Turbulence and Radio Wave Propagation}, 1967.

\bibitem[Martin et~al.(2015)Martin, Balmaseda, Bertino, Brasseur, Brassington,
  Cummings, Fujii, Lea, Lellouche, Mogensen, et~al.]{Martin2015}
M.~J. Martin, M.~Balmaseda, L.~Bertino, P.~Brasseur, G.~Brassington,
  J.~Cummings, Y.~Fujii, D.~Lea, J.-M. Lellouche, K.~Mogensen, et~al.
\newblock Status and future of data assimilation in operational oceanography.
\newblock \emph{Journal of Operational Oceanography}, 8\penalty0
  (sup1):\penalty0 s28--s48, 2015.

\bibitem[Marzouk and Najm(2009)]{Marzouk2009}
Y.~M. Marzouk and H.~N. Najm.
\newblock Dimensionality reduction and polynomial chaos acceleration of
  bayesian inference in inverse problems.
\newblock \emph{Journal of Computational Physics}, 228\penalty0 (6):\penalty0
  1862--1902, 2009.

\bibitem[Marzouk et~al.(2007)Marzouk, Najm, and Rahn]{Marzouk2007}
Y.~M. Marzouk, H.~N. Najm, and L.~A. Rahn.
\newblock Stochastic spectral methods for efficient bayesian solution of
  inverse problems.
\newblock \emph{Journal of Computational Physics}, 224\penalty0 (2):\penalty0
  560--586, 2007.

\bibitem[Mons et~al.(2017)Mons, Margheri, Chassaing, and Sagaut]{Mons2017}
V.~Mons, L.~Margheri, J.-C. Chassaing, and P.~Sagaut.
\newblock Data assimilation-based reconstruction of urban pollutant release
  characteristics.
\newblock \emph{Journal of Wind Engineering and Industrial Aerodynamics},
  169:\penalty0 232 -- 250, 2017.

\bibitem[Morrow et~al.(2019)Morrow, Fu, Ardhuin, Benkiran, Chapron, Cosme,
  d’Ovidio, Farrar, Gille, Lapeyre, Le~Traon, Pascual, Ponte, Qiu, Rascle,
  Ubelmann, Wang, and Zaron]{Morrow2019}
R.~Morrow, L.-L. Fu, F.~Ardhuin, M.~Benkiran, B.~Chapron, E.~Cosme,
  F.~d’Ovidio, J.~T. Farrar, S.~T. Gille, G.~Lapeyre, P.-Y. Le~Traon,
  A.~Pascual, A.~Ponte, B.~Qiu, N.~Rascle, C.~Ubelmann, J.~Wang, and E.~D.
  Zaron.
\newblock Global observations of fine-scale ocean surface topography with the
  surface water and ocean topography (swot) mission.
\newblock \emph{Frontiers in Marine Science}, 6:\penalty0 232, 2019.

\bibitem[Morvan et~al.(2008)Morvan, Knight, Wright, Tang, and
  Crossley]{Morvan2008}
H.~Morvan, D.~Knight, N.~Wright, X.~Tang, and A.~Crossley.
\newblock The concept of roughness in fluvial hydraulics and its formulation in
  1d, 2d and 3d numerical simulation models.
\newblock \emph{Journal of Hydraulic Research}, 46:\penalty0 191--208, 03 2008.

\bibitem[Mouradi et~al.(2020)Mouradi, Thual, Goeury, , Tassi, and
  Zaoui]{Mouradi2020}
R.-S. Mouradi, O.~Thual, C.~Goeury, , P.~Tassi, and F.~Zaoui.
\newblock Sensitivity of tidal modelling in coastal configurations: an
  uncertainty study based on field-measurement reduction.
\newblock \emph{Proceedings of the TELEMAC-MASCARET User Conference 2020},
  2020.

\bibitem[Mouradi et~al.(2021)Mouradi, Goeury, Thual, Zaoui, and
  Tassi]{Mouradi2021}
R.-S. Mouradi, C.~Goeury, O.~Thual, F.~Zaoui, and P.~Tassi.
\newblock Physically interpretable machine learning algorithm on
  multidimensional non-linear fields.
\newblock \emph{Journal of Computational Physics}, 428:\penalty0 110074, 2021.
\newblock ISSN 0021-9991.
\newblock \doi{https://doi.org/10.1016/j.jcp.2020.110074}.

\bibitem[Muller(2008)]{Muller2008_phd}
M.~Muller.
\newblock \emph{On the POD method: an abstract investigation with applications
  to reduced-order modeling and suboptimal control}.
\newblock PhD thesis, 2008.

\bibitem[Navon(2009)]{Navon2009}
I.~M. Navon.
\newblock Data assimilation for numerical weather prediction: a review.
\newblock In \emph{Data assimilation for atmospheric, oceanic and hydrologic
  applications}, pages 21--65. Springer, 2009.

\bibitem[Park and Jung(2001)]{Park2001}
H.~Park and W.~Jung.
\newblock The karhunen–loève galerkin method for the inverse natural
  convection problems.
\newblock \emph{International Journal of Heat and Mass Transfer}, 44\penalty0
  (1):\penalty0 155 -- 167, 2001.

\bibitem[Pham and Lyard(2012)]{Pham2012}
C.-T. Pham and F.~Lyard.
\newblock Use of tidal harmonic constants databases to force open boundary
  conditions in telemac.
\newblock In \emph{Proceedings of the XIXth TELEMAC-MASCARET User Conference
  2012, 18 to 19 October 2012, St Hugh's College, Oxford}, pages 165--172,
  2012.

\bibitem[Qian et~al.(2016)Qian, Lv, Cao, and Shao]{Qian2016}
S.~Qian, X.~Lv, Y.~Cao, and F.~Shao.
\newblock Parameter estimation for a 2d tidal model with pod 4d var data
  assimilation.
\newblock \emph{Mathematical Problems in Engineering}, 2016, 2016.

\bibitem[Robert et~al.(2005)Robert, Durbiano, Blayo, Verron, Blum, and {Le
  Dimet}]{Robert2005}
C.~Robert, S.~Durbiano, E.~Blayo, J.~Verron, J.~Blum, and F.-X. {Le Dimet}.
\newblock A reduced-order strategy for 4d-var data assimilation.
\newblock \emph{Journal of Marine Systems}, 57\penalty0 (1):\penalty0 70 -- 82,
  2005.

\bibitem[Rochoux et~al.(2014)Rochoux, Ricci, Lucor, Cuenot, and
  Trouv{\'e}]{Rochoux2014a}
M.~C. Rochoux, S.~Ricci, D.~Lucor, B.~Cuenot, and A.~Trouv{\'e}.
\newblock Towards predictive data-driven simulations of wildfire spread--part
  i: Reduced-cost ensemble kalman filter based on a polynomial chaos surrogate
  model for parameter estimation.
\newblock \emph{Natural Hazards and Earth System Sciences}, 14\penalty0
  (11):\penalty0 2951--2973, 2014.

\bibitem[Rolnick et~al.(2019)Rolnick, Donti, Kaack, Kochanski, Lacoste,
  Sankaran, Ross, Milojevic-Dupont, Jaques, Waldman-Brown, et~al.]{Rolnick2019}
D.~Rolnick, P.~L. Donti, L.~H. Kaack, K.~Kochanski, A.~Lacoste, K.~Sankaran,
  A.~S. Ross, N.~Milojevic-Dupont, N.~Jaques, A.~Waldman-Brown, et~al.
\newblock Tackling climate change with machine learning.
\newblock \emph{arXiv preprint arXiv:1906.05433}, 2019.

\bibitem[Scott and Mason(2007)]{Scott2007}
T.~Scott and D.~Mason.
\newblock Data assimilation for a coastal area morphodynamic model: Morecambe
  bay.
\newblock \emph{Coastal Engineering}, 54\penalty0 (2):\penalty0 91 -- 109,
  2007.

\bibitem[S{\'e}n{\'e}gas et~al.(2001)S{\'e}n{\'e}gas, Wackernagel, Rosenthal,
  and Wolf]{Senegas2001}
J.~S{\'e}n{\'e}gas, H.~Wackernagel, W.~Rosenthal, and T.~Wolf.
\newblock Error covariance modeling in sequential data assimilation.
\newblock \emph{Stochastic environmental research and risk assessment},
  15\penalty0 (1):\penalty0 65--86, 2001.

\bibitem[Shenefelt et~al.(2002)Shenefelt, Luck, Taylor, and
  Berry]{Shenefelt2002}
J.~Shenefelt, R.~Luck, R.~Taylor, and J.~Berry.
\newblock Solution to inverse heat conduction problems employing singular value
  decomposition and model-reduction.
\newblock \emph{International Journal of Heat and Mass Transfer}, 45\penalty0
  (1):\penalty0 67 -- 74, 2002.

\bibitem[{Shom 2015}(2015)]{Shom2015}
{Shom 2015}.
\newblock {MNT Bathymétrique de façade Atlantique}.
\newblock \emph{(Projet Homonim)}, 2015.
\newblock \doi{http://dx.doi.org/10.17183/MNT_ATL100m_HOMONIM_WGS84}.

\bibitem[Sirovich(1987)]{Sirovich1987}
L.~Sirovich.
\newblock {Turbulence and the Dynamics of Coherent Structures: I, II and III}.
\newblock \emph{Quarterly Applied Mathematics}, 45:\penalty0 561, 1987.

\bibitem[Smith et~al.(2013)Smith, Thornhill, Dance, Lawless, C.~Mason, and
  K.~Nichols]{Smith2013}
P.~Smith, G.~Thornhill, S.~Dance, A.~Lawless, D.~C.~Mason, and N.~K.~Nichols.
\newblock Data assimilation for state and parameter estimation: Application to
  morphodynamic modelling.
\newblock \emph{Quarterly Journal of the Royal Meteorological Society},
  139:\penalty0 314--327, 01 2013.

\bibitem[Soize(2017)]{Soize2017}
C.~Soize.
\newblock \emph{Uncertainty quantification}.
\newblock Springer, 2017.

\bibitem[S{\o}rensen and Madsen(2004)]{Sorensen2004}
J.~V.~T. S{\o}rensen and H.~Madsen.
\newblock Data assimilation in hydrodynamic modelling: on the treatment of
  non-linearity and bias.
\newblock \emph{Stochastic Environmental Research and Risk Assessment},
  18\penalty0 (4):\penalty0 228--244, 2004.

\bibitem[Sudret(2008)]{Sudret2008}
B.~Sudret.
\newblock Global sensitivity analysis using polynomial chaos expansions.
\newblock \emph{Reliability Engineering \& System Safety}, 93\penalty0
  (7):\penalty0 964 -- 979, 2008.
\newblock ISSN 0951-8320.

\bibitem[{SWOT}(2021)]{Swot2021}
{SWOT}.
\newblock {Surface Water and Ocean Topography by NASA (National Aeronautics and
  Space Administration) and CNES (Centre National d'Etudes Spatiales) in
  partnership with CSA (Canadian Space Agency) and UKSA (UK Space Agency)}.
\newblock \url{https://swot.jpl.nasa.gov/home.htm}, 2021.

\bibitem[Taddei(2020)]{Taddei2020}
T.~Taddei.
\newblock A registration method for model order reduction: data compression and
  geometry reduction.
\newblock \emph{SIAM Journal on Scientific Computing}, 42\penalty0
  (2):\penalty0 A997--A1027, 2020.

\bibitem[Taira et~al.(2017)Taira, Brunton, Dawson, Rowley, Colonius, McKeon,
  Schmidt, Gordeyev, Theofilis, and Ukeiley]{Taira2017}
K.~Taira, S.~L. Brunton, S.~T.~M. Dawson, C.~W. Rowley, T.~Colonius, B.~J.
  McKeon, O.~T. Schmidt, S.~Gordeyev, V.~Theofilis, and L.~S. Ukeiley.
\newblock Modal analysis of fluid flows: An overview.
\newblock \emph{AIAA Journal}, 55\penalty0 (12):\penalty0 4013--4041, 2017.

\bibitem[Tandeo et~al.(2018)Tandeo, Ailliot, Bocquet, Carrassi, Miyoshi,
  Pulido, and Zhen]{Tandeo2018}
P.~Tandeo, P.~Ailliot, M.~Bocquet, A.~Carrassi, T.~Miyoshi, M.~Pulido, and
  Y.~Zhen.
\newblock Joint estimation of model and observation error covariance matrices
  in data assimilation: a review.
\newblock \emph{Monthly Weather Review}, 2018.

\bibitem[Tian et~al.(2008)Tian, Xie, and Dai]{Tian2008}
X.~Tian, Z.~Xie, and A.~Dai.
\newblock An ensemble-based explicit four-dimensional variational assimilation
  method.
\newblock \emph{Journal of Geophysical Research: Atmospheres}, 113\penalty0
  (D21), 2008.

\bibitem[Torre et~al.(2019)Torre, Marelli, Embrechts, and Sudret]{Torre2019}
E.~Torre, S.~Marelli, P.~Embrechts, and B.~Sudret.
\newblock Data-driven polynomial chaos expansion for machine learning
  regression.
\newblock \emph{Journal of Computational Physics}, 388:\penalty0 601 -- 623,
  2019.

\bibitem[Vermeulen and Heemink(2006)]{Vermeulen2006}
P.~T.~M. Vermeulen and A.~W. Heemink.
\newblock {Model-Reduced Variational Data Assimilation}.
\newblock \emph{Monthly Weather Review}, 134\penalty0 (10):\penalty0
  2888--2899, 10 2006.

\bibitem[Wan and Karniadakis(2006)]{WanKarniadakis2005}
X.~Wan and G.~Karniadakis.
\newblock An adaptive multi-element generalized polynomial chaos method for
  stochastic differential equations.
\newblock \emph{Journal of Computational Physics}, 2006.

\bibitem[Xiu and Karniadakis(2002)]{XiuKarniadakis2002}
D.~Xiu and G.~E. Karniadakis.
\newblock The wiener--askey polynomial chaos for stochastic differential
  equations.
\newblock \emph{SIAM Journal on Scientific Computing}, 24\penalty0
  (2):\penalty0 619--644, 2002.

\bibitem[Xiu and Karniadakis(2003)]{XiuKarniadakis2003_flow}
D.~Xiu and G.~E. Karniadakis.
\newblock Modeling uncertainty in flow simulations via generalized polynomial
  chaos.
\newblock \emph{Journal of Computational Physics}, 187\penalty0 (1):\penalty0
  137 -- 167, 2003.

\bibitem[Zhu et~al.(1997)Zhu, Byrd, Lu, and Nocedal]{Zhu1997}
C.~Zhu, R.~H. Byrd, P.~Lu, and J.~Nocedal.
\newblock Algorithm 778: L-bfgs-b: Fortran subroutines for large-scale
  bound-constrained optimization.
\newblock \emph{ACM Transactions on Mathematical Software (TOMS)}, 23\penalty0
  (4):\penalty0 550--560, 1997.

\end{thebibliography}

\end{document}